\documentclass{article}
\usepackage[utf8]{inputenc}
\usepackage[margin=1.25in]{geometry}

\usepackage{subfigure}
\usepackage{wrapfig}

\usepackage[colorlinks=true,linkcolor=black,citecolor=blue,allbordercolors={1 1 1}]{hyperref}
\usepackage{graphicx,url,multirow,array,comment}
\usepackage{amsfonts}       
\usepackage{nicefrac}       
\usepackage{microtype}      
\usepackage{amsthm}
\usepackage{amsmath,mathtools}
\usepackage{empheq}
\usepackage{amssymb}
\usepackage{enumitem}
\usepackage[ruled,vlined]{algorithm2e}
\usepackage{bm}
\usepackage{xspace}
\usepackage{multirow}
\usepackage{color}
\usepackage[usenames,dvipsnames]{xcolor}
\usepackage{hyperref}
\usepackage{booktabs,tabularx}
\usepackage{bold-extra}

\usepackage[flushleft]{threeparttable} 

\PassOptionsToPackage{sort}{natbib}
\usepackage[square,numbers]{natbib}

\usepackage{pifont}

\theoremstyle{plain}

\newtheorem{theorem}{Theorem}

\newtheorem{lemma}{Lemma}

\newtheorem{remark}{Remark}

\PassOptionsToPackage{capitalise,noabbrev}{cleveref}
\usepackage{cleveref}
\crefname{equation}{}{}
\Crefname{equation}{}{}

\graphicspath{ {./images/} }

\usepackage[affil-sl]{authblk}

\makeatletter
\renewcommand\AB@affilsepx{,\quad \protect\Affilfont}
\makeatother

\newcommand{\vc}{\bm{c}}
\newcommand{\vx}{\bm{x}}

\newcommand{\obj}{F}
\newcommand{\objErm}{F^{\tiny{\mbox{{ERM}}}}}
\newcommand{\numClients}{\ensuremath{M}}

\newcommand{\localStep}{\tau}
\newcommand{\algsymbol}{\ensuremath{\mathcal{A}}}
\newcommand{\data}{\ensuremath{\mathcal{D}}}
\newcommand{\clientDist}{\ensuremath{\mathcal{P}}}

\newcommand{\activeClients}{\mathcal{S}}
\newcommand{\sgrad}{g}

\newcommand{\localChange}{\Delta}
\newcommand{\modelSize}{d}
\newcommand{\E}{\mathbb{E}}
\newcommand{\R}{\ensuremath{\mathbb{R}}}
\newcommand{\lr}{\eta}
\newcommand{\slr}{\lr_{s}}

\newcommand{\cC}{{\cal C}}

\newcommand{\Exs}{\ensuremath{{\mathbb{E}}}}

\newcommand{\ie}{\textit{i.e.,}\xspace}
\newcommand{\eg}{\textit{e.g.,}\xspace}

\newcommand{\fedavg}{\textsc{FedAvg}\xspace}
\newcommand{\fedavgm}{\textsc{FedAvgM}\xspace}
\newcommand{\fedsgd}{\textsc{FedSGD}\xspace}
\newcommand{\fedprox}{\textsc{FedProx}\xspace}
\newcommand{\fednova}{\textsc{FedNova}\xspace}
\newcommand{\fedopt}{\textsc{FedOpt}\xspace}
\newcommand{\scaffold}{\textsc{SCAFFOLD}\xspace}

\newcommand{\gd}{\textsc{GD}\xspace}
\newcommand{\sgd}{\textsc{SGD}\xspace}
\newcommand{\adam}{\textsc{Adam}\xspace}

\newcommand{\yogi}{\textsc{Yogi}\xspace}
\newcommand{\adagrad}{\textsc{Adagrad}\xspace}

\newcommand{\feddane}{\textsc{FedDane}\xspace}
\newcommand{\fedadam}{\textsc{FedAdam}\xspace}
\newcommand{\fedyogi}{\textsc{FedYogi}\xspace}

\newcommand{\mime}{\textsc{Mime}\xspace}
\newcommand{\mimelite}{\textsc{MimeLite}\xspace}
\newcommand{\fedpa}{\textsc{FedPA}\xspace}
\newcommand{\feddualavg}{\textsc{FedDualAvg}\xspace}

\newcommand{\adalter}{\textsc{AdaAlter}\xspace}
\newcommand{\secagg}{\textsc{SecAgg}\xspace}

\newcommand{\serveropt}{\textsc{ServerOpt}\xspace}
\newcommand{\clientopt}{\textsc{ClientOpt}\xspace}

\usepackage[colorinlistoftodos,prependcaption]{todonotes}

\definecolor{darkgreen}{rgb}{0,0.4,0.1}

\definecolor{darkred}{rgb}{0.6,0,0.2}

\title{A Field Guide to Federated Optimization}
\author[1]{Jianyu Wang$^*$}
\author[3]{Zachary Charles$^*$}
\author[3]{Zheng Xu\thanks{Jianyu Wang, Zachary Charles, Zheng Xu, Gauri Joshi, H. Brendan McMahan conceived, coordinated, and edited this work. Part of the work was done while Jianyu Wang was an intern at Google. Correspondence to Zheng Xu \{\url{xuzheng@google.com}\} and the section editors detailed in acknowledgements and notes.}}
\author[1]{Gauri Joshi$^*$}
\author[3]{H. Brendan McMahan$^*$}
\author[3]{Blaise Ag\"uera y Arcas}
\author[1]{Maruan Al-Shedivat} 
\author[3]{Galen Andrew} 
\author[13]{Salman Avestimehr} 
\author[3]{Katharine Daly}
\author[9]{Deepesh Data}
\author[9]{Suhas Diggavi} 
\author[3]{Hubert Eichner}
\author[1]{Advait Gadhikar}
\author[3]{Zachary Garrett}
\author[9]{Antonious M.~Girgis}
\author[8]{Filip Hanzely} 
\author[3]{Andrew Hard} 
\author[13]{Chaoyang He}
\author[4]{Samuel Horv\'{a}th}
\author[3]{Zhouyuan Huo}
\author[3]{Alex Ingerman} 
\author[2]{Martin Jaggi}
\author[10]{Tara Javidi}
\author[3]{Peter Kairouz} 
\author[3]{Satyen Kale} 
\author[2]{Sai Praneeth Karimireddy}
\author[3]{Jakub Kone\v{c}n\'{y}} 
\author[11]{Sanmi Koyejo} 
\author[1]{Tian Li}  
\author[3]{Luyang Liu} 
\author[3]{Mehryar Mohri}
\author[3]{Hang Qi}
\author[3]{Sashank J. Reddi}
\author[4]{Peter Richt\'{a}rik}
\author[3]{Karan Singhal}
\author[1]{Virginia Smith} 
\author[13]{Mahdi Soltanolkotabi} 
\author[3]{Weikang Song}
\author[3]{Ananda Theertha Suresh}
\author[2]{Sebastian U. Stich} 
\author[1]{Ameet Talwalkar} 
\author[14]{Hongyi Wang} 
\author[8]{Blake Woodworth}
\author[3]{Shanshan Wu} 
\author[3]{Felix X. Yu} 
\author[6]{ Honglin Yuan} 
\author[3]{Manzil Zaheer} 
\author[5]{Mi Zhang} 
\author[3,7]{Tong Zhang} 
\author[3]{Chunxiang Zheng} 
\author[12]{Chen Zhu}
\author[3]{Wennan Zhu} 

\affil[1]{\small Carnegie Mellon University}
\affil[2]{\small \'{E}cole Polytechnique F\'{e}d\'{e}rale de Lausanne}
\affil[3]{\small Google Research}
\affil[4]{\small King Abdullah University of Science and Technology}
\affil[5]{\small Michigan State Univeristy}
\affil[6]{\small Stanford University}
\affil[7]{\small The Hong Kong University of Science and Technology}
\affil[8]{\small Toyota Technological Institute at Chicago}
\affil[9]{\small University of California, Los Angeles}
\affil[10]{\small University of California, San Diego}
\affil[11]{\small University of Illinois Urbana-Champaign}
\affil[12]{\small University of Maryland, College Park}
\affil[13]{\small University of Southern California}
\affil[14]{\small University of Wisconsin–Madison}
\date{}

\begin{document}
\maketitle

\begin{abstract}
Federated learning and analytics are a distributed approach for collaboratively learning models (or statistics) from decentralized data, motivated by and designed for privacy protection. The distributed learning process can be formulated as solving federated optimization problems, which emphasize communication efficiency, data heterogeneity, compatibility with privacy and system requirements, and other constraints that are not primary considerations in other problem settings. This paper provides recommendations and guidelines on formulating, designing, evaluating and analyzing federated optimization algorithms through concrete examples and practical implementation, with a focus on conducting effective simulations to infer real-world performance. The goal of this work is not to survey the current literature, but to inspire researchers and practitioners to design federated learning algorithms that can be used in various practical applications.  
\end{abstract}

\clearpage
\tableofcontents
\clearpage

\setlength{\parskip}{0.15cm}

\section{Introduction} \label{sec:intro}


Federated learning (FL) was initially proposed by \citet{mcmahan17fedavg} as an approach to solving learning tasks by a loose federation of mobile devices. However, the underlying idea of training models without collecting raw training data in a single location has proven to be useful in other practical scenarios. This includes, for example, learning from data silos such as businesses or medical institutions which cannot share data due to confidentiality or legal constraints, or applications in edge networks~\citep{yang2019federated, li2020federated,lim2020federated,he2020fedml}.
In light of this,  
\citet{kairouz2019advances} proposed a broader definition:
\begin{quote}
\textbf{Federated learning} is a machine learning setting where multiple entities (clients) collaborate in solving a machine learning problem, under the 
coordination of a central server or service provider. Each client's raw data is stored locally and not exchanged or transferred; instead, focused updates intended for immediate aggregation are used to achieve the learning objective.
\end{quote}

The goal of providing strong privacy protection is implicit in this definition, and a central motivator for FL. Storing data locally rather than replicating it in the cloud decreases the attack surface of the system, and the use of focused and ephemeral updates and early aggregation follow the principle of data minimization. Stronger privacy properties are possible when FL is combined with other technologies such as differential privacy and secure multiparty computation (SMPC) protocols such as secure aggregation. The need to adhere to these privacy principles and ensure compatibility with other privacy technologies puts additional constraints on the design space for federated optimization algorithms.

Federated learning has received increasing attention from both academic researchers and industrial practitioners. 
The field has exploded from a handful of papers in 2016 including \citep{mcmahan17fedavg, FEDOPT2016, FEDLEARN2016}, to over 3000 new publications using the term in 2020. Google makes extensive use of federated learning in the Gboard mobile keyboard for applications including next word prediction \citep{hard18gboard}, emoji suggestion~\citep{gboard19emoji} and out-of-vocabulary word discovery \citep{gboard19oov}. Federated learning is also being explored for improving the “Hey Google”  detection models in Assistant \citep{googleassistant2021}, suggesting replies in Android Messages \citep{androidmessages2020}, and improving user experience on Pixel phones \citep{googledps2020}.

Federated learning is being adopted by industrial practitioners beyond Google. Apple uses federated learning in iOS 13 for applications like the QuickType keyboard and the vocal classifier for “Hey Siri” \citep{apple19wwdc}. Applications of federated learning in the finance space are undertaken by WeBank for money laundering detection \citep{webank2020}, as well as by Intel and Consilient for financial fraud detection \citep{intel2020}. In the medical space, federated learning is used by MELLODDY consortium partners for drug discovery \citep{melloddy2020}, by NVIDIA for predicting COVID-19 patients’ oxygen needs \citep{nvidia2020}, by Owkin for medical images analysis \citep{owkin2020} and others.

A newer application, federated analytics (FA) \citep{federatedanalytics}, poses a class of problems different from the training of deep networks, but nevertheless potentially addressable by optimization methods. As defined in \citep{federatedanalytics}, FA is ``\emph{the practice of applying data science methods to the analysis of raw data that is stored locally on users’ devices. 
Like federated learning, it works by running local computations over each device’s data, and only making the aggregated results — and never any data from a particular device — available to product engineers.}''
Recent years have seen a dramatic rise in the training of large models from potentially private user data, driving demand for federated learning, but the collection of statistics and use of dashboards for software telemetry is even more ubiquitous, and is fraught with similar privacy concerns. 
FA thus represents what is likely to become a very common use case, and is already being used in products including Google Health Studies to power privacy-preserving health research~\citep{ghs2020}. Federated optimization algorithms can be highly relevant to federated analytics: FA can frequently be viewed as training (estimating) a small ``model'' that is often a simple statistic, such as a count or marginal distribution, rather than a complete distribution (as in a generative model) or conditional distribution (as in a classifier). Moreover, FA settings can present or accentuate challenges that are relevant to the design of optimization algorithms; we discuss some of these below.

There have been a number of recent works surveying, reviewing, and benchmarking federated learning, including~\citep{caldas2018leaf,yang2019federated,kairouz2019advances,li2020federated,lim2020federated,he2020fedml}. However, there remains a lack of consensus about core concepts in federated learning, such as the formalization of the optimization problem(s), the definition of data heterogeneity, system constraints, evaluation metrics, relevant experimental settings, and approaches to parameter tuning. This paper provides recommendation and commentary on these areas, among others.

Our goal is also to help bridge the gap between optimization theory and practical simulations, and between simulations and real-world systems. This paper emphasizes the practical constraints and considerations that can motivate the design of new federated optimization algorithms, rather than a specific set of knobs.  We do not present new theoretical results, but rather provide suggestions on how to formulate problems in federated optimization, which constraints may be relevant to a given problem, and how to go about empirically analyzing federated optimization. 
In short, we hope this paper can serve as a guide for researchers designing and evaluating new federated learning algorithms, as well as a concise handbook for federated optimization practitioners.

The goal for this guide is not to be a literature review, a comprehensive survey, nor an empirical benchmark of state-of-the-art methods. In \cref{sec:practical_algorithm_design,sec:theory}, in addition to the general suggestions, we briefly review advanced techniques in recent research for both practice and theory, and refer readers to related papers for more detailed discussion. The discussion of such techniques is intended to provide representative examples that apply our general guidelines, not to survey all the work in FL. The algorithms are often chosen due to the popularity of the method, as well as authors’ familiarity. We encourage readers to focus on the general ideas of how to formulate and analyze FL problems, rather than the specific algorithms discussed.

\paragraph{Disclaimer} The practical experience shared in this draft is biased towards the application on mobile devices and some of the data are based on Google's federated learning system \citep{bonawitz19sysml}.


\subsection{Federated Optimization}
\label{sec:federated_optimization}
Optimization methods for federated learning must contend with many key issues that centralized methods usually do not have to address; we focus on the common features here, and provide an incomplete list for further discussion in the following sections. Federated learning is interdisciplinary research and some other considerations such as fairness and personalization are also of interest and discussed in \cref{sec:generalized_problem} and \cref{sec:connection_to_other_topics}.

\begin{itemize}
   \item \emph{Communication efficiency.} Federated learning is one kind of distributed computing with decentralized data. Minimizing communication between the server and clients is desired both for system efficiency and to support the privacy goals of federated learning. Communication can be the main bottleneck when clients are mobile devices that have limited bandwidth and availability for connection. These needs have led to significant research, including to the development of a plethora of compression mechanisms for reducing communication costs, and improved optimization methods that can reduce the total number of communication rounds. Taking multiple local updates on private data before a global sync (as in federated averaging, see Section~\ref{sec:basics}) often achieves surprisingly good results in practice, though a full theoretical characterization of why such algorithms perform well in practice is still lacking. 

   \item \emph{(Data) heterogeneity.} Data is stored across the clients (which typically generate it in the first place), and will never be directly communicated to the server or other clients due to privacy reasons. The totality of data  is typically highly {\em imbalanced} (clients have different quantities of training data) and {\em statistically heterogeneous} (the training samples on clients may come from different distributions). In contrast, in conventional distributed optimization settings, a single curator partitions and distributes data among the workers to achieve various goals, such as load balancing, reduction of effective statistical heterogeneity, and so on. The non-IID (identically and independently distributed) data partitioning in federated settings can be challenging when the goal is to train a single global model for all clients, particularly under a limited communication budget. 
   
    Aspects of data heterogeneity can be particularly important in federated analytics where online learning or estimation are desirable. In such settings, a small but more time-varying model may be required, as opposed to a possibly static and large one. Analytics applications may require additional statistical rigor in addressing concerns raised from data heterogeneity like estimating confidence intervals for predictions or model parameters, time-of-day variation, and bias from various sources.

    \item \emph{Computational constraints.} Clients can have different local computation speeds due to inherent hardware differences, or due to competing background tasks, which can lead to challenging computational heterogeneity. If clients are mobile devices or use GPU accelerators, then there may also be strict memory limits for local computation.

    \item \emph{Privacy and security.} Privacy is an explicit goal in federated learning settings, and the definition of FL itself suggests that algorithms should constrain themselves to only access information from data through aggregates for this reason. Further, the ability to combine federated learning with other privacy technologies, e.g., differential privacy \citep{mcmahan18learning} and secure aggregation \citep{bonawitz2017practical}, is of great importance. For many applications, federated optimization methods need to be compatible with such privacy protection methods, or otherwise achieve similar privacy guarantees. 
    
    \item \emph{System complexity.} Practical FL systems are often complex, e.g., hierarchical aggregation architectures between the coordinating server in the data center and client edge devices. Various protocols are designed to build such complicated systems \citep{bonawitz19sysml,paulik2021federated} and they can introduce some implicit constraints for federated optimization.   For example, the existence of stragglers (devices that are slow to report updates to the server) can affect algorithmic choices on synchronous or asynchronous optimization. The robustness of the system has to depend on close collaboration of different FL components such as clients for data storage and local computation, communication channels and aggregators for collecting and transferring information, and server for orchestration and global computation. 
\end{itemize}


\subsection{Applications}\label{section:application_patterns}


There are many practical applications of the general idea of federated learning. We highlight two settings that may significantly affect the design of practical algorithms: {\em cross-device FL} and {\em cross-silo FL}. Cross-device FL targets machine learning across large populations of mobile devices, while cross-silo federated learning targets collaborative learning among several organizations. Both cross-device and cross-silo federated optimization have data and computational heterogeneity as well as communication and privacy constraints, but they also have some nuanced differences \citep[Table 1]{kairouz2019advances}:

\begin{itemize}
    \item \emph{Number of clients.} The total number of clients in cross-device FL is typically much larger than in cross-silo FL. The number of clients (organizations) in cross-silo FL could be as small as two, more often dozens or hundreds, while the number of clients (devices) in cross-device FL can easily go beyond millions. The amount of private data on each client in cross-device FL can be smaller than in cross-silo FL. This may also affect the heterogeneity of data, particularly if inherited clustering structures can be found in the large number of clients of cross-device FL.  
    
    \item \emph{Client availability.} Cross-silo clients are typically hosted in data centers, implying high availability likely enabling all the clients to participate in each round. In contrast, the large and intermittently-available population of typical cross-device settings implies the server can only access clients via some sampling process which can be thought of as given, and the server has only limited ability to control it. In cross-device settings, client availability to participate in training may also correlate with its local training data distribution, introducing diurnal variations that optimization algorithms may need to address \citep{eichner2019semi}.
   
    \item \emph{Connection topology.} Direct (non-server-mediated) peer-to-peer connections are generally less well-supported on mobile device platforms, whereas in cross-silo settings such direct client-to-client connections may be easier to establish. In cross-silo FL, researchers have explored a variety of communication patterns, e.g., decentralized FL \citep{tsitsiklis1986,lian17decentralized,wang2018cooperative,he2019central,SwarmSGD2019,koloskova2020unified,D-DIANA,OptDecentralized2020}, vertical FL \citep{Hydra,Hydra2,hardy2017private,cheng2019secureboost}, split learning \citep{osia2017hybrid,gupta2018distributed,vepakomma2018split,he2020group}, and hierarchical FL \citep{wainakh2020enhancing,liao2019federated,briggs2020federated}. 

    \item \emph{Computation and communication constraints.} The computation and communication constraints are more strict in cross-device settings. Edge devices often have limited computation and communication power, while clients in cross-silo setting can often be data centers that are able to deploy hardware accelerators and high bandwidth communication links. 
 
    \item \emph{Client-local computation state.} Given the large populations, client-level privacy goals, and necessity of sampling, in cross-device settings we typically assume devices have no identifier and might only participate once in the entire training process. Thus, in this setting algorithms without client-side state are generally necessary. In contrast, since most clients will participate in most rounds of cross-silo training, stateful algorithms are appropriate.
   
\end{itemize}

Note that cross-device and cross-silo settings, mainly defined by their system-imposed constraints, are only two of many possible modalities of federated learning. These two have been the subject of much of the research on federated learning to date, in part due to their immediate application to and practical adoption by industry. Other FL scenarios include networks where ultra-reliable low-latency communication is necessary, such as vehicles on the road \citep{samarakoon2018federated}, or wireless edge networks in general \citep{park2019wireless, lim2020federated, chen2020joint}.

\subsection{Organization and Background}
This paper is organized as follows. \Cref{sec:formulation} formulates the canonical federated learning problem and introduces the basic optimization framework. \Cref{sec:practical_algorithm_design} provides guidelines for developing practical algorithms and presents representative techniques for advanced federated learning algorithm design. \Cref{sec:evaluation} discusses the use of simulations to evaluate federated optimization algorithms, including concrete examples. \Cref{sec:system} provides suggestions for deploying FL algorithms on real-world systems. \Cref{sec:theory} reviews the theoretical tools and some recent advances for analyzing FL algorithms. \Cref{sec:connection_to_other_topics} draws connections between federated optimization and other important aspects of FL, such as privacy, robustness, personalization and fairness.

\paragraph{Optional background} This manuscript is intended to be a self-contained reference on federated optimization and related topics. However, given the wide scope of federated learning and the myriad system-level differences between federated settings, several sections of \citet{kairouz2019advances} may serve as additional background to the reader who wants more detail:
\cite[Table 1, Section 1]{kairouz2019advances} describes various characteristics and regimes of federated learning; 
\cite[Section 2]{kairouz2019advances} discusses applications beyond \Cref{section:application_patterns} including the unique characteristics of \emph{cross-silo} federated learning;
\cite[Section 3.2]{kairouz2019advances} provides an overview of federated optimization methods, theoretical convergence rates, and open problems in federated optimization;
\cite[Section 4.2]{kairouz2019advances} discusses privacy techniques that can be used in conjunction with federated optimization methods;
\cite[Sections 5 and 6]{kairouz2019advances} provides more discussion of robustness and fairness, and \cite[Section 7]{kairouz2019advances} introduces system challenges.

\section{Problem Formulation}\label{sec:formulation}


In the basic conception of federated learning, we would like to minimize the objective function,
\begin{align}
    \obj(\vx) = \Exs_{i \sim \clientDist}[ \obj_i(\vx)], \quad \text{where} \quad \obj_i(\vx) = \Exs_{\xi \sim \data_i}[f_i(\vx, \xi)], \label{eqn:global_obj}
\end{align}
where $\vx \in \mathbb{R}^\modelSize$ represents the parameter for the global model, $\obj_i: \mathbb{R}^\modelSize \rightarrow \mathbb{R}$ denotes the local objective function at client $i$, and $\clientDist$ denotes a distribution on the population of clients $\mathcal{I}$.  The local loss functions $f_i(\vx,\xi)$ are often the same across all clients, but the local data distribution $\data_i$ will often vary, capturing data heterogeneity.

Algorithms designed for the cross-device setting cannot directly compute $\obj(\vx)$ or $\nabla \obj(\vx)$ because they are assumed to only have access to a random sample $\activeClients$ of the clients in each communication round. However, the objective function $\obj(\vx)$ can be used as a mathematical object in the analysis of such an algorithm, or even computed numerically in a simulation as part of an empirical evaluation procedure. If we model cross-device FL with a fixed dataset, to match the population risk \Cref{eqn:global_obj}, we would typically use a held-out set of \emph{clients}\footnote{While in the cross-device setting the server cannot access client IDs to partition clients into disjoint train and test sets, devices can locally flip a coin to decide whether to participate in a training or test population.}, rather than a held-out set of examples for each client from a fixed set. We consider this a ``stylized'' scenario because while such a simplified model of client selection might be suitable for analyzing or comparing some optimization algorithms, other approaches where a client selection strategy is part of the algorithm will require a richer model of device availability and participation.

The cross-silo setting can generally be well-modeled as a finite number of clients, i.e., $\obj^{\mathrm{silo}}(\vx) = \sum_{i=1}^\numClients p_i \obj_i(\vx)$.
A train/test split can generally be made by splitting the per-client datasets into local train and test sets, with generalization measured against the held-out per-client data. 

In both cross-device and cross-silo setting, the objective function in \Cref{eqn:global_obj} can take the form of an empirical risk minimization (ERM) objective function with finite clients and each client has finite local data: 
\begin{align}
    \objErm(\vx) = \sum_{i=1}^\numClients p_i \objErm_i(\vx), \quad \text{where} \ \objErm_i(\vx)=\frac{1}{|D_i|}\sum_{\xi \in D_i}f_i(\vx,\xi) \ \text{and} \ \sum_{i=1}^\numClients p_i =1. \label{eqn:global_obj_erm}
\end{align}
Note that $\numClients=|\mathcal{I}|$ denotes the total number of clients and $p_i$ is the relative weight of client $i$. Setting $p_i=|D_i|/\sum_{i=1}^\numClients |D_i|$ makes the objective function $\objErm(\vx)$ equivalent to the empirical risk minimization objective function of the union of all the local datasets. The objective in \eqref{eqn:global_obj} is equal (in expectation) to the ERM objective that one would optimize centrally if we randomly selected some number of clients and constructed a central training dataset from the union of their local datasets.

Compared to the centralized training, we want to highlight several key properties of \Cref{eqn:global_obj,eqn:global_obj_erm}:
\begin{itemize}
	\item \textbf{Heterogeneous and imbalanced data}: The local
	datasets $D_i$’s can have different distributions and sizes. As a consequence, the local objectives $\obj_i(\vx)$’s can be different. For example, they may have arbitrarily different local minima.
	\item \textbf{Data privacy constraints}: The local datasets $D_i$'s cannot be shared with the server or shuffled across clients. 
	\item \textbf{Limited client availability (more common in cross-device FL)}: In cross-device FL, the number of clients $\numClients$  in \Cref{eqn:global_obj_erm}  can be  extremely large and  not even necessarily well-defined (do we count a device that exists, but never meets the eligibility criteria to participate in training?). At any given time, only a subset (typically less than $1\%$) of clients are available to connect with the server and participate in the training. The client distribution $\clientDist$, total number of clients $\numClients$ or the total number of data samples $\sum_{i=1}^\numClients |D_i|$ are not known a priori before the training starts. 
\end{itemize}


\subsection{Federated Optimization Basics}\label{sec:basics}

Problem \eqref{eqn:global_obj} can potentially be solved by {\em gradient descent} (\gd), which performs iterations of the form
$ \vx^{(t+1)} = \vx^{(t)} - \eta_t \nabla \obj(\vx^{(t)})$, $t=0,1,2,\dots,$,
where $\eta_t$ is an appropriately chosen learning rate. Under appropriate regularity conditions, we can swap differentiation and expectation, which gives the following formula for the gradient:
$\nabla \obj(\vx) = \nabla \Exs_{i \sim \clientDist}[ \obj_i(\vx)] =  \Exs_{i \sim \clientDist}[ \nabla\obj_i(\vx)]. $
Note that the gradient of the global loss function $\obj$ is equal to the expectation (or ``average'') of the gradients of the local functions $\obj_i$. 
In many federated learning settings, the clients can't communicate among themselves directly, but can communicate indirectly via an orchestrating server. 
When applying \gd to the ERM formulation \eqref{eqn:global_obj_erm}, server has to calculate  
$\nabla \objErm(\vx) = \sum_{i=1}^\numClients p_i \nabla \objErm_i(\vx)$ by weighted average of {\em all} the local gradients from the clients. 

While \gd can be conceptually applied in the context of FL, it is not used in practice for various constraints and considerations discussed in \cref{sec:intro}. A number of techniques can be used to enhance \gd to make it theoretically or practically efficient as a method for solving federated optimization problems. Moreover, many of these techniques are (orthogonal) enhancements that can be combined for a more dramatic effect. Having said that, many of the possible combinations are not well understood and are still subject of active research.

{\em Partial participation} is a requirement for cross-device FL and some cross-silo settings. In communication round $t$, only a (finite) subset $\activeClients^{(t)}$ of clients can connect to the server, and the update rule becomes
$ \vx^{(t+1)} = \vx^{(t)} - \eta_t \frac{1}{|\activeClients^{(t)}|}\sum_{i\in  \activeClients^{(t)}} \nabla \obj_i(\vx^{(t)}).$ In practice, the sequence of active clients $\activeClients^{(t)}$ is typically dictated by complicated circumstances beyond the control of the orchestrating server (for example mobile devices might only participate when idle, connected to particular unmetered networks, and charging; see \cref{sec:system}). In theory, assumptions on client sampling are necessary to guarantee convergence (see \cref{sec:theory} for more discussion), and partial participation can be expected to lead to an increase in the number of communication rounds. 
    
Independently of whether partial participation is necessary, clients can use the {\em stochastic approximation (SA)}, replacing the exact gradient of their local loss function with  an unbiased stochastic gradient $\sgrad_i(\vx^{(t)})$ such that $\Exs_{\xi \sim \data_i}[\sgrad_i(\vx^{(t)})]  = \nabla  \obj_i(\vx^{(t)})$.  
SA is preferred when the size of $D_i$ is large and the calculation of the exact gradient is inefficient. 

{\em Local steps} is a popular technique to reduce communication costs. Each active client updates their local model for $\tau_i$ steps before the server aggregates the model deltas $\Delta_i^{(t)}=\vx_i^{(t,\localStep_i)}-\vx^{(t)}.$ Combining partial participation, stochastic approximation and local steps leads to federated averaging, a popular practical algorithm for federated optimization, which is further discussed in \cref{subsec:fedavg}. We defer the discussion of additional techniques like compression, momentum and acceleration, adaptive method, and control variates to \cref{sec:practical_algorithm_design} and the theoretical analysis to \cref{sec:theory}.

\subsection{The Generalized Federated Averaging Algorithm} \label{subsec:fedavg}
A common algorithm to solve \eqref{eqn:global_obj} is \emph{federated averaging} (\fedavg), proposed by \citet{mcmahan17fedavg}. The algorithm divides the training process into rounds. At the beginning of the $t$-th round ($t \geq 0$), the server broadcasts the current global model $\vx^{(t)}$ to a \emph{cohort} of participants: a random subset of clients $\activeClients^{(t)}$ (often uniformly sampled without replacement in simulation). Then, each sampled client in the round's cohort performs $\localStep_i$ local SGD updates on its own local dataset and sends the local model changes $\Delta_i^{(t)}=\vx_i^{(t,\localStep_i)}-\vx^{(t)}$ to the server. Finally, the server uses the aggregated $\Delta_i^{(t)}$ to update the global model:
\begin{align}
    \vx^{(t+1)} 
    = \vx^{(t)} + \frac{\sum_{i \in \activeClients^{(t)}} p_i \Delta_i^{(t)}}{\sum_{i \in \activeClients^{(t)}} p_i}, \label{eqn:upadte_fedavg}
\end{align}
where $p_i$ is the relative weight of client $i$. The above procedure will repeat until the algorithm converges. In the \emph{cross-silo} setting where all clients participate in training on every round (each cohort is the entire population), we have $\activeClients^{(t)}=\{1,2,\dots,\numClients\}$. 

\begin{algorithm}[ht]
    \DontPrintSemicolon
    \SetKwInput{Input}{Input}
    \SetAlgoLined
    \LinesNumbered
    \Input{Initial model $\vx^{(0)}$; \clientopt, \serveropt with learning rate $\lr, \slr$}
     \For{$t \in \{0,1,\dots,T-1\}$ }{
      Sample a subset $\activeClients^{(t)}$ of clients\;
      
      \For{{\it \bf client} $i \in \activeClients^{(t)}$ {\it \bf in parallel}}{
        Initialize local model $\vx_i^{(t,0)}=\vx^{(t)}$\;
        \For {$k =0,\dots,\localStep_i-1$}{
            Compute local stochastic gradient $\sgrad_i(\vx_i^{(t,k)})$\;
            Perform local update $\vx_i^{(t,k+1)} = \clientopt(\vx_i^{(t,k)}, \sgrad_i(\vx_i^{(t,k)}), \lr, t)$\;
        }
        Compute local model changes $\localChange_i^{(t)} = \vx_i^{(t,\localStep_i)} - \vx_i^{(t,0)}$\;
      }
      Aggregate local changes $\localChange^{(t)} = \sum_{i \in \activeClients^{(t)}} p_i \localChange_i^{(t)} / \sum_{i \in \activeClients^{(t)}} p_i$\;
      Update global model $\vx^{(t+1)} = \serveropt(\vx^{(t)}, -\localChange^{(t)},\slr,t)$\;
     }
     \caption{Generalized \fedavg (also known as \fedopt ~\citep{reddi2021adaptive})}
     \label{algo:generalized_fedavg}
\end{algorithm}

\fedavg can be easily generalized to a flexible framework that allows the algorithm designer to change the client update rule~\citep{wang2020tackling,yuan2020federated,wang2021local}, the update rule of the global model~\citep{reddi2021adaptive,hsu2019measuring,Wang2020SlowMo}, or the aggregation method applied to updates~\citep{lin2020ensemble,he2020group}. In particular, \citet{reddi2021adaptive} proposed a generalized version of \fedavg, the pseudo-code of which is presented in \Cref{algo:generalized_fedavg}. The algorithm is parameterized by two gradient-based optimizers: \clientopt and \serveropt with client learning rate $\lr$ and server learning rate $\slr$, respectively. While \clientopt is used to update the local models, \serveropt treats the negative of aggregated local changes $-\Delta^{(t)}$ as a pseudo-gradient and applies it to the global model. The original \fedavg algorithm implicitly set \serveropt and \clientopt to be SGD, with a fixed server learning rate $\slr$ of $1.0$.

The \fedavg algorithm can be viewed as a generalization of Local SGD (also called local-update SGD or periodic averaging SGD), which is studied for reducing communication cost in classic distributed settings~\citep{mcdonald2010distributed,stich2018local,yu2019parallel,wang2018cooperative, LSGDunified2020}. Some distinguishing properties of \fedavg are that unlike classic distributed settings, only a subset of clients participate in each training round. In terms of analysis, convergence analyses of Local SGD often assume that the local data is homogeneous and each client performs the same number of local updates, which may not hold in the federated setting.

\subsection{Related Problem Formulations} \label{sec:generalized_problem}

Going beyond the canonical formulation of federated learning of \eqref{eqn:global_obj}, several alternative formulations have been proposed in recent research.  We review some such alternative formulations below. This list is admittedly incomplete as federated optimization is rapidly being adopted for myriad new applications. There are several other possible generalizations of the federated learning problem formulation such as 1) federated neural architecture search \citep{he2020fednas} that searches for efficient neural architecture under FL setting, 2) considering time-varying models $\vx(t)$ that adapt to changes to the clients participating in training and their local training data, 3) hierarchical model training, and many more that are beyond the scope of this guide. 

\paragraph{Personalization and multi-task learning}  
The data heterogeneity in FL does not only present challenges to the design of provably efficient training methods for solving global objective~\eqref{eqn:global_obj}, but also inevitably raises questions about the utility of such a global solution to individual users. Indeed, a global model trained across all the data from all devices might be so different from the typical data and usage patterns experienced by an individual user such that the resulting model becomes worthless. In personalization, every client is allowed to have a different model that is adapted to their local data (i.e., a personalized model). One approach to learning a personalized model is to train a global model and use meta-learning to refine it and obtain personalized models \citep{chen2018federated,jiang2019improving, li2019differentially,fallah2020personalized}. Another line of work uses multi-task learning \citep{smith2017federated, evgeniou2004regularized, FL-personal-mixture2020, dinh2020personalized,ditto} to regularize local models towards the global average or towards some reference point. Section~\ref{sec:personalization} provides additional detail and discussion of personalized models.

\paragraph{Federated composite optimization}
Standard FL algorithms such as \fedavg are primarily geared towards \emph{unconstrained} settings, and $f_i$'s in \eqref{eqn:global_obj} are often assumed to be smooth for theoretical analysis. However, many potential FL applications involve non-smooth regularization terms to promote sparsity, low-rankness, monotonicity or enforce hard constraints. Such problems can be formulated as the following \emph{Federated Composite Optimization} (FCO) problem \citep{yuan2020federated2}: $\mathrm{min}_{\vx}\ \obj(\vx) + \psi(\vx)$, with $\obj(\vx) = \E_{i \sim \clientDist} [F_i(\vx)]$ and $\psi(\vx)$ a convex, possibly non-smooth, non-finite additive regularizer. 
See \citep{Nesterov-MP13,Flammarion.Bach-COLT17} for detailed discussions on the analysis and applications of classic composite optimization problems.
To approach the non-smooth (FCO) problem, \citet{yuan2020federated2} propose the \emph{Federated Dual Averaging} (\feddualavg) algorithm, which performs stochastic dual averaging during client local optimization and dual state aggregation during server aggregation. 

\paragraph{Probabilistic view}
Taking a Bayesian perspective~\citep{mackay1992bayesian}, we can formulate federated learning as inference of the \emph{global posterior distribution} over the model state $\vx$ given the data $\data$: $\mathbb{P}\left( \vx \mid \data \equiv \data_1 \cup \data_2 \cup \dots \right)$. 
If the posterior exists, it can be exactly decomposed into a multiplicative average of local posterior distributions (or sub-posteriors): $
    \mathbb{P}\left( \vx \mid \data \right) \propto \prod_{i=1}^M \mathbb{P}\left( \vx \mid \data_i \right)$.
Using this decomposition, \citet{alshedivat2020federated} show that the modes of the global posterior correspond to the optima of the objective $F(\vx)$ and propose to solve federated learning by using a combination of local posterior inference on the clients and global posterior inference on the server.
The proposed federated posterior averaging algorithm (\fedpa) resembles generalized \fedavg (Algorithm~\ref{algo:generalized_fedavg}), has the same computation and communication complexity, but also allows to effectively utilize local computation to de-bias client updates in heterogeneous settings (see discussion in Section~\ref{sec:reduce-bias-in-local-updates}). Taking this probabilistic view, \citet{lalitha2018fully} developed a posterior averaging algorithm to handle non-iid data and general communication graphs where a variational updating of Bayesian posterior density enables a general collaborative learning framework. Unfortunately, however, the variational methods are known to fall short in handling larger models. A handful of recent studies focus on stochastic differential equation (SDE) methods for decentralized sampling~\citep{gurbuzbalaban2021decentralized, parayil2020decentralized},  making the probabilistic approach more amenable to high dimensions.

\section{Practical Algorithm Design}
\label{sec:practical_algorithm_design}


In this section, we will discuss how to design practical federated learning algorithms. We first introduce some general guidelines in \Cref{sec:practical_guidelines} to help readers understand the challenges in designing practical algorithms, and then present several representative algorithms as examples in \Cref{sec:representative_techniques} that closely follow these guidelines and improve over vanilla \fedavg. When designing new FL algorithms, readers can use \Cref{sec:practical_guidelines} to check on the practicality and \Cref{sec:representative_techniques} to get inspirations from previous works.

\subsection{Guidelines for Developing Practical Algorithms}
\label{sec:practical_guidelines}


\subsubsection{Specify the Application Setting}
\label{subsec:application_settings}
Federated learning can apply to many different applications  including the cross-silo and cross-device settings discussed in \Cref{section:application_patterns}. These settings have distinct properties such that a single algorithm may not be suitable for all applications. When proposing new optimization algorithms, we recommend researchers clearly specify the most suitable application settings for their algorithms and consider the constraints and requirements of that specific setting.

\paragraph{Stateless clients in cross-device settings} In cross-device FL, it is often necessary to design algorithms where clients do not maintain persistent states across rounds. This is due to privacy requirements, but more so due to the fact that populations are typically very large, and so often a particular client will only participate in training a single time. Even in the best case, many rounds are likely to pass between the participations of a single client, and so state would likely be stale and adversely affect convergence.

Large client populations and intermittent availability also imply the server cannot compute any global information that would require an aggregation across all clients. Examples of such global information include the global objective value $\obj(\vx)$, the total number of clients $\numClients$, the total number of samples $\sum_{i \in \mathcal{I}}|\data_i|$, etc. This highlights the fact that the cross-device setting is better modeled by \eqref{eqn:global_obj}, which makes it clear that the above quantities are either ill-defined or can only be estimated.

\subsubsection{Improve Communication Efficiency}\label{sec:algo_comm_constraint}

In federated learning (especially in the cross-device setting), clients may experience severe network latency and bandwidth limitations. Practical FL algorithms generally use communication-reduction mechanisms to achieve a useful computation-to-communication ratio. In this subsection, we are going to introduce three common methods to reduce the communication cost: (i) reduce the communication frequency by allowing local updates; (ii) reduce communication volume by compressing messages; (iii) reduce communication traffic at server by limiting the participating clients per round. These three methods are orthogonal, and can be combined with each other.

\begin{remark}[Communication Efficiency]
In this subsection, we consider a method communication-efficient if it can reduce the communication cost per model local update (\ie per local iteration). However, the total communication cost is a product of number of local iterations and the communication cost per iteration. Therefore, when introducing the communication-efficient methods, we also discuss their impacts on the total number of iterations to achieve a target accuracy.
\end{remark}
\paragraph{Utilize multiple local updates}
When clients perform $\localStep$ model updates, the communication cost per client model update can be effectively reduced by a factor of $\localStep$. Many works have empirically validated the effectiveness of local updates in reducing the number of communication rounds to learn an accurate model~\citep{mcdonald2010distributed,moritz2015sparknet,zhang2016parallel,povey2014parallel,su2015experiments,mcmahan17fedavg, chaudhari2017parle,smith2018cocoa,wang2018adaptive,lin2018don}. While demonstrably effective, this can make analyzing the convergence behavior of methods employing local updates (such as \fedavg) more difficult than methods such as \sgd. In the case of \fedavg (and Local \sgd in many of these works), \citet{stich2018local,wang2018cooperative,yu2019parallel} show that there is an additional error term monotonically increasing with the local steps, though it is of a higher order compared to other terms and could be negligible when decaying the learning rate. When client data are non-IID, the side effects of the additional error term will be further exacerbated as shown in~\citep{li2019convergence,karimireddy2019scaffold,bayoumi2020tighter}. Accordingly, it is worth noting that there is a trade-off between communication reduction and convergence.
When using a larger number of local updates, the algorithm will communicate less frequently and save communication; however, it may incur higher error at convergence~\citep{charles2020learningrates, charles2021convergence}. This trade-off is succinctly captured by the lower bounds for FedAvg in \citep{karimireddy2019scaffold,woodworth2020minibatch}. In practice, one can select a best value of local steps to balance this trade-off or adapt it over time~\citep{wang2018adaptive}.

An alternative to taking multiple local steps is to use a large mini-batch size. Although the local-update SGD approach empirically outperforms the large mini-batch approach~\citep{lin2018don,mcmahan17fedavg}, its superiority is not well-understood from a theoretical perspective~\citep{woodworth2020localSGD}. We refer interested readers to \Cref{sec:fedavg-proof} for more discussion.

\paragraph{Employ compression techniques}
Reducing the bits transmitted between the clients and the server is another effective approach to save communication costs, and has spurred a large number of methods for sparsifying or quantizing the model updates without substantially reducing accuracy~\citep{wen2017terngrad, wang2018atomo, bernstein2018signsgd, vogels2019powersgd, rothchild2020fetchsgd, suresh2017distributed, FEDLEARN2016}. Compression techniques are especially effective for high-dimensional models.  We defer to \citep{xu2020survey} for a detailed survey of such methods. The output of compression methods can be viewed as stochastic estimates of the model update, and can therefore be \emph{unbiased} or \emph{biased}. Although biased compression methods can achieve higher compression ratios, their direct application can lead to convergence failures~\citep{alistarh2018sparse,stich2018sparse,karimireddy2019ef}. In settings with full participation and where clients have persistent state (\eg cross-silo FL), these errors can be provably mitigated using error-feedback strategies~\citep{Qsparse-local-sgd19,seide2014compr,strom2015scalable,lin2017deep,xie2020cser,karimireddy2019ef,stich2019,Koloskova2019-DecentralizedEC-2019,DoubleSqueeze2019,biased2020,EC-LSVRG,EC-Katyusha,EC-SGD,induced}. Both biased and unbiased compression methods, as well as error feedback schemes, can also be combined with optimization methods, such as momentum, for improved convergence~\citep{squarm_decentralized_20,vogels2019powersgd}. Another approach incorporates gradient compression into the model architecture itself. For instance, factorized model architectures using low-rank approximation can greatly reduce the amount of communication needed to communicate a model update~\citep{FEDLEARN2016,wang2021pufferfish}. 

Another important facet of compression methods to consider is their compatibility with aggregation protocols in distributed systems. For example, many distributed systems use all-reduce style techniques to aggregate gradients or local model updates across compute nodes~\citep{patarasuk2009bandwidth}. In such systems, compression techniques that are not compatible with all-reduce may provide less communication-efficiency, despite their higher compression ratio~\citep{vogels2019powersgd, wang2021pufferfish,agarwal2021utility}. In such settings, it is important to make sure that the compression operation \emph{commutes} with addition. That is, the sum of compressed gradients must be equal (at least in expectation) to the compression of a sum of gradients. This is also necessary for many federated aggregation protocols, such as secure aggregation~\citep{bonawitz2017practical}.

\paragraph{Sample a subset of clients}
Reducing the number of participating clients at each round by sampling a smaller subset can potentially reduce the communication cost as the per-round communication delay monotonically increases with the number of receivers/senders. Moreover, this strategy can mitigate the straggler effects when clients have random or heterogeneous computing speeds~\cite{dutta2018slow,chen2016revisiting}. For instance, if we assume that each client's computing delay is an exponential random variable, then the additional time of waiting for the slowest one is about $\mathcal{O}(\log M)$, where $M$ is the number of participating clients. It is important to note that the number of participating clients influence not only the communication time but also the convergence properties of the algorithm. Using too few clients per round may significantly increase the stochastic noise in the training process. How to set a proper number of participating clients is still open and less-explored in literature.

\paragraph{Evaluate communication-efficient methods in practice}
In real-world deployment, the communication efficiency of above methods should be evaluated via carefully designed wall-clock time measurement. The wall-clock time of one round of communication will be influenced by algorithmic design choices as well as system properties, including encoding/decoding overhead, fixed network latency (\eg time to establish handshake that is independent of the number of bits transmitted), and the per-bit communication time multiplied by the model size. Practitioners can choose the most suitable methods according to their specific settings (more discussions can be found in \Cref{sec:evaluate_communication,sec:basicmodel}). For example, compression methods are often evaluated in terms of the total number of bits transmitted. While useful as a rough reference, it does not account for the encoding/decoding times and the fixed network latency in the system. If these delays are much higher than the per-bit communication time, then the benefit of using compression can be incremental or negligible \citep{vogels2019powersgd}. On the other hand, if the per-bit communication time dominates other delays, the communication savings of compression can be significant. Also, many standard communication stacks will automatically compress network payloads using a  general purpose compression scheme like arithmetic coding~\citep{rissanen1979arithmetic} or the Lempel-Ziv method \citep{ziv1978compression}. So in practice, the gradient/model compression methods are typically used together with standard communication compression algorithms. Understanding the effects of this joint usage is highly useful but still remains under-explored.

\subsubsection{Design for Data and Computational Heterogeneity}
Due to the imbalance and heterogeneity of client data, the local objectives at clients are generally not identical and may not share common minimizers. Thus, when performing local updates from the same global model, clients will drift towards the minima of local objectives and end up with different local models. This phenomenon is often referred to as \emph{client drift}, and reduces the benefits of performing multiple local steps~\citep[e.g.][]{karimireddy2019scaffold,charles2021convergence}. Detailed discussions on the effects of client drift and its relation to data heterogeneity are given in \Cref{sec:theory}.

Another important but less well-studied source of heterogeneity in federated learning is computational heterogeneity.
Unlike centralized settings where computations are performed on dedicated and homogeneous machines, clients in federated settings may have varying compute speeds due to hardware differences. For example, some clients may be mobile phones, while others could be laptops equipped with GPUs. Similarly, some clients may dedicate their resources to federated training, while others may need to share resources with background tasks.

While many algorithms assume clients take the same number of steps, this may induce stragglers due to computational heterogeneity, and can increase the runtime of an algorithm. Production systems may enforce a timeout limit, causing the updates from stragglers to be discarded. To avoid such issues, \citet{hsu2020fedvision} caps the number of steps any client can take locally, but samples clients with probability proportionally to the amount of data they possess. Another solution is to allow variable number of local steps across clients. For example, \citet{li2018federated} and \citet{wang2020federated} propose allowing clients to perform as many local steps as possible within a given time window. Unfortunately, in the presence of data heterogeneity, \citet{wang2020tackling} show that this ``variable-step'' approach exacerbates client drift and causes a non-vanishing \emph{inconsistency} between the converged point of \fedavg-style algorithms and the minimizer of actual empirical loss function. They also show that this inconsistency caused by ``variable-step'' can be removed entirely with careful re-weighting of client updates.

In short, we recommend that researchers and practitioners consider allowing data heterogeneity and variable amounts of local computation by design. If taking this approach, we recommend modifying the client weighting strategy to account for this variability in steps (\eg giving less weight to clients which take more local steps \cite{wang2020tackling}). Besides, in some practical scenarios, clients may have very limited computation or memory capacity to finish even one local update. To address this, \citet{he2020group,horvath2021fjord} explored a strategy allowing clients use different sizes of models based on their own computation or memory budgets.

\subsubsection{Compatibility with System Architectures and Privacy-Preserving Protocols}
Practical federated learning systems are often complex and incorporate many privacy-preserving protocols (such as secure aggregation~\citep{bonawitz2017practical} and differential privacy mechanisms). This architectural complexity may introduce additional constraints for practical federated learning algorithms. While it is challenging to design an algorithm  compatible with all possible system architectures and privacy-preserving protocols, we recommend that researchers (i) clearly specify the targeted systems and privacy guarantees, (ii) discuss the limitations and incompatibilities of their methodology, and (iii) consider modifications of their proposed algorithms that would allow them to function in different systems. We discuss two examples of additional constraints due to system architectures and privacy-preserving protocols below.

\paragraph{Client inactivity and sampling strategies} While cross-device settings may have a large number of clients in the training population, the server may only be able to communicate with a small subset of clients at a given time due to various system constraints. In many experimental simulations, the cohort of clients used to perform training at a given round is obtained by sampling uniformly at random from the total population. In order to derive the theoretical guarantees, it is often assumed 
that all clients are always available to be sampled. This assumption is often not true in practical cross-device federated learning systems, where clients decide whether to be available based on their own status. For example, a mobile client could only be available when it is connected to WiFi and is charging \citep{bonawitz19sysml}. This constraint on client sampling is referred to as \emph{client inactivity} in \citet{ruan2020towards}. \citet{eichner2019semi} study a related scenario where there are two groups of clients that participate in training at different times of day. For example, users in different geographic locations may only be available at different times of day due to time zone differences (if clients only satisfy the mobile connection criteria at night). This non-uniform client availability leads to optimization challenges with significant practical relevance.

\paragraph{Privacy protection and weighting scheme}
Privacy protection methods are often applied to the aggregation process (client to server communication) of FL algorithms, for example, secure aggregation \citep{bonawitz2017practical} and differential privacy \citep{mcmahan18learning}. In many variants of federated averaging, the aggregation process can be formulated as a weighted summation of client updates (see equation \cref{eqn:upadte_fedavg}). A natural choice of the weights for aggregation is the number of examples on clients, so that the expectation of the weighted sum equals the batch gradient for the ERM objective \cref{eqn:global_obj} when each client only performs one iteration of gradient update (i.e., \fedsgd). Such sample-based weighting scheme is used in the original \fedavg algorithm \citep{mcmahan17fedavg} and many other federated learning papers~\cite[e.g.][]{li2019convergence,wang2020tackling}. 
However, non-uniform weighting can increase the influence of one client's update on the aggregated updated, which runs counter to privacy goals (and technically makes bounding sensitivity, a key requirement for differential privacy more challenging). In practice, DP-\fedavg \citep{mcmahan18learning} applies a uniform weighting scheme (i.e., $p_i=1/|\activeClients^{(t)}|$ in \cref{eqn:upadte_fedavg}) when applying differential privacy in federated learning. It remains an open question whether uniform weighting by itself can potentially benefit the privacy protection, or robustness to outliers,  or fairness to clients with small number of examples.

\subsection{Representative Techniques for Improving Performance}\label{sec:representative_techniques}
In this section we present a few representative techniques and optimization algorithms illustrating how various challenges in federated learning can be addressed. We emphasize that this is not meant to be an exhaustive list of federated optimization algorithms. Instead, the techniques and algorithms mentioned below are chosen to highlight design principles and elucidate settings where they can be applied.

\subsubsection{Incorporate Momentum and Adaptive Methods}
It is well-known that momentum and adaptive optimization methods (such as \adam \cite{kingma2014adam}, \adagrad \cite{duchi2011adaptive}, \yogi \cite{reddi2018adaptive}) have become critical components for training deep neural networks, and can empirically improve the optimization and generalization performance of SGD \citep{QIAN1999145,sutskever2013importance,zhang2020adaptive}. A natural question is whether one can carry over the benefits of these methods to the federated setting. To answer this question, many recent papers study how to incorporate momentum and adaptivity in FL, and validate their effectiveness in accelerating convergence.

\paragraph{Server and client optimization framework}
As discussed in \Cref{sec:formulation}, \citet{reddi2021adaptive} found that vanilla \fedavg implicitly enables a two-level optimization structure. Accordingly, they proposed a general algorithmic framework named \fedopt (see \Cref{algo:generalized_fedavg} for the pseudo-code). While clients use \clientopt to minimize the local training loss, the server takes the aggregated local model changes as a pseudo-gradient and uses it as input of \serveropt to update the global model. By setting \clientopt or \serveropt to be \textsc{SgdM}, \adam, \yogi, etc., the framework \fedopt naturally allows the use of any momentum and adaptive optimizers.

\paragraph{Server momentum and adaptive methods}
The idea of treating local changes as a pseudo-gradient can be traced back to \citep{chen2016scalable} and \citep{nichol2018first}, which apply it to distributed speech recognition and meta-learning, respectively. In the context of federated learning, \citet{Wang2020SlowMo,hsu2019measuring} show the effectiveness of server-side momentum (\ie \serveropt is \textsc{SgdM} and \clientopt is SGD) both theoretically and empirically. In intuition, the server-side momentum produces a similar effect to increasing the number of selected clients every round. Because of the exponentially weighted averaging of pseudo-gradients, model changes from clients selected in previous rounds also contributes to the global model update in the current round. 

In the same spirit, \citet{reddi2021adaptive} further demonstrate the benefits of using server-side adaptivity in cross-device federated learning. In particular, by setting \adam and \yogi as \serveropt and \sgd as \clientopt respectively, \fedadam and \fedyogi incorporate adaptive learning rates into federated optimization, and achieve the much faster convergence and higher validation performance than vanilla \fedavg on multiple benchmark federated learning datasets.  Moreover, the results show that the adaptive methods are easier to tune (\ie more robust to hyperparameter changes).

Another benefit of using server-side momentum or adaptive methods while keeping \clientopt as \sgd is that it does not increase the computational complexity of clients or the communication cost per round. It is also compatible with extremely low client sampling ratio (less than $1\%$). All the above benefits highlight the utility of server-side momentum or adaptive methods in cross-device FL settings.

\paragraph{Client momentum and adaptive methods}
Given the server-side momentum and adaptive methods, a natural question that arises is how can we apply momentum or adaptivity directly to the clients? Intuitively, using first- and second-order moments at each step might be better than using them every $\localStep$ steps as in server-side adaptive methods. However, when clients locally perform momentum or adaptive optimization methods, their optimizer states are separately updated, and hence, may deviate from each other due to the non-IID data distributions. To address this problem, a few works proposed the synchronized states strategy, where both the local models and the client optimizer states are synchronized by the server at each round. Examples include \citep{yu2019linear} and \citep{yuan2020federated}, both of which apply momentum at clients using the synchronized strategy and assuming all-clients participation. Recently, \citet{wang2021local} show that even when the client optimizer states are reset to the default values at the beginning of each round, client adaptive methods can be beneficial,  and it is possible to combine with the server adaptive methods to further boost the performance. Besides, this resetting states strategy will not incur any additional communications between the server and clients.


\paragraph{Lazy-update frameworks for momentum and adaptivity}
Beyond the server and client optimization framework, there also exist other approaches to apply adaptive optimization methods to FL. To avoid deviations among client optimizer states (\ie first- and second-order moments in adaptive optimizers), a common idea is to fix the optimizer states during local training and lazily update them on the server at the end of each round. Compared to the server-side adaptive methods mentioned above, there are two key differences: (i) The lazy-update algorithms allow the clients to locally use momentum or adaptive methods, but their optimizer states are fixed or keep synchronized; (ii) In these lazy-update algorithms, the optimizer states are updated on the server either by synchronizing all clients' local optimizer states or by using the batch gradient evaluated on the current global model. 

For instance, \citet{xie2020adaalter} proposed Local \adalter, which combines Local SGD with \adagrad via lazy updates of the optimizer states. In local \adalter, the clients use \adagrad instead of SGD to perform local updates. However, the optimizer states at different clients are always the same at each local iteration, as they are updated in the exact same way without any local information. This algorithm is designed for the setting where all clients can participate in every round of training and its effectiveness has been evaluated on classic distributed training tasks. Moreover, in order to synchronize the clients' optimizer states, it requires doubled communication cost compared to \fedavg and the server-side adaptive methods.

A similar idea of using lazily updated optimizer states at clients appears in \citep{karimireddy2020mime}. The authors proposed the \mime framework to adapt centralized stateful optimization methods to cross-device FL settings. The proposed algorithms split the model update and optimizer states update between the server and the clients. In particular, at each round, the server broadcast the global model $\vx$ and the optimizer states $\bm{s}$ to a random subset of clients. Then, the selected clients perform local updates according to the update rule of momentum or adaptive optimizer, but their optimizer states $\bm{s}$ remain unchanged during local training. Finally, the server uses the aggregated local model changes and the synchronized local batch gradient $\frac{1}{|\activeClients|}\sum_{i \in \activeClients} \nabla F_i(\vx)$ to update the global model and optimizer states, respectively. The above procedure repeats until convergence. It is worth noting that \citet{karimireddy2020mime} showed that using momentum in this way, along with an additional \emph{control variates} technique, can help to reduce the client drift caused by taking local steps. Furthermore, compared to server-side adaptive method (with a fixed server learning rate in simulation), \mime (and its more practical variant, \mimelite) can achieve better or comparable performance on several federated training tasks, albeit at the expense of additional communication and computation costs. 




\subsubsection{Reduce the Bias in Local Model Updates}\label{sec:reduce-bias-in-local-updates}
As mentioned, \fedavg and other federated optimization algorithms generally reduce the communication cost by allowing clients to take multiple SGD steps per round, before synchronizing with the server.
However, in settings with heterogeneous client data, taking more local steps in fact hinders convergence because the resulting client updates ($\Delta_i^{(t)}$'s in Algorithm~\ref{algo:generalized_fedavg}) become biased towards to the local minimizers. Here, we introduce two methods that can help to reduce the bias of each client's local model updates.

\paragraph{Control variates}
{\em Control variates} is a technique developed in standard convex optimization literature to reduce the variance of stochastic gradients in finite sum minimization problems and thereby speed up convergence. It is possible to adapt the technique to the federated learning setting to \emph{reduce the variance across clients}. A representative algorithm employing this technique in FL is \scaffold \citep{karimireddy2019scaffold}, as we describe below.

The \scaffold algorithm is applicable to the {\em cross-silo} setting and makes use of a persistent state stored with each client. This state is a control variate $\bm{c}_i$ for client $i$ which is meant to estimate the gradient of the loss with respect to the client's local data $\bm{c}_i \approx \nabla \obj_i(\vx)$. The server maintains the average of all client states as its own control variate, $\bm{c}$, which is communicated to all the selected clients in each round. Clients perform multiple steps of SGD in each round, just like \fedavg, but adding a correction term $\bm{c} - \bm{c}_i$ to each stochastic gradient. The effect of the correction is to \emph{de-bias} the update step on each client, ensuring they are much closer to the global update $ \nabla \obj_i(\vx) + \bm{c} - \bm{c}_i \approx  \nabla \obj(\vx)$. This enables \scaffold to provably converge faster than vanilla \fedavg without any assumptions on the data heterogeneity.

When implementing \scaffold, there are many possible choices of the control variates. For example, one can choose to use the the averaged local gradients in the last round as $\bm{c}_i$. Specifically, after local updates, the local control variate is updated as follows:
\begin{align}
    \bm{c}_i^{(t+1)} = 
    \begin{cases}
    \bm{c}_i^{(t)} - \bm{c}^{(t)} + \dfrac{1}{\lr \localStep}(\vx^{(t)} - \vx_i^{(t,\localStep)}), & \text{if} \ i \in \activeClients^{(t)} \\
    \bm{c}_i^{(t)}, & \text{otherwise}
    \end{cases} \label{eqn:scaffold_cv}
\end{align}
where the superscript $(t)$ denotes the index of communication round. This strategy is used in \citep{karimireddy2019scaffold}. It also appears in \citep{liang2019variance} when assuming all-clients participation. It is worth noting that \Cref{eqn:scaffold_cv} requires clients to have persistent states across rounds. In order to overcome this limitation, the \mime algorithm \cite{karimireddy2020mime} explores another option, that is, using the stochastic gradient evaluated on the global model $\nabla \obj_i(\vx; \xi_i)$ as the local variate $\bm{c}_i$ and the synchronized full-batch gradient $\frac{1}{|\activeClients|}\sum_{i \in \activeClients} \nabla\obj_i(\vx)$ as the global control variate $c$. By doing this, \mime can also reduce the variance in local mini-batch gradients and is applicable to the cross-device setting. 

\paragraph{Local posterior sampling}
Another way to de-bias client updates is computing client deltas as follows \citep{alshedivat2020federated}:
\begin{equation}
    \label{eq:fedpa-client-delta}
    \hat \Delta_i^{(t)} = \hat \Sigma_i^{-1} (\hat \mu_i - \vx^{(t)}),
\end{equation}
where $\hat \mu_i$ and $\hat \Sigma_i$ are estimates of the mean and covariance of the local posterior distribution on the $i$-th client, $p(\vx \mid \mathcal{D}_i)$. As the number of posterior samples used to estimate $\hat \mu_i$ and $\hat \Sigma_i$ increases, the bias in $\hat \Delta_i^{(t)}$ vanishes. In other words, when clients can use some extra compute per round, instead of running SGD for many steps or epochs (which would result in biased deltas in heterogeneous settings) we can instead run local Markov chain Monte Carlo (MCMC) to produce approximate local posterior samples and use them to reduce the bias. Under the uniform prior assumption, the global posterior can be estimated by the local posteriors.

\citet{alshedivat2020federated} designed \fedpa algorithm that allows to compute $\hat \Delta_i^{(t)}$ efficiently, by adding only a small constant factor computational overhead and no communication overhead compared to \fedavg, and used it as a drop-in replacement in Algorithm~\ref{algo:generalized_fedavg}.
\fedpa works with stateless clients, and is applicable for cross-device FL.
Methods similar to \fedpa have been developed in the distributed inference literature, before FL was introduced, such as consensus Monte Carlo~\citep{scott2016bayes} and expectation propagation (EP)~\citep{vehtari2020expectation, hasenclever2017distributed}.
In fact, EP can be seen as a generalization of \fedpa, although it requires stateful clients, and hence can be used only in cross-silo settings. The generalization to arbitrary graphs for collaborative learning was discussed in \citet{lalitha2018fully} and \citet{nedic2017distributed}; however, the challenges of communicating posteriors efficiently and accurately remain largely unaddressed.

\subsubsection{Regularize Local Objective Functions}
In vanilla \fedavg and \fedopt, clients perform multiple local updates at each round. As discussed in \Cref{sec:algo_comm_constraint}, while more local steps will reduce the average communication delay per iteration, it may incur additional errors at the end of training due to the heterogeneous local objectives. If we let clients perform too many local steps, then the local models will directly converge to the mimima of the local objectives, which can be far away from the global one. In order to avoid local models drift towards their local minima, a natural idea is to penalize local models that are far away from the global model by regularizing the local objectives. Formally, one can augment the local objective function at the $t$-th round as follows:
\begin{align}
    \widetilde{\obj}_i^{(t)}(\vx) = \obj_i(\vx) + \psi_i(\vx, \vx^{(t)}) \label{eqn:regularized_local_obj}
\end{align}
where $\vx^{(t)}$ is the global model and also the starting point of local training at round $t$, and function $\psi_i$ denotes a kind of distance measure between the local model $\vx$ and the current global model $\vx^{(t)}$. By carefully choosing the regularizer $\psi_i$, we can ensure the local models are close to the global one, or the surrogate local objectives \Cref{eqn:regularized_local_obj} have the same stationary points as the original global objective \Cref{eqn:global_obj}. Some examples of $\psi_i$ in literature are listed below.


\paragraph{Example: \fedprox~\citep{li2018federated}.} This algorithm uses the Euclidean distance between local models and the global model as the regularization function. That is, $\psi_i(\vx, \vx^{(t)}) = \frac{\mu}{2}\| \vx - \vx^{(t)}\|^2$ where $\mu \geq 0$ is a tunable hyperparameter. This easy form of regularization is particularly applicable to the cross-device setting of FL and does not need any additional computation or communication compared to vanilla \fedavg. Nonetheless, in theory, it cannot offer better rate of convergence than vanilla \fedavg, see the discussions and analyses in \citep{karimireddy2019scaffold,wang2020tackling,charles2021convergence}.
    
\paragraph{Example: \feddane~\citep{li2019feddane}.} Inspired by the \textsc{Dane} algorithm in classic distributed optimization, \citet{li2019feddane} further add a linear term to the regularization function as follows: 
    \begin{align}
        \psi_i(\vx, \vx^{(t)}) = \left\langle\frac{1}{|\activeClients^{(t)}|}\sum_{i \in \activeClients^{(t)}}\nabla \obj_i(\vx^{(t)}) - \nabla \obj_i(\vx^{(t)}), \vx - \vx^{(t)}\right\rangle + \frac{\mu}{2}\| \vx - \vx^{(t)}\|^2. \label{eqn:psi_feddane}
    \end{align}
    In order to compute the linear term (inner product) in \Cref{eqn:psi_feddane}, the selected clients need to compute the full-batch local gradients and synchronize them, resulting in doubled communication cost per round. Also, as noted by the authors, despite encouraging theoretical results, \feddane demonstrates underwhelming empirical performance compared to \fedprox and \fedavg. In certain cases (\eg client optimizer is SGD and $\mu=0$), \feddane has a similar client-side update rule to \mime~\citep{karimireddy2020mime}. Specifically, \feddane adds a correction term $\frac{1}{|\activeClients^{(t)}|}\sum_{i \in \activeClients^{(t)}}\nabla \obj_i(\vx^{(t)}) - \nabla \obj_i(\vx^{(t)})$ to each local stochastic gradient, while \mime uses $\frac{1}{|\activeClients^{(t)}|}\sum_{i \in \activeClients^{(t)}}\nabla \obj_i(\vx^{(t)}) - \nabla \obj_i(\vx^{(t)};\xi_i^{(t,k)})$ where $\xi_i^{(t,k)}$ denotes a random mini-batch sampled at local iteration $k$ and round $t$.
    
\paragraph{Example: \textsc{FedPD}~\citep{zhang2020fedpd} and \textsc{FedDyn}~\citep{acar2021federated}.} These two algorithms similarly modify the local objectives, and define the regularizer of local objective at client $i$ as follows:
\begin{align}
    \psi_i(\vx, \vx^{(t)}) = \langle\bm{\lambda}_i^{(t)}, \vx - \vx^{(t)}\rangle + \frac{\mu}{2}\| \vx - \vx^{(t)}\|^2 \label{eqn:psi_fedpd}
\end{align}
where $\bm{\lambda}_i^{(t)}$ is an auxiliary variable and updates at the end of each round: $\bm{\lambda}_i^{(t+1)} = \bm{\lambda}_i^{(t)} + \mu (\vx_i^{(t,\localStep_i)} - \vx^{(t)})$. By doing this, \citet{acar2021federated} proves that the surrogate local objectives, in the limit, have the same stationary points as the global objective. As a consequence, when the surrogate local functions are minimized exactly at each round, \textsc{FedDyn} can achieve faster convergence rate than \fedavg for convex smooth functions. Although the regularization \Cref{eqn:psi_fedpd} does not introduce extra communications, it requires clients to maintain persistent states or memory across rounds. Hence, it is more suitable for cross-silo FL settings.


\subsubsection{Consider Alternative Aggregation Methods}
As noted earlier, the global objective $\obj(\vx)$ defines a weighting over the individual client objectives $\obj_i(\vx)$. Typically, clients' local updates may be weighted by number of examples or weighted uniformly at each round. Though the weighting schemes may not influence the convergence rate of FL algorithms, they do control where the global model finally converges to. By tuning the weighting scheme, it is possible that the final global model is biased towards the minima of certain clients. We illustrate these effects in detail below.

\paragraph{Objective inconsistency problem}
When all clients participate in training at each round and have homogeneous computing speeds, the server just needs to aggregate all local changes in the same way as the global objective. However, when there is client sampling mechanism as in cross-device FL settings or clients take different local steps per round, the weighting scheme closely relates to many implementation choices. If the weighting scheme is not properly chosen, then the global model will instead converge to the stationary points of a surrogate objective, which is inconsistent with the original one. 

For example, \citet{li2019convergence} show that, in order to guarantee convergence, the weighting scheme should be selected according to the client sampling scheme. Specifically, if clients are sampled with replacement based on their local sample sizes, then the local model changes should be uniformly averaged; if clients are uniformly sampled without replacement, then the weights of each local model should be re-weighted according to their sample sizes. \emph{The key is that we need to ensure the expectation of aggregated local changes weights each local client in the same way as the global objective}. Otherwise, there is a non-vanishing term in the error upper bound, which does not approach to zero even if learning rate is small enough or gradually decayed~\citep{cho2020client}. Researchers often assume weighted sampling with replacement and uniform averaging for the simplicity of theoretical analysis. However, in real-world cross-device FL settings, the server has both limited knowledge of the population and limited control over clients sampling, and so weight in aggregation ends up effectively a time-varying function depending on the active clients set $\activeClients^{(t)}$ as shown in \Cref{algo:generalized_fedavg} (line 11). The convergence and consistency in practical scenarios remain an open problem. 

Recently, \citet{wang2020tackling} further showed that the weighting scheme is influenced by the computational heterogeneity across clients as well. If we still use the original weighting scheme, then When clients perform different number of local steps in vanilla \fedavg and many other FL algorithms, the algorithm will implicitly assign higher weights to the clients with more local steps. \citet{wang2020tackling} give an analytical expression of the surrogate objective function which the original \fedavg actually optimizes. The authors propose a simple method \fednova to eliminate the inconsistency between the surrogate loss and the original one by ``normalizing'' (or dividing) the local updates with the number of local steps. This method can also be considered as a new weighting scheme, as it equivalently assigns lower weights to the clients with more local steps. Another recent work \cite{wang2021local} further generalizes the above analysis by showing that the inconsistency problem can appear even with the same number of local steps when using adaptive methods (\eg \adagrad, \adam) on clients.

We should mention that while the objective inconsistency problem can greatly hurt the performance of FL algorithms even in some simple quadratic models~\citep{wang2020tackling}, its influence may differ across practical training tasks, depending on the properties of the datasets and neural network models. Due to the non-convex nature of neural networks, it is possible that the minimizer of the surrogate objective is also one of the original objective. On the other hand, if the convergence of the original global objective must be guaranteed in certain applications, then one may carefully choose the weighting scheme according to the suggestions above.


\paragraph{Beyond weighted averaging -- neuron matching algorithms}
In \fedavg, parameters of local models are averaged coordinate-wise with weights proportional to sizes of the client datasets~\citep{mcmahan17fedavg}. One potential shortcoming of \fedavg is that coordinate-wise averaging of weights may have drastic detrimental effects on the performance of the averaged model and adds significantly to the communication burden. This issue arises due to the permutation invariance of the hidden layers in a neural network. That is, for most neural networks, one can form many functionally equivalent networks (\ie each input is mapped to the same output) simply by permuting neurons within each layer. Unfortunately, applying coordinate-wise averaging to such functionally equivalent networks may result in a network that produces drastically different output values.

Several recent techniques address this problem by matching the neurons of client NNs before averaging them. For example, Probabilistic Federated Neural Matching (PFNM)~\citep{yurochkin2019bayesian} utilizes Bayesian non-parametric methods to adjust global model size according to the heterogeneity in the data. In many cases, PFNM has better empirical performance and communication efficiency than \fedavg. Unfortunately, structural limitations of the method restrict it to relatively simple neural architectures, such as fully connected neural networks with limited depth~\citep{wang2020federated}.

Federated Matched Averaging (FedMA) \citep{wang2020federated} and Model Fusion \citep{singh2020model} extend PFNM to other architectures, including CNNs and LSTMs. To address the reduced performance of PFNM on deeper networks, FedMA conducts a layer-wise neural matching scheme. First, the server gathers only the weights of the first layers from the clients and performs one-layer matching to obtain the first layer weights of the federated model. The server then broadcasts these weights to the clients, which proceed to train all consecutive layers on their datasets, keeping the matched federated layers frozen. This procedure is then repeated up to the last layer for which we conduct a weighted averaging based on the class proportions of data points per client. Empirically, FedMA has the potential to achieve better communication efficiency compared to \fedavg on several FL tasks~\citep{wang2020federated}.
In the same vein as FedMA and PFNM, \cite{singh2020model} propose the use of optimal transport to match the client neurons before averaging them.

\section{On Evaluating Federated Optimization Algorithms}
\label{sec:evaluation}

In this section, we discuss how to perform effective evaluations of federated optimization algorithms.
Our focus is on topics that are either not present in centralized algorithm evaluation, or exacerbated by facets of federated learning. In particular, we focus on \emph{simulated} federated learning settings, not actual deployments of federated learning. For discussion of production federated learning systems, see Section \ref{sec:system}. In this section, we aim to provide examples, discussions, and recommendations focused on federated learning research, though they may also be useful for the practitioner alike.

\subsection{Example Evaluations}
In order to facilitate this discussion and make it concrete, we implement our recommendations in representative simulated evaluations of three federated optimization algorithms across four datasets. 

\paragraph{Federated optimization methods}
We perform our empirical studies using three existing federated optimization algorithms. We refer to these as Algorithm A, Algorithm B, and Algorithm C. The interested reader can find the details of these algorithms in Appendix~\ref{appendix:empirical_details}.
We omit their names throughout this section since we are not interested in determining which algorithm outperforms the others. Instead, we are interested in how to effectively evaluate the methods. For the purposes of this section, the salient details are that all three algorithms are a special case of Algorithm \ref{algo:generalized_fedavg}, and employ the same amount of communication and client computation per round. Algorithms A, B, and C all have a client learning rate $\eta$ and server learning rate $\eta_s$, which we tune via grid-search. Instead of performing $\tau_i$ steps of \clientopt as in Algorithm \ref{algo:generalized_fedavg}, we perform $E$ epochs of training over each client's dataset. This is due to the fact that in practical settings, clients may have local datasets of varying size. Thus, specifying a fixed number of steps can cause some clients to repeatedly train on the same examples, and cause others to only see a small fraction of their data. This epoch-based approach was used in the first work on \fedavg~\citet{mcmahan17fedavg}. Unless specified, in a given figure, we set $E = 1$. 

\paragraph{Datasets} We use four federated datasets: GLD-23k, GLD-160k~\citep{gldv2, hsu2020fedvision}, CIFAR-10~\citep{krizhevsky2009learning}, and Stack Overflow~\citep{stackoverflow}. The Landmark datasets (GLD-23k and GLD-160k) and Stack Overflow dataset have naturally-arising federated structures. In the Landmark datasets, clients correspond to individual Wikipedia image contributors (for details, see the 2019 Landmark Recognition Challenge~\cite{weyand2020google}). In the Stack Overflow dataset, clients correspond to users of the Stack Overflow forum. For CIFAR-10, we create a non-IID, synthetic client structure by partitioning CIFAR-10 among 10 clients using a form of Dirichlet allocation \citep{hsu2019measuring}. The partitioning among clients is identical to that provided by TensorFlow Federated~\citep{tff}, which we mention because consistent partitioning is vital to deriving fair comparisons between methods. See Appendix \ref{appendix:datasets} for more details on all datasets.

\paragraph{Models and tasks} 
For the Landmark datasets, we predict the image labels using a version of MobileNetV2~\citep{sandler2018mobilenetv2}, replacing the batch norm layers with group norm~\citep{wu2018group}, as suggested in the context of federated learning by~\citet{hsieh2020non}. For CIFAR-10, we predict the image labels using ResNet-18 (again replacing the batch norm layers with group norm). For Stack Overflow, we train a moderately-sized transformer model to perform next-word-prediction.

\paragraph{Client sampling} For the Landmark datasets and Stack Overflow dataset, we only sample a fraction of the training clients at each round, thus simulating cross-device FL. Specifically, clients are sampled uniformly at random, choosing $M$ clients without replacement in a given round, but with replacement across rounds. For CIFAR-10, we sample all 10 clients at each round, simulating cross-silo FL. 

In our simulations, we vary the client learning rate, server learning rate, and number of training epochs per client. For more details on the algorithms, models, hyperparameters, and implementations, see Appendix \ref{appendix:empirical_details}. We use only one set of experimental setting (dataset and model) to illustrate the general guidelines in \cref{subsec:evaluation_guide}. More results on other experimental settings can be found in \cref{appendix:additional_results}. The code to reproduce all the experimental results is open-sourced\footnote{\url{https://github.com/google-research/federated/tree/aa2df1c7f513584532cac9c939ddd94f434ed430/fedopt_guide}}.

\subsection{Suggestions for Evaluations} \label{subsec:evaluation_guide}
This section contains advice for researchers and practitioners on how to evaluate and compare federated optimization algorithms. We stress that our recommendations are neither comprehensive nor always necessary. Rather, we hope they will inspire the reader to think critically about empirical studies of federated optimization algorithms, especially aspects that might differ from empirical evaluations of centralized optimization algorithms. We hope this can be useful in designing new empirical studies, as well as discerning the merits of past and future empirical work on federated optimization.


\subsubsection{Use Realistic Tuning Strategies} \label{subsec:tuning strategy}

While centralized learning problems often have explicit training/validation/test splits, this may not be the case in federated learning problems. In order to perform effective tuning for the purposes of simulations, one may wish to use some form of held-out validation data. Notably, such held-out data can be formed in a variety of ways in federated learning.

One simple approach to creating held-out data is to use a held-out set of clients for evaluation. These clients will not participate in any training round. This approach is generally amenable to cross-device settings, where there may be more clients than can reasonably be communicated with during the entire training procedure, and the goal is often to train a model that generalizes to clients not seen during training. In fact, if the number of potential training clients is sufficiently large, it may be statistically valid to simply validate across a random set of training clients, as this validation set will have little overlap with the set of training clients. By contrast, this approach may not be suitable in cross-silo settings, where we have a small set of clients and close to full participation per training round.

Another strategy for generating held-out data that is suitable in cross-silo settings would be to reserve some fraction of the data on each client for validation. This approach has the advantage of ensuring that the validation set is representative, but requires each client to have enough data to effectively partition between train and validation sets. Thus, this approach may not be advisable in cross-device settings where clients have limited amounts of data.

We stress that while the approaches above can be effective for simulations and production use-cases, held-out set tuning may not be the ideal way to perform federated hyperparameter tuning. For example, system constraints (such as limited client availability) may make it difficult to re-run an algorithm with many different hyperparameter settings. Thus, it may be beneficial to instead learn hyperparameters in tandem with the model. Such an approach was used by \citet{khodak2021federated}, who found that hyperparameters can be efficiently learned across clients using techniques based on weight-sharing for neural architecture search~\citep{pham2018efficient}. Developing effective federated hyperparameter tuning techniques is an important, but relatively open problem. 

Many previous works in federated optimization (including the original paper on \fedavg~\citep{mcmahan17fedavg}) tune their methods on a held-out centralized test set. This can be useful when trying to understand the fundamental limits of federated learning compared to centralized learning. This ``test-set tuning'' is not usually a practical method, but instead can be viewed as addressing questions like ``\emph{In an ideal scenario, can federated optimization methods recover the same accuracy as centralized optimization methods?}''.

\begin{figure}[ht]
\centering
\begin{subfigure}
    \centering
    \includegraphics[width=0.45\linewidth]{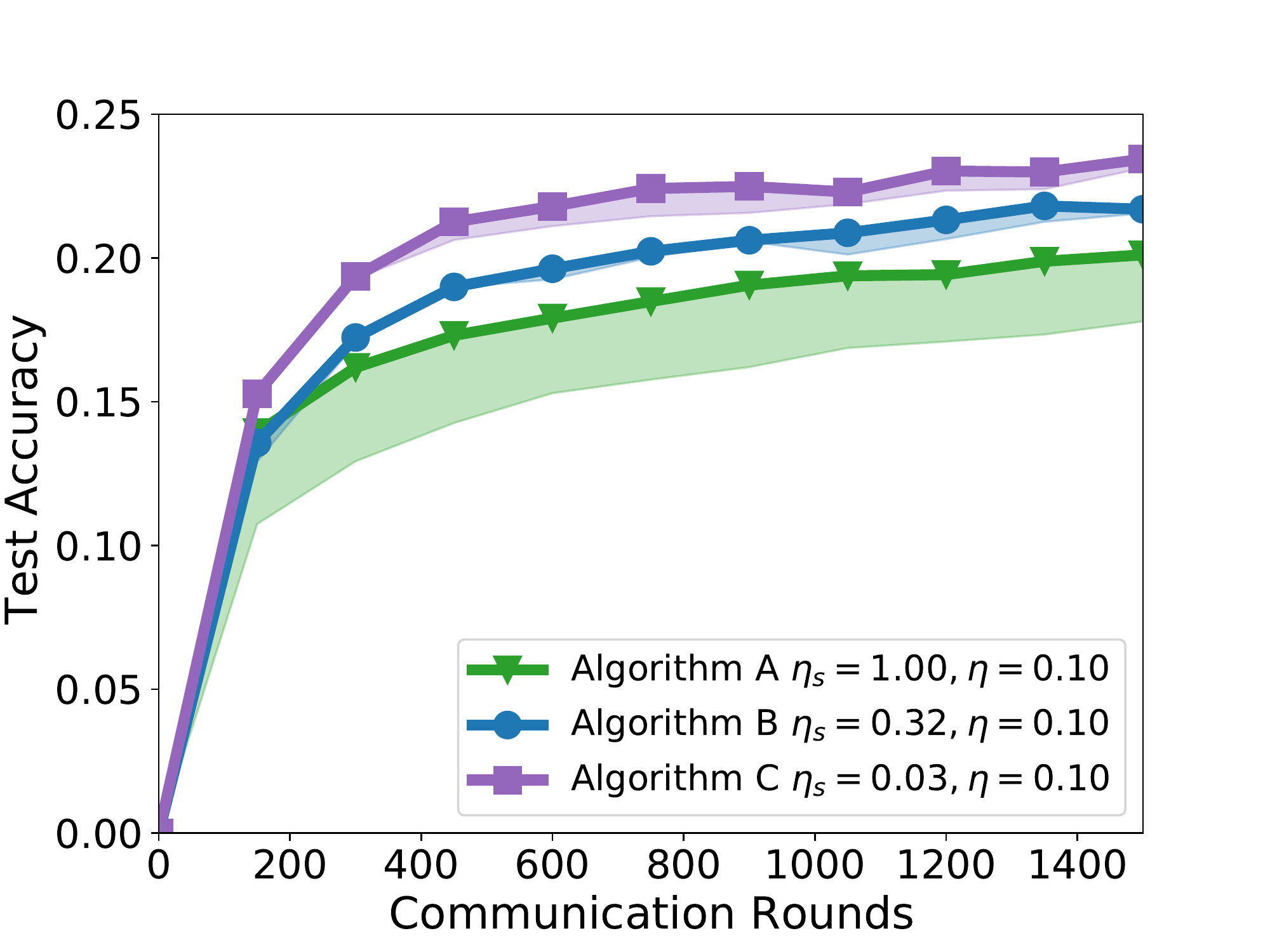}
\end{subfigure}
\caption{Test accuracy on Stack Overflow for Algorithms A, B, and C. The best and second-best performing client and server learning rate combinations are selected for each algorithm based on validation set accuracy (see Figure~\ref{fig:stackoverflow_tune_lr}). The dark lines indicate the test set accuracy for the best-performing combination (given in the legend), with shading indicating the gap between the best and second-best performing runs. This shaded gap provides an indication of the relatively sensitivity of the different algorithms to the resolution of the hyperparameter tuning grid; here we see that Algorithms B and C both perform better and may be easier to tune.}
\label{fig:stackoverflow_compare_tuning}
\end{figure}

\paragraph{Example}

To exemplify one of the tuning strategies discussed above, we give an explicit example on the Stack Overflow task. This dataset partitions clients into explicit train, validation, and test sets. We compare Algorithms A, B, and C on this dataset, tuning the client and server learning rates by selecting the values maximizing the accuracy of the model on the validation set (see Figure~\ref{fig:stackoverflow_tune_lr}). To give a sense for how robust the algorithms are to this tuning, we also determine the second-best configuration of learning rates. The test accuracy for these configurations is given in Figure \ref{fig:stackoverflow_compare_tuning}.

\subsubsection{Tune Client and Server Learning Rates}\label{section:tune_lrs}
The generalized \fedavg formulation provided in \cref{algo:generalized_fedavg} introduces the notion of separate learning rates for the client gradient updates and the server update. Learning rate tuning is important both for finding optimal operating points of a new algorithm, and for generating fair comparisons between algorithms. In particular, a server learning rate of 1.0 is not always optimal, even for \fedavg~\citep{reddi2021adaptive}. Thus, we strongly recommend tuning both learning rates.

Furthermore, because client and server learning rates can be interdependent in some algorithm designs, these should generally be tuned \emph{simultaneously}. This is a key point: to minimize the number of parameter combinations tested, one may wish to tune them serially, perhaps by fixing one, then tuning the other. This can potentially (but not always) lead to sub-optimal operating points.

\begin{figure}[ht]
\centering
\begin{subfigure}
    \centering
    \includegraphics[width=0.3\linewidth]{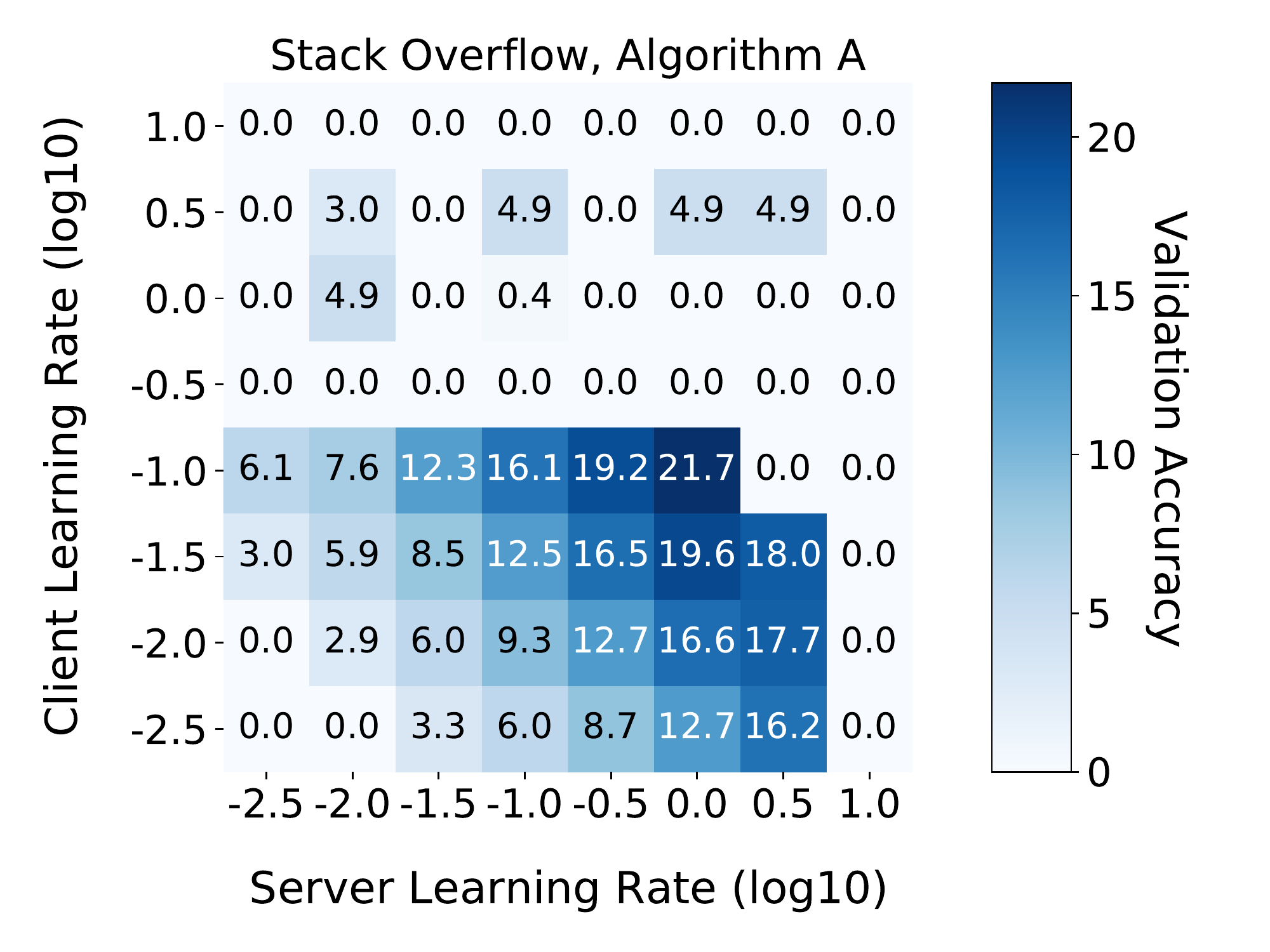}
\end{subfigure}
\begin{subfigure}
    \centering
    \includegraphics[width=0.3\linewidth]{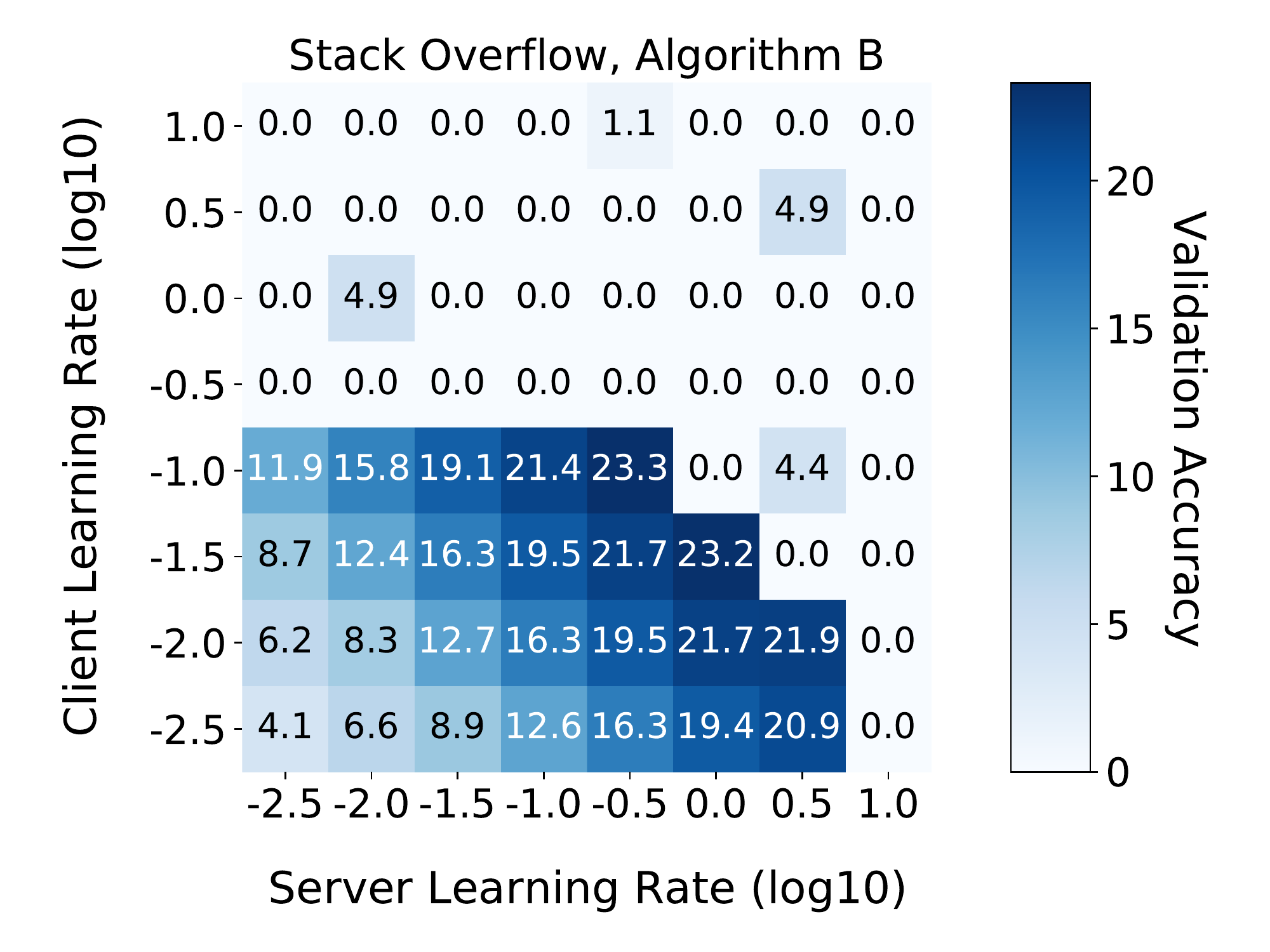}
\end{subfigure}
\begin{subfigure}
    \centering
    \includegraphics[width=0.3\linewidth]{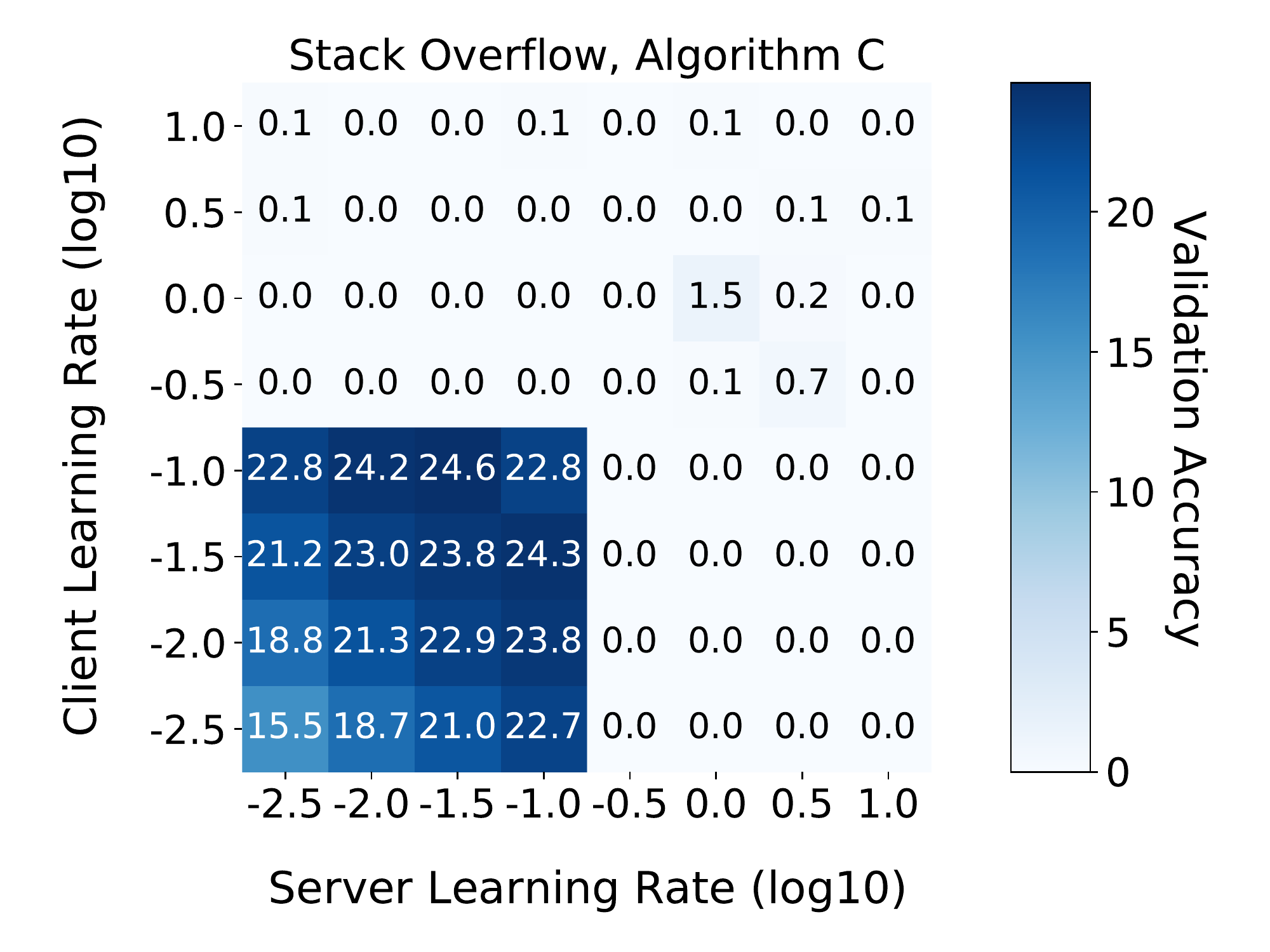}
\end{subfigure}
\caption{Validation accuracy on Stack Overflow at the last training round (2000), for various client and server learning rates. Results for Algorithms A, B, and C are given in the left, middle, and right plots, respectively. The best and second-best $(\eta_s, \eta)$ combinations are used in Figure~\ref{fig:stackoverflow_compare_tuning}.}
\label{fig:stackoverflow_tune_lr}
\end{figure}

For example, in \cref{fig:stackoverflow_tune_lr}, we plot the test accuracy for Algorithms A, B, and C on Stack Overflow as a function of both client and server learning rate. This gives us a full picture of the best possible results achievable by the algorithms with various hyperparameter settings. For Algorithm A, there is only one combination that obtains over 20\% accuracy. For Algorithm B, there are multiple configurations that achieve over 20\%. However, the client and server learning rates appear to be inter-dependent, and must be set inversely proportional to one another in a specific manner.
Thus, algorithms A and B may require tuning client and server learning rates \emph{simultaneously} in order to find near-optimal operating points. At the very least, the tuning must account for the ``staircase'' shape of the tuning grid.

By contrast, we see that for Algorithm C, 20\% accuracy is obtained for all points in our grid where either the client learning rate or server learning rate is $0.1$ (as long as the other is at most $0.1$). Informally, the rectangular shape of the tuning grid suggests that for Algorithm C, we can first tune the server learning rate, then the client learning rate (or vice-versa). This avoids a quadratic blow-up in the amount of hyperparameter tuning needed.

\subsubsection{Analyze Communication-Limited Performance}

\begin{figure}[ht]
\centering
\begin{subfigure}
    \centering
    \includegraphics[width=0.3\linewidth]{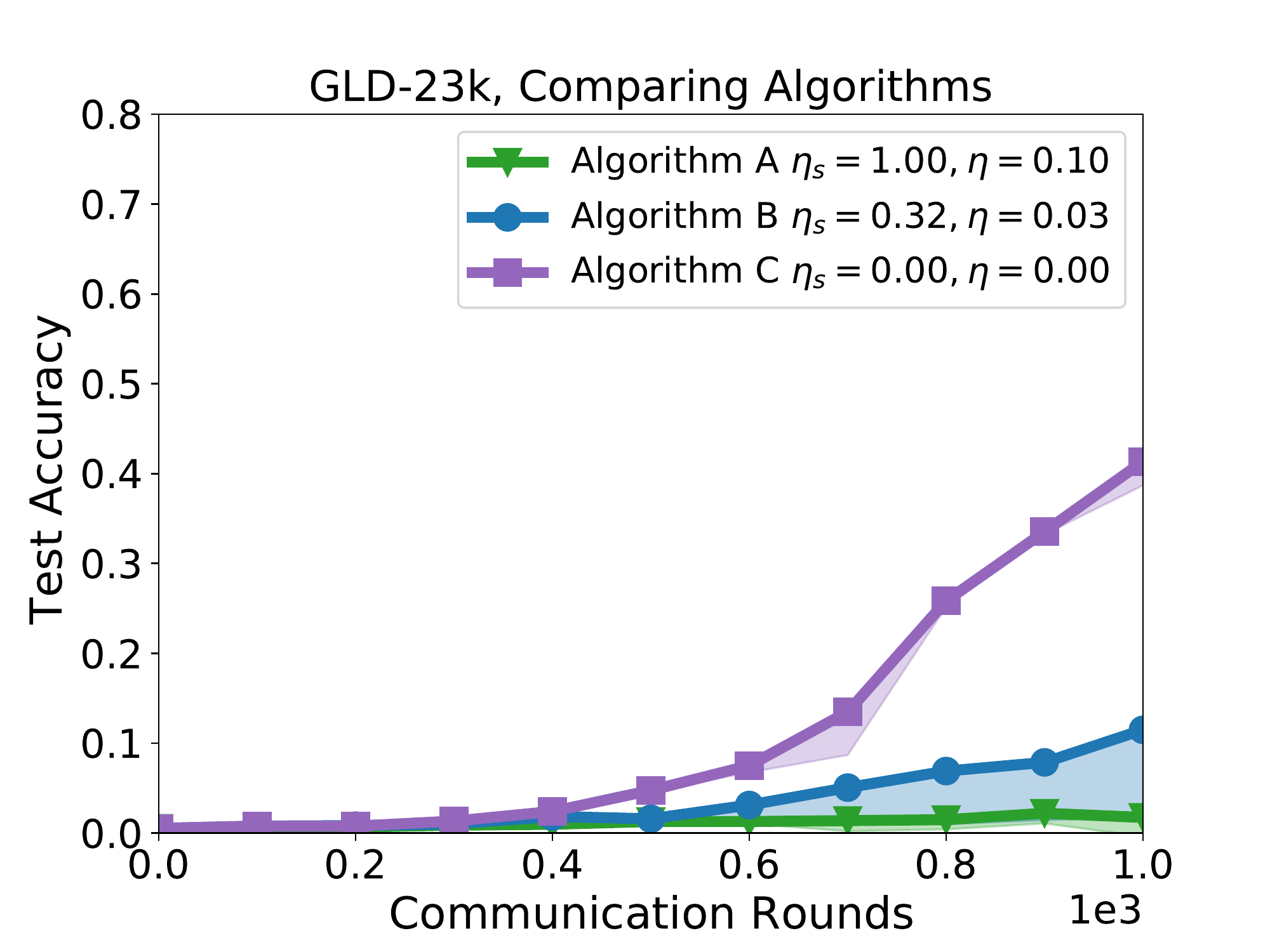}
\end{subfigure}
\begin{subfigure}
    \centering
    \includegraphics[width=0.3\linewidth]{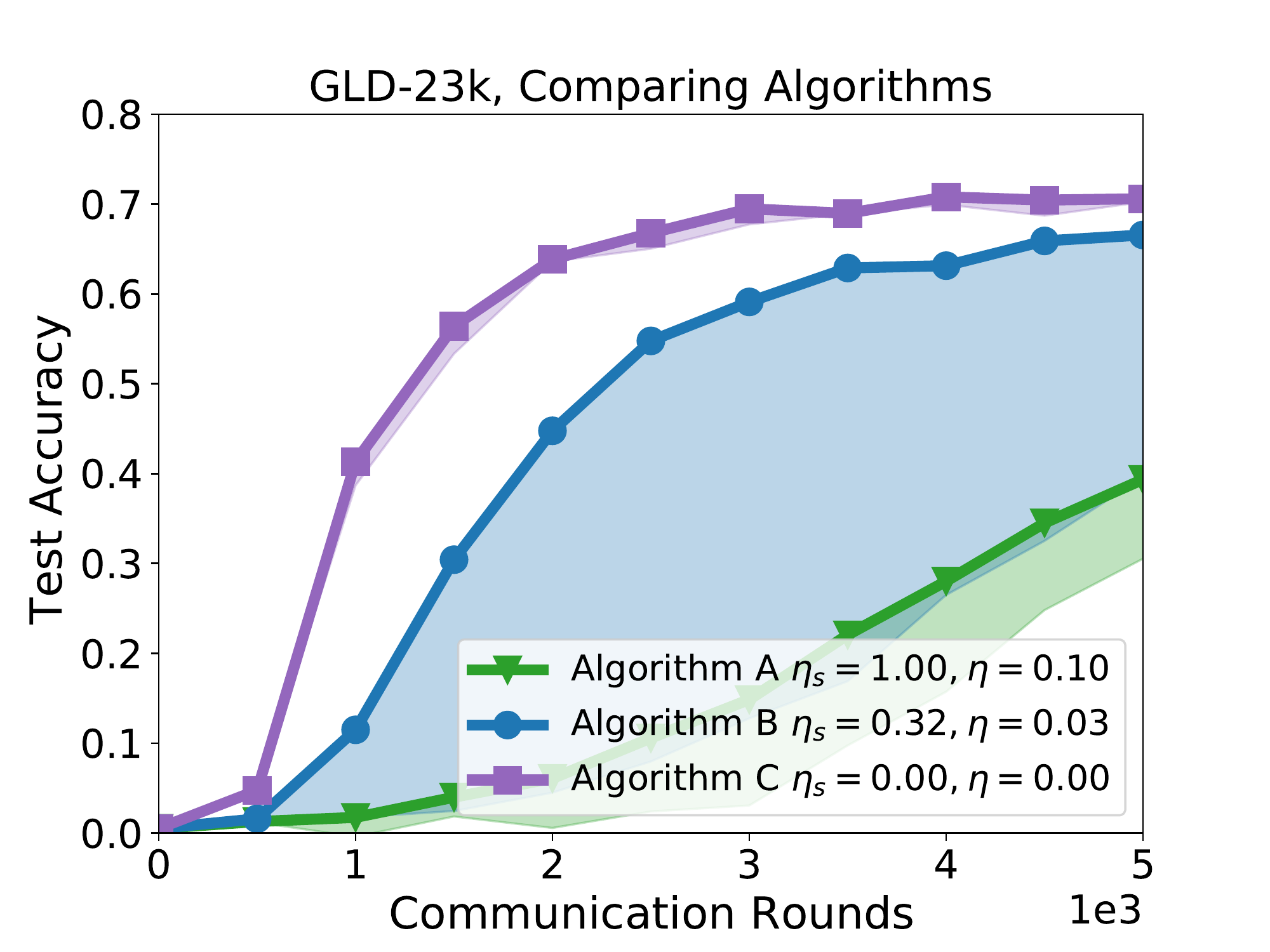}
\end{subfigure}
\begin{subfigure}
    \centering
    \includegraphics[width=0.3\linewidth]{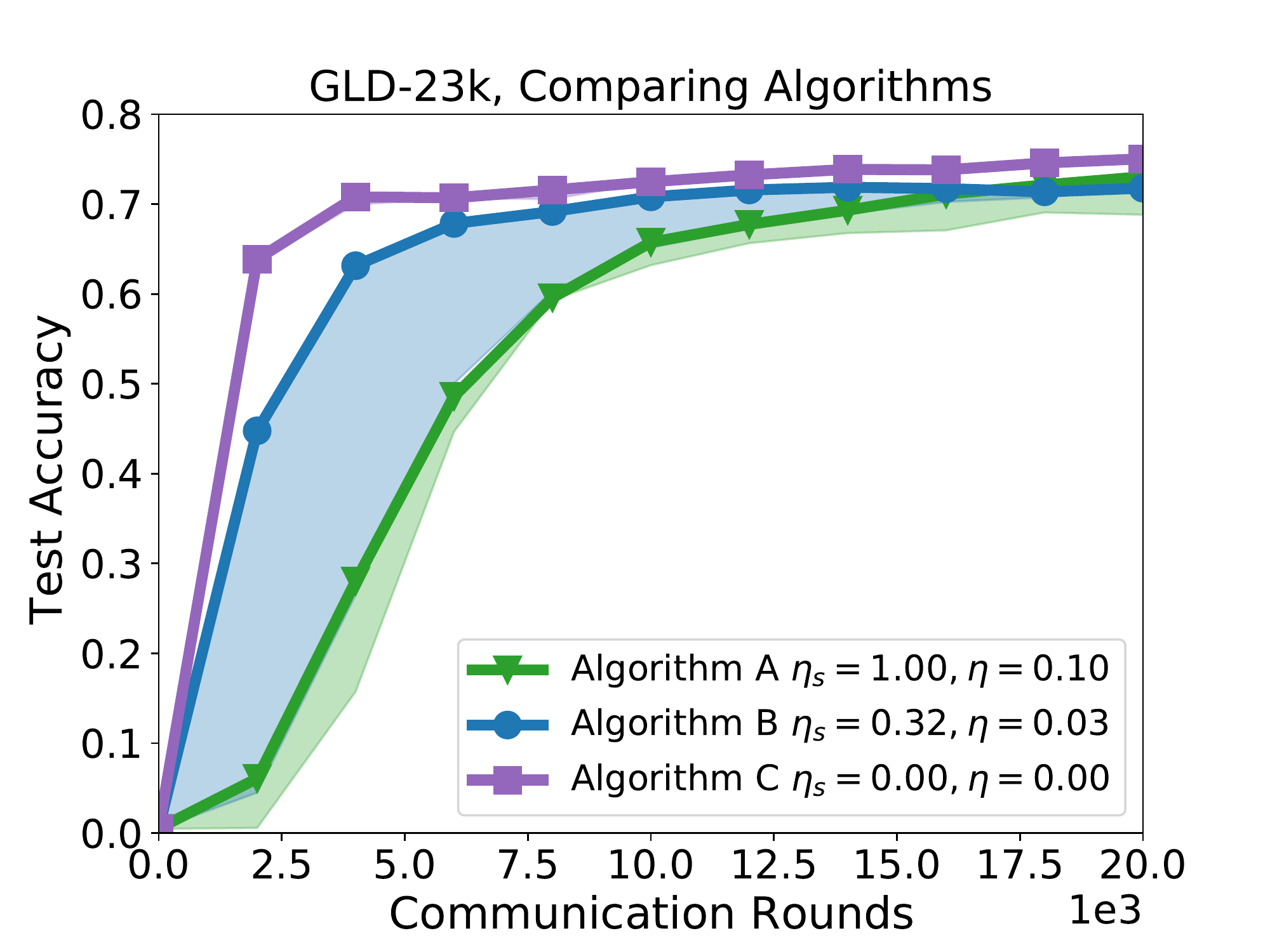}
\end{subfigure}
\caption{Test accuracy on GLD-23k for a total of 1000, 5000, and 20,000 communication rounds (left, middle, and right, respectively).  The best and second-best performing client and server learning rate combinations are selected for each algorithm based on the test set accuracy after the final communication round. The dark lines indicate the test set accuracy for the best-performing combination (given in the legend), with shading indicating the gap between the best and second-best performing runs.}
\label{fig:gld23k_compare_asymptotics}
\end{figure}

In optimization theory, convergence rates are often asymptotic. Even in empirical evaluations, many methods are evaluated on their performance after sufficiently many training rounds. Given that communication is a fundamental bottleneck in practical federated learning, we encourage authors to instead focus on algorithmic performance in communication-limited settings, especially for cross-device applications. We stress that the relative performance of various algorithms may change depending on the number of communication rounds performed.

For example, in \cref{fig:gld23k_compare_asymptotics}, we plot Algorithms A, B, and C for various regimes of communication rounds on the GLD-23k task. Depending on the total number of communication rounds, the conclusions one might draw are different. In the left plot, we might surmise that Algorithms A and B make little progress, while algorithm C is much better. In the middle plot, we might assert that Algorithm A is worse than B and C, and that C is better than B. In the right plot, we see that all algorithms perform equivalently after sufficiently many rounds (as is often required for theoretical results), with B and C achieving slightly better initial accuracy. Notably, Algorithms A and C have little difference between the best and second best learning rate configurations in the right-hand plot, unlike Algorithm B. In short, the actual desired number of communication rounds impacts the conclusions that one might draw about an algorithm. We encourage researchers to be up front about the success and limitations of their algorithms as the amount of communication changes.

\subsubsection{Treat Local Training Steps as a Hyperparameter}\label{section:local_steps_hparam}
According to \citet{mcmahan17fedavg}, one of the key advantages of using \fedavg over \fedsgd is improved communication-efficiency. This is achieved by having clients perform multiple training steps during each communication round (as opposed to \fedsgd, in which clients effectively perform a single training step). Intuitively, larger numbers of local training steps ($\tau_i$ in \cref{algo:generalized_fedavg}) will lead to a reduction in the total number of communication rounds needed to converge. However, as shown repeatedly in the literature on federated optimization, larger numbers of local training steps can also reduce the quality of the learned model~\citep{mcmahan17fedavg, li2018federated, charles2020learningrates, charles2021convergence}. Thus, this strategy often requires appropriate setting of the number of local steps. To obtain the best trade-off between convergence and accuracy, it is beneficial to rigorously tune this hyperparameter.

\begin{figure}[ht]
\centering
\begin{subfigure}
    \centering
    \includegraphics[width=0.45\linewidth]{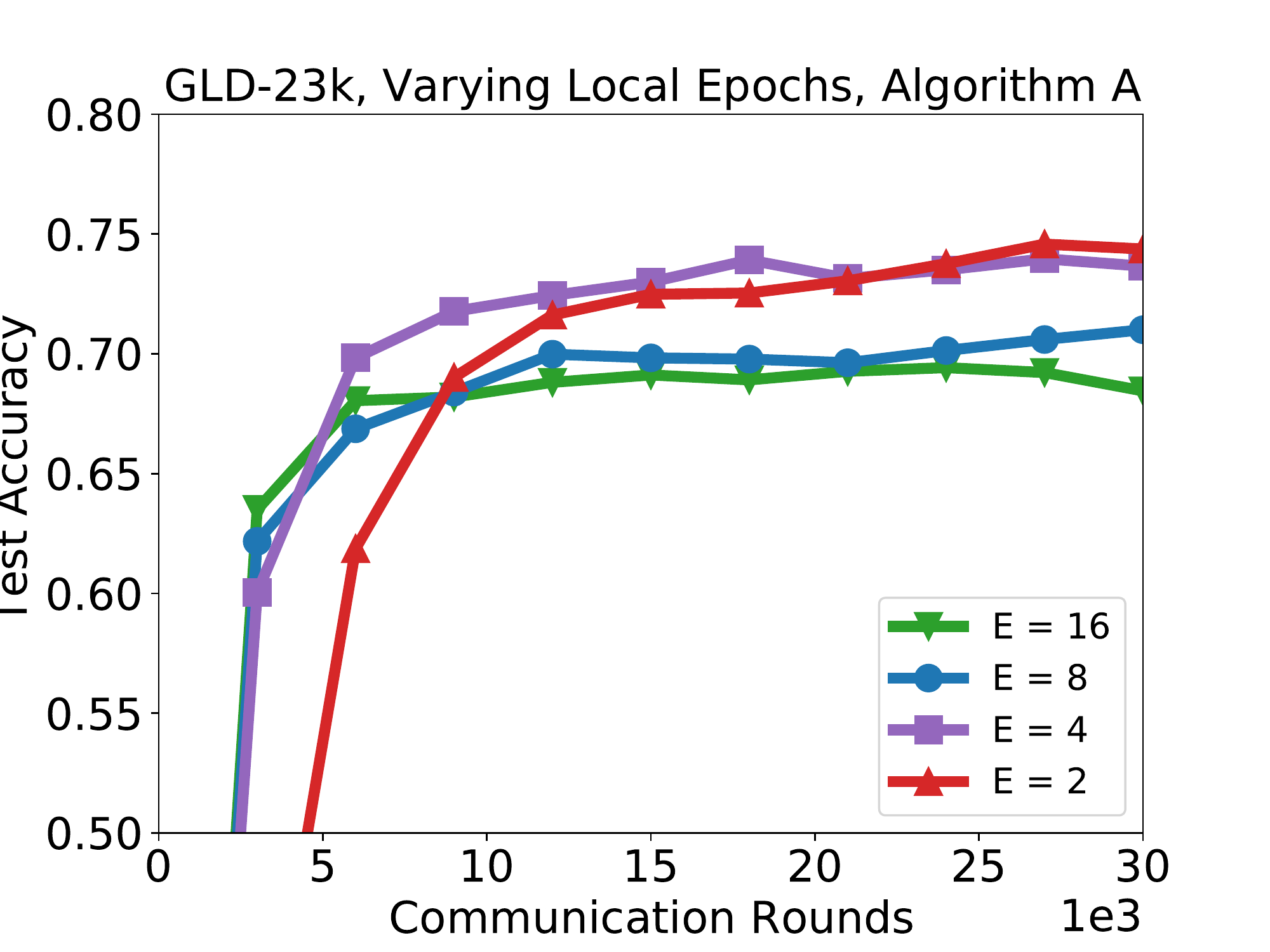}
\end{subfigure}
\begin{subfigure}
    \centering
    \includegraphics[width=0.45\linewidth]{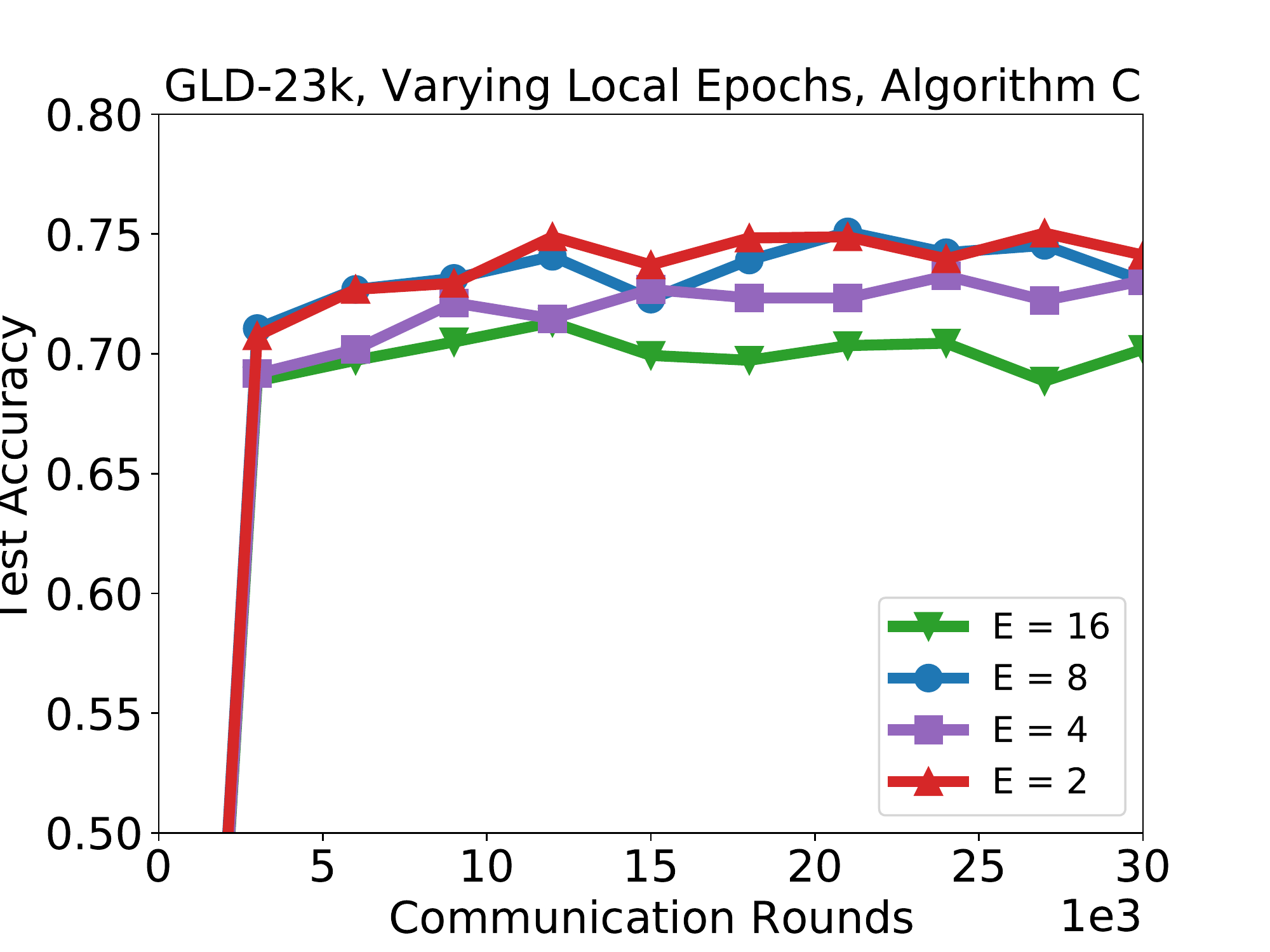}
\end{subfigure}
\caption{Test accuracy on GLD-23k for Algorithms A (left) and C (right) for various numbers of local epochs per round $E$, versus the number of communication rounds. We set $\eta = 0.1, \eta_s = 1.0$ for Algorithm A, and $\eta = 0.01$, $\eta_s = 10^{-5/2}$ for Algorithm C.}
\label{fig:gld23k_compare_num_epochs_num_rounds}
\end{figure}

\begin{figure}[ht]
\centering
\begin{subfigure}
    \centering
    \includegraphics[width=0.45\linewidth]{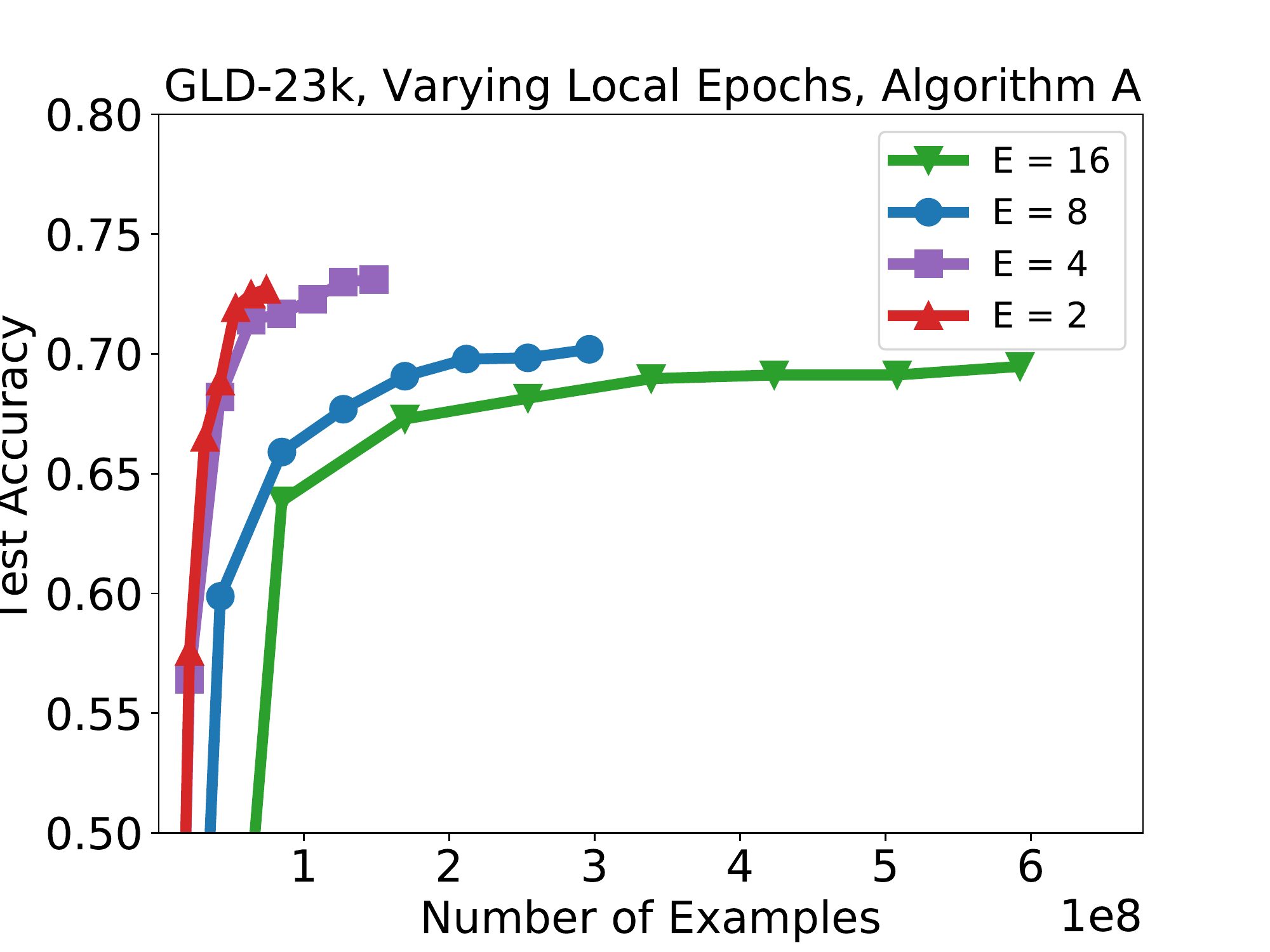}
\end{subfigure}
\begin{subfigure}
    \centering
    \includegraphics[width=0.45\linewidth]{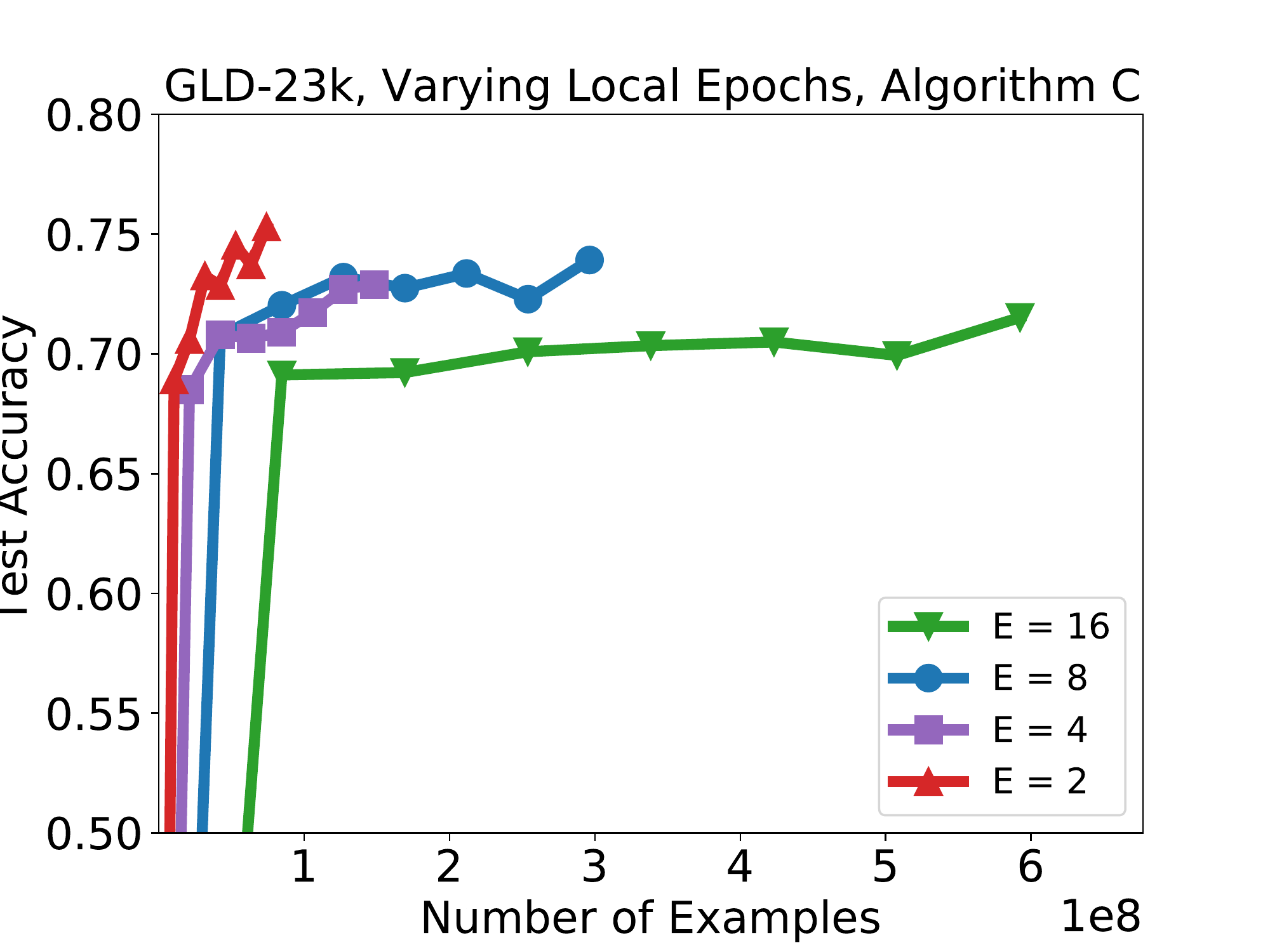}
\end{subfigure}
\caption{Test accuracy on GLD-23k for Algorithms A (left) and C (right) for various numbers of local epochs per round, versus the total number of examples processed by all clients. We set $\eta = 0.1, \eta_s = 1.0$ for Algorithm A, and $\eta = 0.01$, $\eta_s = 10^{-5/2}$ for Algorithm C. These learning rates were chosen as they performed well for $E = 2$. While they also performed near-optimally for other values of $E$, tuning the learning rates jointly with $E$ may produce slightly better results.}
\label{fig:gld23k_compare_num_epochs_num_examples}
\end{figure}

For example, in \cref{fig:gld23k_compare_num_epochs_num_rounds}, we apply Algorithms A and C with different number of local epochs $E$ on the GLD-23k task. We see that for Algorithm A, the choice of number of local epochs can make a large impact on the initial accuracy of the model, and minor differences in later rounds. On the other hand, Algorithm C is robust against the choice of $E$, achieving comparable accuracy for all settings used throughout.

While \cref{fig:gld23k_compare_num_epochs_num_rounds} is useful for judging the number of communication rounds needed, it omits the amount of \emph{total} computation being performed by the clients. For each client, $E = 16$ involves $8\times$ as much work as $E = 2$, something that is particularly important in cross-device settings where clients may have limited compute capacity. To judge this overall computation complexity, we give an alternate view of these results in \cref{fig:gld23k_compare_num_epochs_num_examples}. Instead of plotting the number of communication rounds, we plot the total number of examples processed by all clients throughout the training process. We see that this gives a drastically different view of the algorithms and the role of $E$. We see that for both algorithms, $E = 16$ actually processes many more examples than $E = 2$ to reach a comparable accuracy. Thus, benefits of increasing $E$ saturate, and even decrease. However, we see that Algorithm C is slightly more robust to settings of $E$, achieving roughly the same ``example-efficiency'' for $E = 2, 4, 8$, while Algorithm A only sees comparable accuracy for $E = 2, 4$.

\begin{figure}[ht]
\centering
\begin{subfigure}
    \centering
    \includegraphics[width=0.45\linewidth]{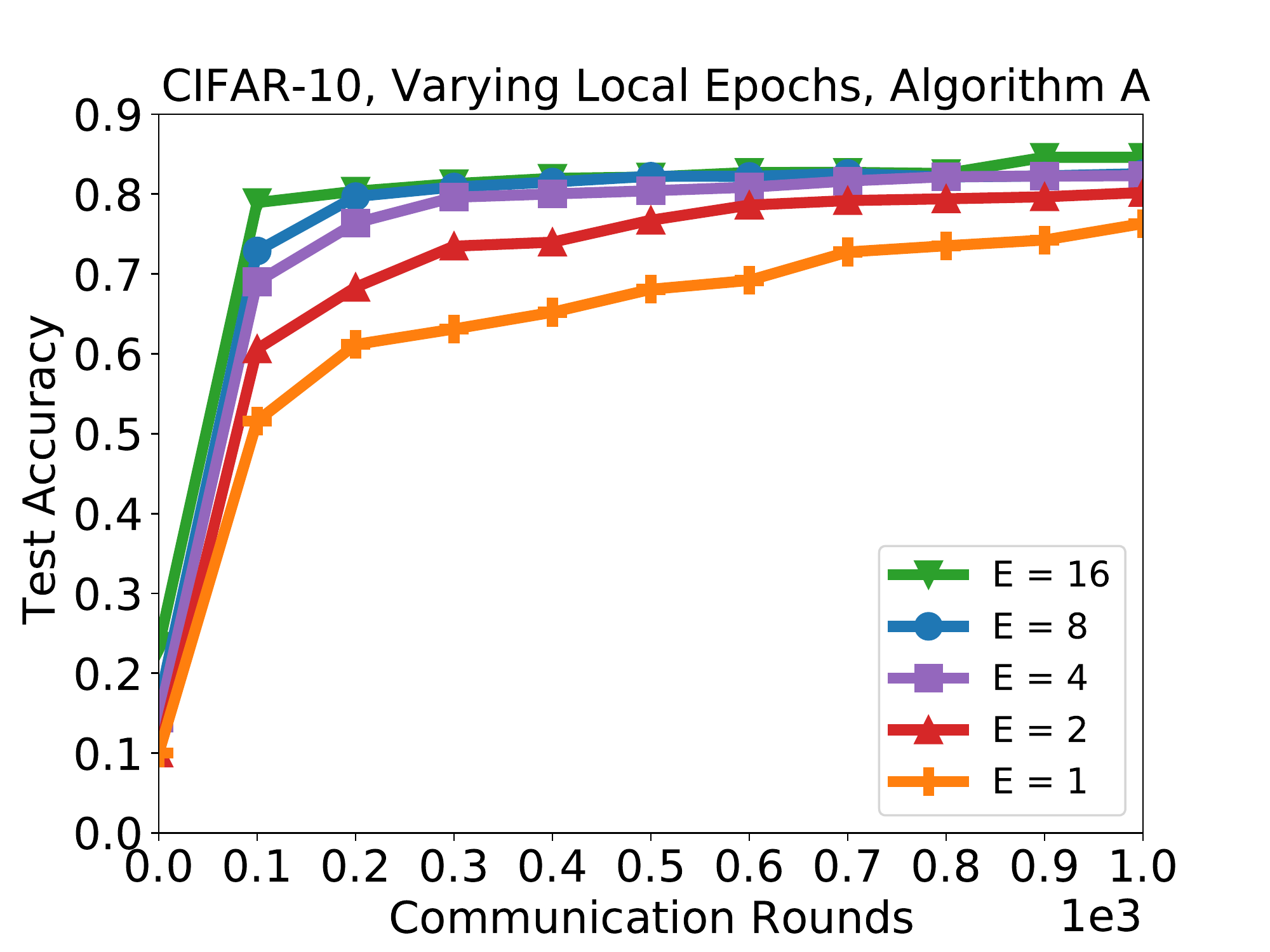}
\end{subfigure}
\begin{subfigure}
    \centering
    \includegraphics[width=0.45\linewidth]{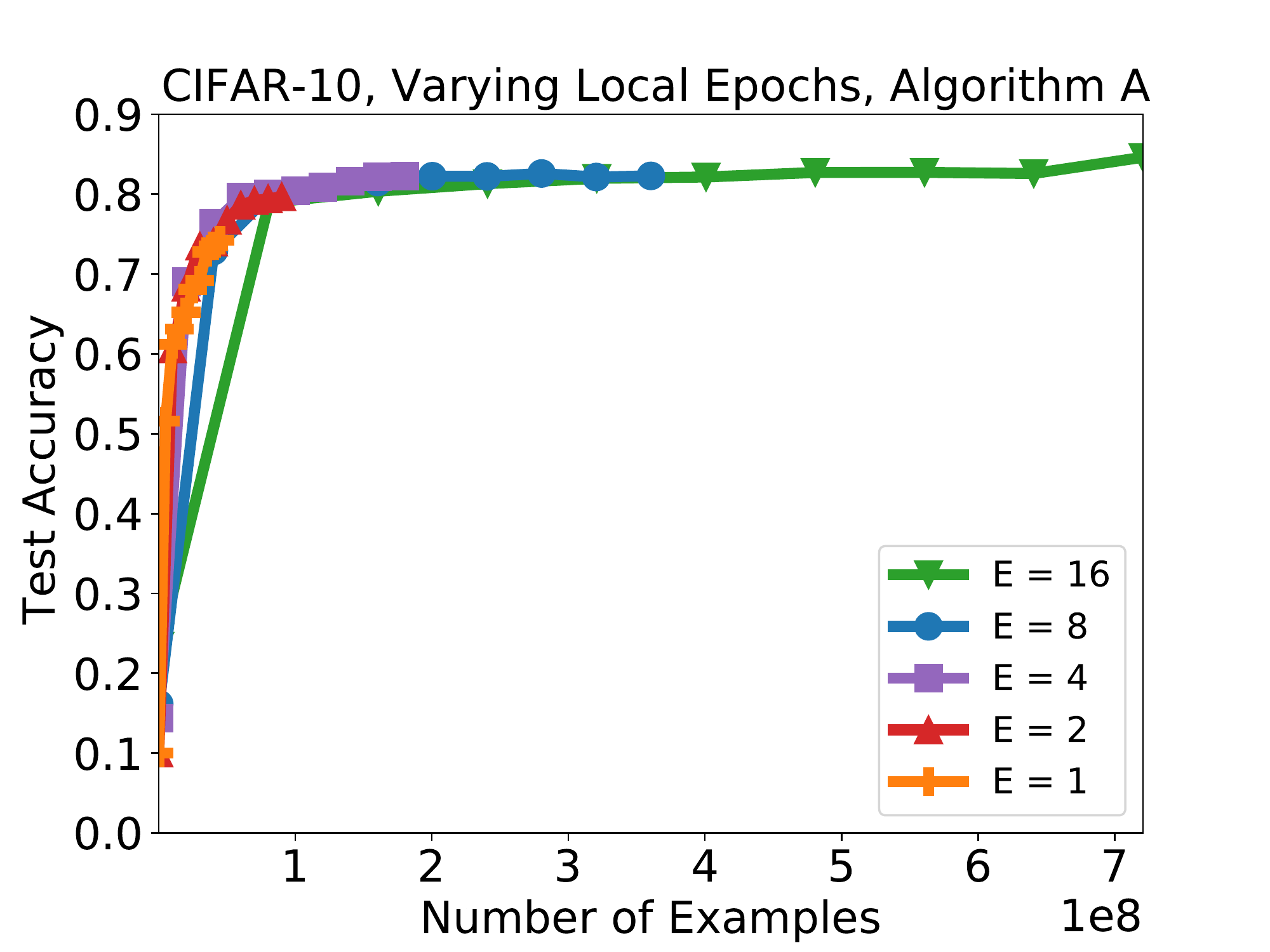}
\end{subfigure}
\caption{Test accuracy on CIFAR-10 for Algorithms A. We plot test accuracy versus the number of communication rounds (left) and versus the total number of examples processed by clients (right) for various numbers of local epochs per round $E$. We use learning rates $\eta = 0.01, \eta_s = 1.0$.  These learning rates were chosen as they performed well for $E = 2$. While they also performed near-optimally for other values of $E$, tuning the learning rates jointly with $E$ may produce slightly better results.}
\label{fig:cifar10_compare_num_epochs}
\end{figure}

We also note that such findings may depend on the federated setting being simulated. For example, we perform the same experiment as above with Algorithm A, but for CIFAR-10. The results are in Figure \ref{fig:cifar10_compare_num_epochs}. While we see similar results to the left-hand plots of Figures \ref{fig:gld23k_compare_num_epochs_num_rounds} and \ref{fig:gld23k_compare_num_epochs_num_examples}, the plots are slightly different when comparing accuracy versus number of examples seen. In particular, while Algorithm A can obtain higher accuracies with fewer examples when using a smaller value of $E$ on GLD-23k, we see roughly comparable numbers of examples processed for any test accuracy on CIFAR-10. In other words, Algorithm A experiences diminishing returns (in terms of data efficiency) on GLD-23k. It does not see such diminishing returns on CIFAR-10. This may be due to the difference in nature of the two datasets. GLD-23k is a more heterogeneous, cross-device dataset, and therefore setting $E$ to be too large can be detrimental~\citep{woodworth2020minibatch}, while CIFAR-10 is more homogeneous and cross-silo, and therefore may benefit from larger $E$~\citep{woodworth2020localSGD}.

\subsubsection{Understand the Impact of Cohort Size}

\begin{figure}[ht]
\centering
\begin{subfigure}
    \centering
    \includegraphics[width=0.45\linewidth]{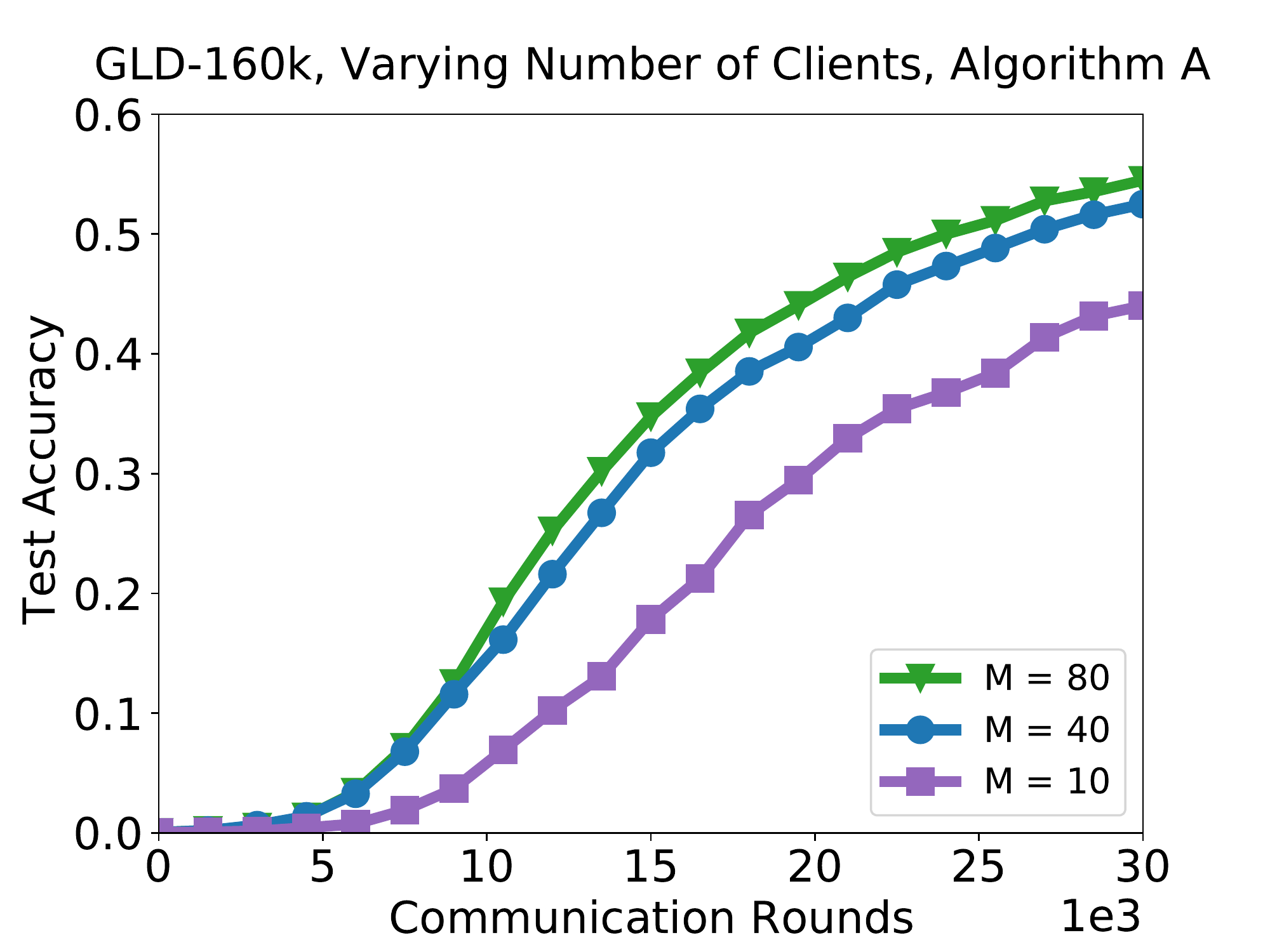}
\end{subfigure}
\begin{subfigure}
    \centering
    \includegraphics[width=0.45\linewidth]{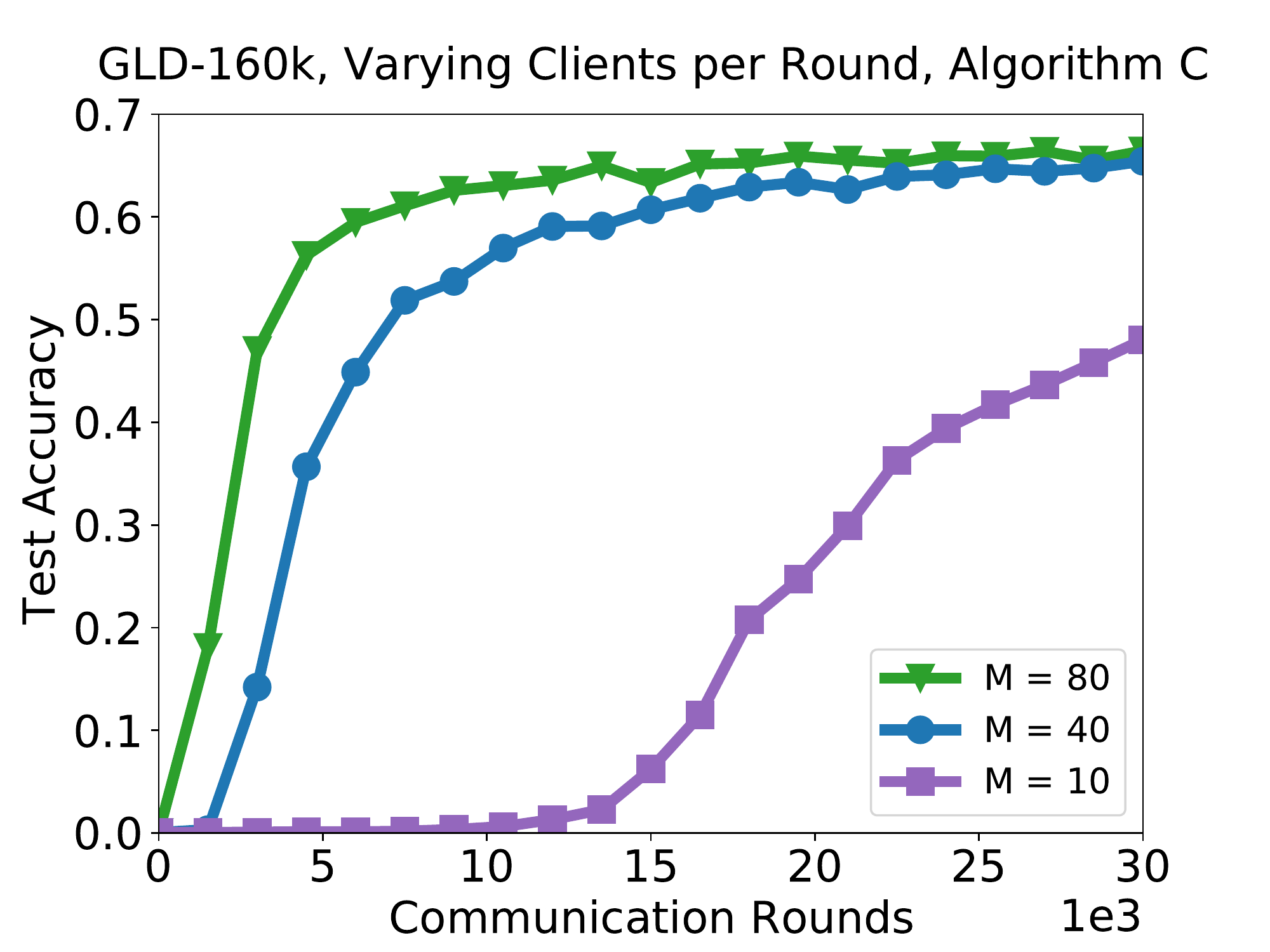}
\end{subfigure}
\caption{Test accuracy on GLD-160k for Algorithms A (left) and C (right) for various cohort size $M$, versus the number of communication rounds. We use learning rates $\eta = 0.1, \eta_s = 1$ for Algorithm A, and $\eta = 0.01, \eta_s = 10^{-5/2}$ for Algorithm C.  These learning rates were chosen as they performed well for $M = 10$. While they also performed near-optimally for other values of $M$, tuning the learning rates jointly with $M$ may produce slightly better results.}
\label{fig:gld160k_compare_num_clients_num_rounds}
\end{figure}

One fundamental aspect of (cross-device) federated optimization that we have not yet broached is the effect of the number of clients $M$ sampled at each round (i.e., the cohort size). Intuitively, the larger $M$ is, the less variance in the aggregate client updates. This in turn could lead to a reduction in the total number of training rounds needed to obtain a given accuracy. For example, in \cref{fig:gld160k_compare_num_clients_num_rounds}, we plot the accuracy of Algorithms A and C on GLD-160k, varying the cohort size $M$ over $\{10, 40, 80\}$. We see that for both algorithms, increasing $M$ leads to a better final accuracy. For Algorithm C, we also see a dramatic increase in the initial accuracy of the learned model.

\begin{figure}[ht]
\centering
\begin{subfigure}
    \centering
    \includegraphics[width=0.45\linewidth]{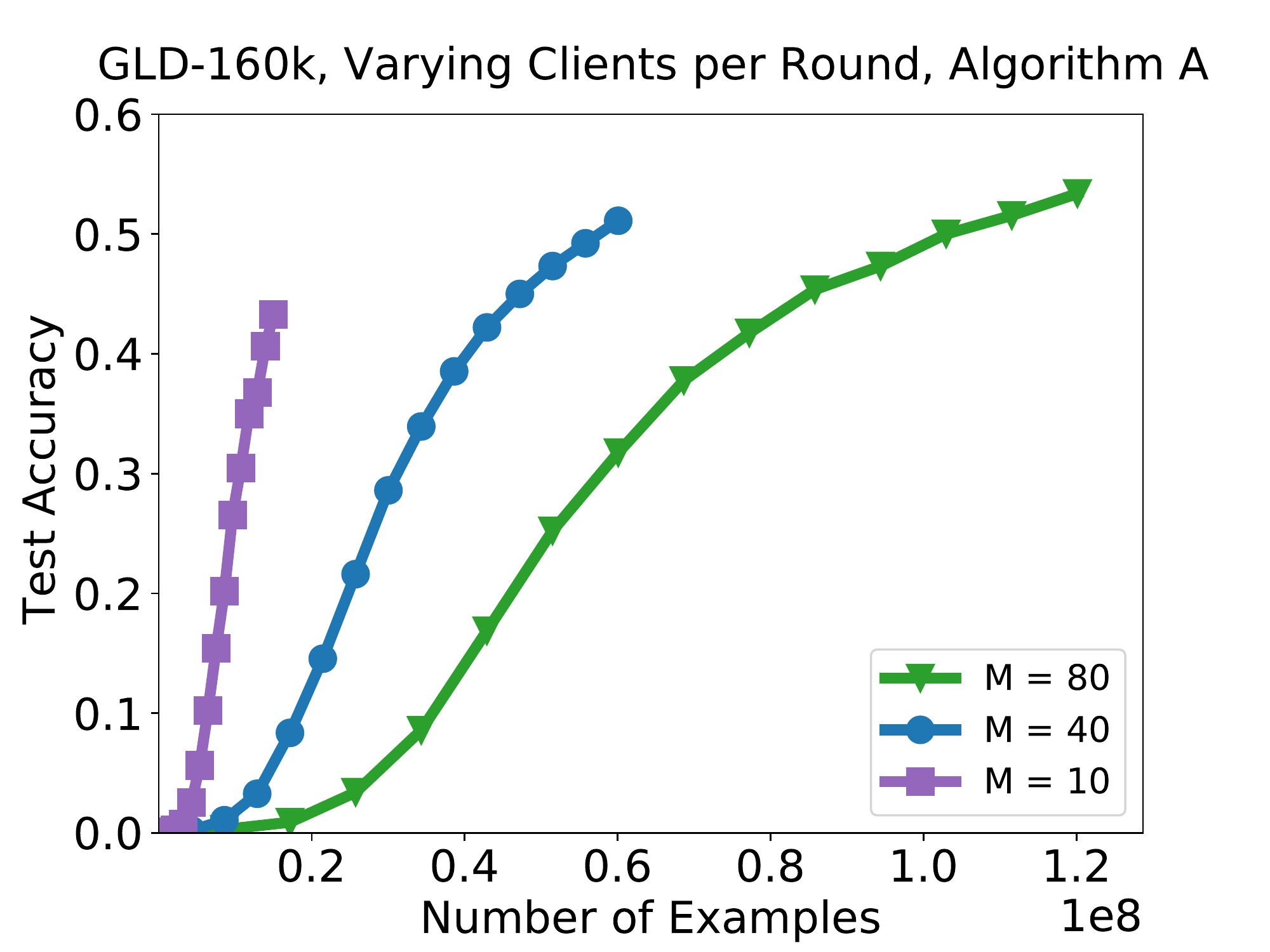}
\end{subfigure}
\begin{subfigure}
    \centering
    \includegraphics[width=0.45\linewidth]{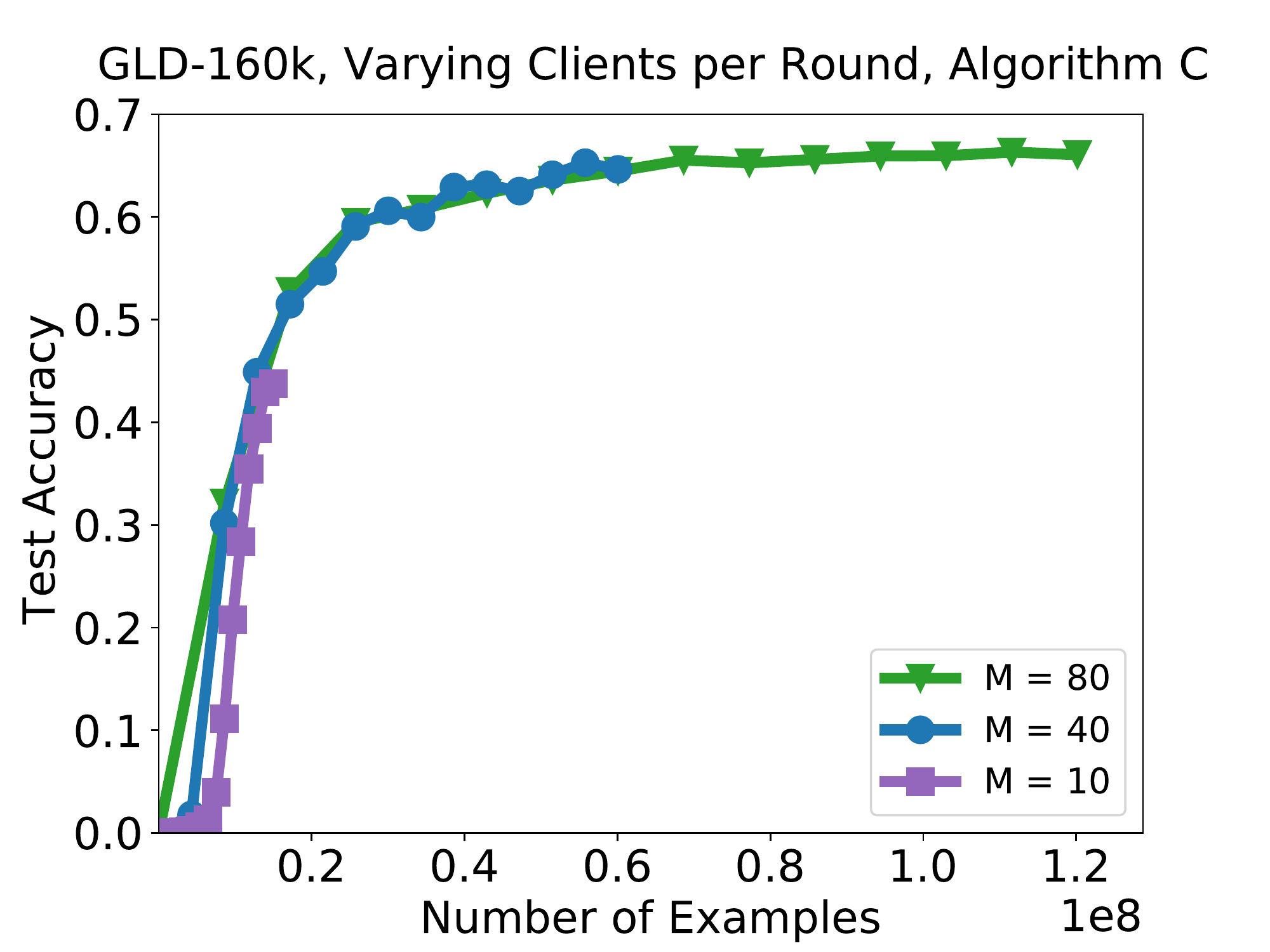}
\end{subfigure}
\caption{Test accuracy on GLD-160k for Algorithms A (left) and C (right) for various cohort sizes $M$, versus the number of examples processed by the clients. We use learning rates $\eta = 10^{-3/2}, \eta_s = 1$ for Algorithm A, and $\eta = \eta_s = 10^{-5/2}$ for Algorithm C.}
\label{fig:gld160k_compare_num_clients_num_examples}
\end{figure}

Just as in \cref{section:local_steps_hparam}, comparing these settings with respect to the number of communication rounds omits some key information. While all clients are now performing the same amount of computation per round, the total amount of client computation per round varies depending on $M$. Increased total client computation time can have negative impacts, both due to electricity usage and carbon emissions, as well as delayed round completion times stemming from straggler effects. In \cref{fig:gld160k_compare_num_clients_num_examples}, we give an alternative view of this experiment, plotting test accuracy versus the total number of examples processed at all clients. We see that the story changes dramatically for Algorithm A and C. Algorithm C seems to achieve comparable accuracy for a given number of examples, for all settings of $M$. For Algorithm A, higher values of $M$ actually require many more examples to be processed to reach smaller accuracy thresholds. This is an important point: The benefits and drawbacks of increasing $M$ can vary significantly depending on the federated optimization algorithm employed. More experimental results and discussion on how the number of clients sampled per round can affect federated training dynamics can be found in \citep{charles2021large}.

\subsubsection{Other Considerations}

\paragraph{Motivate simulation designs with well-specified target applications}
The guidelines in \cref{subsec:evaluation_guide} are generally applicable to both cross-device and cross-silo settings. While the example algorithms A, B, C were initially designed for cross-device federated learning, they are also applicable to cross-silo federated learning. However, as discussed in \cref{section:application_patterns}, cross-device and cross-silo settings have important distinctions. Echoing the suggestion in \cref{subsec:application_settings} for algorithm design, we encourage researchers to clearly specify the application setting of any evaluations. For example, when evaluating a method in cross-silo settings, it is not necessary for it to scale to millions of clients. By contrast, evaluating scalability of an algorithm may be vital when considering cross-device settings. While methods suitable for cross-device settings can often apply to cross-silo settings without significant modifications, the reverse is generally not true. More sophisticated cross-silo algorithms may require global information across clients, or require stateful clients (for example, SCAFFOLD~\citep{karimireddy2019scaffold}). Thus, it is important to be clear about what settings are being considered in any evaluation of federated optimization.

Further, it will usually not be realistic to explore all possible combinations of local epochs $E$, cohort sizes $M$, client and server learning rates, rounds of communication, etc. Thus, when decisions need to be made which limit the scope of experiments, it is best to make choices based on what is likely to be practical and realistic for the targeted application; comparing to the choices made in prior works can also be valuable, but if those regimes are not relevant to the target application, then such comparisons can be optional.

\paragraph{Using alternate evaluation metrics}

In the examples above, we report the accuracy of a model over all test samples (\ie the accuracy is the number of correct predictions over the entire test set, divided by the number of test samples). This ignores potential data imbalances across test clients. To perform better evaluations of federated algorithms, one may wish to use alternate evaluation metrics. For example, one could look at the mean test accuracy across all test clients. More generally, one could look at the distribution of a model's accuracy across clients. This is particularly important in areas such as fairness, where quantities such as the minimum accuracy across all clients play an important role (see \cref{sec:fairness} for more discussion on this topic). Developing better evaluation metrics for federated learning is an open problem, and we encourage researchers to clearly specify the metrics used for experimental evaluation.

Alternate metrics aggregations are particularly useful when trying to measure the fairness of a model. For example, one measure of fairness is the minimum accuracy across all clients. Of course, there are a number of other ways this could be examined, and we caution the reader that there is no single way of computing the fairness of a model (see work by \citet{li2019fair} and \citet{hu2020fedmgda+} for various approaches to measuring fairness in empirical evaluations). We encourage authors to be explicit about the measure they are using, and its potential drawbacks. For example, the minimum accuracy measure above guarantees a floor on client accuracy, but does not measure the variance of accuracy across clients (such as the difference between the maximum and minimum client accuracy). Developing useful fairness metrics is an ongoing and important area of study within federated learning.

\paragraph{Isolate sources of randomness} Empirical evaluations are typically stronger when they try to account for any sources of randomness (e.g., model initialization). Federated learning has notable sources of extra randomness. In cross-device settings, which clients are sampled at which round has the potential to significantly affect the learned model (particularly when $M$ is small), in much the same way that the order of examples seen while performing SGD (especially non-random orders) can alter the underlying training dynamics~\citep{eichner2019semi}. In order to derive fair comparisons, we recommend using pseudo-random client sampling in simulations to see how algorithms behave with the same sequence of sampled clients (see \citep{reddi2021adaptive} for an example of doing this in federated settings). More generally, we recommend varying the random seed governing the pseudo-random sampling and plotting the mean and variance of results across trials. More sophisticated empirical analyses may even perform biased sampling of clients in order to simulate things like client availability and heterogeneity in device capabilities~\citep{eichner2019semi}.

\paragraph{Use learning rate scheduling} One common technique used to train centralized models is learning rate scheduling, in particular warmup and decay~\citep{goyal2017accurate}. Such techniques can be vital for training large models such as transformers and deep ResNet models. Thus, learning rate scheduling may also be useful in federated learning. In fact, \citet{li2019convergence} show that some form of learning rate decay is \emph{necessary} to guarantee that \fedavg reaches critical points of the empirical loss function. How exactly to configure such scheduling is an open question. For example, in \cref{algo:generalized_fedavg}, learning rate scheduling can be done on the client or on the server (or both). If clients perform many local training steps, then client learning rate decay may be effective. However, if clients only perform a small number of updates relative to the number of communication rounds, server learning rate decay may be a more effective strategy.

\subsection{On the Role of Toy Problems and Artificial Datasets}

Due to the nascent and sometimes counter-intuitive findings of federated optimization, synthetic datasets can be useful for performing an initial validation of a method. Such datasets can be particularly useful for understanding federated optimization algorithms in controlled environments. For example, one might use such datasets to isolate and analyze the effect of single factor. Such datasets are often most useful as a way to gain insight over a single facet of federated learning, especially in service of subsequent theoretical analyses. Below we discuss some examples of toy datasets, their utility, and limits on their ability to provide meaningful simulations of federated learning.

\subsubsection{Quadratic and Convex Problems on Synthetic Data}
A frequently used class of toy problems are quadratic problems, \ie problems where the client objective functions $\obj_i(\vx)$ in~\eqref{eqn:global_obj} are quadratic functions. 
For instance, \citet{charles2020learningrates} consider
functions of the form 
$\obj_i(\vx) = \frac{1}{2} \big\| A_i^{1/2}(\vx - \vc_i) \big\|^2$ for $A_i \in \R^{d \times d}$ symmetric and positive semi-definite and $\vc_i \in \R^d$. Such formulations are useful for understanding the impact of heterogeneity among client datasets, as one can reduce the heterogeneity of the datasets to the heterogeneity of the pairs $(A_i, \vc_i)$. The quadratic nature also frequently allows for closed-form expressions of quantities of interest. For example, for the quadratic functions above, each client's optimal model is given by $\vx^\star_i := A_i^\dagger A_i \vc_i$ (where $A^\dagger$ denotes the pseudo-inverse of a matrix $A$). These optimal points for each client are in general different from the globally optimal solution, $\vx^\star = (\Exs_{i \sim \clientDist} A_i )^\dagger (\Exs_{i \sim \clientDist} A_i \vc_i)$.

Another approach to generating a synthetic quadratic problem was formulated by \citet{li2018federated} using heterogeneous Gaussian distributions. In their approach, each client $i$ has samples $x$ drawn from a Gaussian $\mathcal{N}(\mu_i, \Sigma)$. Each $x$ has label given by a generalized linear model $y = \sigma(W_ix + b_i)$. Notably, $\mu_i, W_i, b_i$ are all random variables. While the exact details of the distributions can vary, this toy problem allows one to control the heterogeneity of the $\mu_i$ (\ie the heterogeneity of the clients' examples) separately from the heterogeneity of the $w_i, b_i$ (\ie the heterogeneity of the clients' generalized linear models).

Thematically similar quadratic and convex toy problems have been used in various works on distributed optimization \citep{shamir2013dane,li2018federated,koloskova2020unified,wang2020tackling}. While frequently useful, especially from a theoretical point of view, there are a number of pitfalls to be cognizant of when using them. For example, quadratic functions are not a representative benchmark for balanced, IID, datasets due to their functional simplicity~\citep{woodworth2020localSGD}. More generally, convex problems are not representative benchmarks of non-convex problems, due to their inherent property that $F(\Exs_{i \sim \clientDist} \vx_i^\star) \leq \Exs_{i \sim \clientDist} F(\vx_i^\star)$, \ie the average of the clients' models over-performs. This property does not transfer to non-convex tasks.

\subsubsection{Artificial Partitioning of Centralized Datasets}

Another common way to create synthetic federated datasets is to take a centralized dataset, and employ some scheme to partition the dataset among a set of clients. Compared to the synthetic data approach above, this has the advantage of allowing one to use more realistic examples and well-understood models. One particularly prevalent approach is to take a labeled dataset (such as CIFAR-10) and partition examples by randomly distributing the set of all examples with label $\ell$ across some set of clients $S_\ell$. While this approach does result in a heterogeneous dataset, it is not necessarily indicative of realistic heterogeneous datasets. Thus, we caution the reader that such pathological partitioning should be done primarily for the purpose of understanding specific and extreme forms of heterogeneity. In the example of label partitioning above, this approach is most useful when trying to specifically understand the impact of label heterogeneity (\ie when each client may only have a small subset of the total set of labels) on a federated algorithm.

Recent work has attempted to partition centralized datasets using more realistic, but controllable partitioning schemes. Notably, \citet{hsu2019measuring} propose a useful partitioning scheme based on Latent Dirichlet Allocation from topic modeling. This has the advantage of allowing one to control the amount of heterogeneity by changing the parameter $\alpha$ of the underlying distribution. In fact, varying $\alpha$ in $[0, \infty)$ allows the user to interpolate between a completely IID partitioning of examples and the pathological label partitioning scheme above (wherein each client's dataset has a single label). Later work by \citet{reddi2021adaptive} used the Pachinko Allocation Method \citep{li2006pachinko} to partition datasets with a hierarchical label structure (such as CIFAR-100).

While these topic modelling approaches can generate more realistic artificial federated datasets, we caution that they are not necessarily emblematic of realistic federated datasets. For example, such methods assume that all clients generate data from the same stochastic process, which may not reflect datasets with dissimilar users. Instead, we argue that these topic modelling approaches are most useful when trying to understand the impact of heterogeneity on an algorithm. Because the underlying topics are generally controllable via some user-specified distribution, the user can evaluate an algorithm for various amounts of heterogeneity in order to perform effective ablation studies (see \citep{hsu2019measuring} for valuable examples of this).

\subsubsection{Datasets with Examples Shuffled Across Users}
As discussed in Section \ref{sec:federated_optimization}, a challenge for many federated algorithms is the presence of heterogeneous data across clients. Notably, this also poses a challenge for performing effective ablation studies on a federated optimization algorithm. Importantly, the effect of heterogeneous data across clients can be effectively removed by amalgamating all client datasets, and creating a new dataset where each client's dataset is drawn uniformly at random from this super-set of examples. More sophisticated ablation studies could also interpolate between this ``shuffled'', homogeneous dataset and the original dataset, for instance by having each client use a kind of mixture between their original dataset, and some global pool of examples.

While similar to standard server evaluation, this type of evaluation allows one to test novel federated algorithms in a simplified setting. This can be useful as a first step towards determining the plausibility of an idea, or as a way to measure the effect of client drift on an algorithm (eg. by evaluating an algorithm on both the original and shuffled dataset). However, such shuffled datasets are not indicative of the performance of federated optimization algorithms in practical settings. We therefore recommend such ``pathological'' datasets as ways to perform more fine-grained evaluation, not as a way to actually benchmark federated optimization methods.

\section{System Constraints and Practices}\label{sec:system}

In \cref{sec:practical_algorithm_design} we discussed methods to improve algorithm performance by considering heterogeneity, communication efficiency and many other factors; and in \cref{sec:evaluation} we evaluated algorithms in the absence of the deployment system. In this section, we will examine the dimensions of heterogeneity and other factors that appear in practice and discuss evaluating federated algorithms on real-world federated learning systems such as \citet{bonawitz19sysml,paulik2021federated}. We examine how the costs of communication, computation, and memory differ from the simulations used for developing algorithms, and what effect that has on real-world deployment. Then we will propose a basic model parameterized by system properties that can be used to evaluate algorithm deployment costs. We aim to show that co-designing numerical optimization and system optimization can achieve practical federated learning systems and applications. Finally, we will conclude with suggestions for deploying algorithms on real-world systems.

\subsection{Communication Costs} \label{sec:system_communication}

When developing and experimenting with federated learning algorithms, researchers and engineers typically use simulation environments. In simulated environments, communication can be effectively free, and typically always available, as the computation is often co-located either on a single machine or within a highly interconnected data center. This setup does not reflect real-world federated learning systems. In cross-device settings, compute power and communication capacity can exhibit extreme heterogeneity. Even in cross-silo settings, where compute resources may be less constrained, participants in a federated learning algorithm may be geographically separated, incurring large communication costs.

When we talk about communication-efficient algorithms in academia, we often measure using the frequency of communication, which is only part of the whole picture. In contrast, in real-world deployments, the elapsed time required to train a model is an important but often overlooked concern. It is important to also consider what factors contribute to where time is spent, as there are multiple dimensions to consider even within communication cost.

\paragraph{Redundancy in model updates} As mentioned in \cref{sec:practical_guidelines}, compressed communication can often help in federated learning. Communicated matrices may have many forms of redundancy (\eg sparsity, spatial), allowing for significant compression ratios. For instance, if the model updates are low rank, adopting low-rank/sketching-based compression methods~\citep{FEDLEARN2016,wang2018atomo,vogels2019powersgd} can achieve high communication efficiency. This may indicate a metric that understands the distribution of expected values, not only the shapes and sizes of the communications, is desirable.

\paragraph{Bandwidth heterogeneity and dynamics} Unlike distributed training in the data center setting, where network bandwidths are usually homogeneous and stable over time, in the cross-device setting edge devices typically communicate with a central server via potentially low quality network connections. In real-world deployments, the available bandwidth of wireless networks may have vastly different download versus upload capabilities. Furthermore, wireless networks can be noisy, are prone to interference, and vary in quality over time.

\paragraph{Stragglers} In synchronous learning algorithms with fixed cohorts (such as \fedavg as described in \cref{sec:basics}), communication cost of the system can be dominated by the slowest participant, as round completion is blocked by stragglers \citep{bonawitz19sysml}. Further, it is not always the case that the weakest compute node is also the weakest network link: communication-computation ratios can be highly imbalanced within a round. To mitigate the effect of extreme stragglers both \citet{bonawitz19sysml} and \citet{paulik2021federated} describe participant reporting deadlines for synchronization points, and recent research demonstrates algorithms that can mitigate the effect of slow participants~\citep{CCFed2020,reisizadeh2020stragglerresilient,li2018federated}. These may not be necessary in cross-silo settings, where dedicated compute hardware and network availability can be more reliable.

\paragraph{Extra cost for privacy and security: secure aggregation}
It is desirable for FL system to accommodate various privacy and security methods. However, such methods may introduce extra cost in the system. One example is the  cryptographic protocol \secagg \citep{bonawitz2017practical} used to protect the privacy of the participants during the communication back to the central aggregator. 
\secagg relies on shared secret keys for each pair of participating users, which adds quadratic complexity to the beginning of every communication round.  Alternatively, \cite{he2020robustsecure} requires a second non-colluding auxiliary server, giving a secure scheme with computation and communication overhead of at most $2\times$.
To perform single-server \secagg efficiently, \citet{bell2020secagg} presents a protocol where both client computation and communication depend logarithmically the number of participating clients, potentially enabling scaling the protocol to much greater numbers of clients per communication round. Turbo-Aggregate~\citep{so2020turbo} employs a multi-group circular strategy for efficient model aggregation, and leverages additive secret sharing and coding techniques to handle user dropouts while guaranteeing user privacy, which has a complexity that grows (almost) linearly with the number of the users, while providing robustness to 50\% user dropout. 
Furthermore, encryption generally reduces compression ratios, as it tries to make the data indistinguishable from random values. 
Compatibility limitations of secure computation with federated algorithms are further discussed in \Cref{sec:private_agg}.

\subsection{Computation and Memory Costs} \label{sec:system_computation}
As described in \cref{sec:practical_guidelines}, performing optimization computations at edge devices, such as local SGD and client drift mitigation, incurs computation cost of the client. In cross-device settings, participants are generally resource-constrained devices such as smartphones. We must take such computation limitations into consideration to design computation-efficient federated optimization strategies that are viable for real-world deployments.

\paragraph{Heterogeneous resources} Large variability in training participant resources can lead to difficulty in real-world systems, particularly in cross-device settings. High-end devices with powerful computation and memory resources can train larger models while low-end devices can only train smaller models. Systems that ``close" a round of communication once a threshold of participants are finished may unfairly bias results towards those resource-rich participants. How to take advantage of these high-end devices and efficiently aggregate heterogeneous models at the server side remains an open research question.

\paragraph{System policies} In cross-device settings where participants are generally smartphone devices, resources are usually shared with other non-learning processes running on the system. System policies may introduce requirements such as the device being in an idle state, connected to particular networks, or charging \citep{bonawitz19sysml,paulik2021federated}, which can severely limit the devices' ability to participate in training. Compared to dedicated data center compute resources, this will result in unreliable compute availability and fluctuating resources \citep{yang2018,fangzeng2018nestdnn}.

\paragraph{Memory constraints} Smartphone clients participated in federated optimization have much less available memory than the high performance servers used in centralized data center training and federated learning simulation. The trend in machine learning is ever larger and deeper models. Combined with more complex optimization techniques that require additional parameters (momentum, etc.), training a model quickly becomes infeasible on devices with limited resources. Techniques such as gradient checkpointing \citep{chen2016sublinearmemory} can reduce memory usage at the cost of more computation. \citet{he2020group} investigate performing more computation on the resource-rich server, though such techniques may not apply to fully decentralized federated learning regimes.

\subsection{Analytical Communication and Computation Costs} \label{sec:evaluate_communication}

Building a large-scale federated learning system is sometimes infeasible for academic researchers, and inconvenient for exploring and prototyping research ideas. We now discuss approaches to evaluate the practical communication and computation cost before deploying the algorithms in a real system. We'll examine the scenarios where the communication and computation costs can be analyzed directly, e.g., if two algorithms have identical amounts of computation and communication in one round of training. In this case the conventional metric of “number of total rounds to complete training” is a reliable metric for comparison. This analytical comparison can be used when one algorithm is strictly better on computation and/or communication cost, e.g., fast lossless compression for communication. However, we defer the discussion to \cref{sec:basicmodel} when two algorithms have different computation and communication costs and it is difficult to tell which one is better from simulation only. 

In one of the earliest federated optimization papers, \citet{mcmahan17fedavg} measured the performance of various algorithms in terms of how the training loss varies with the number of communication rounds (which we often refer to as simply a round). Note that this is a useful method for comparing methods with the same amount of communication and local computation. For example, this is used by \citet{li2018federated} to accurately compare \fedavg with \fedprox, and by \citet{reddi2021adaptive} to compare \fedavg with methods such as \fedadam. However, many works on federated optimization alter the number of local gradient evaluations or the amount of communication per round. In such cases, it is not sufficient to compare the convergence of different methods only in terms of the number of communication rounds. Instead, we suggest using a metric of estimated wall-clock time spent per round in a real-world FL deployment.

In federated learning, wall-clock time can be broken down into the local computation time $T_{\text{comp}}$ and the communication time $T_{\text{comm}}$. The local computation time is proportional to the number of gradient evaluations per round, which depends on the optimization algorithm. The communication time $T_{\text{comm}}$ can be formulated as $\alpha_{|\activeClients|} + \beta \modelSize$. The parameter $\alpha_{|\activeClients|}$ is a fixed latency for every communication with the central server irrespective of the bits transmitted, which can be an increasing function of the number of active clients $|\activeClients|$. The size of the updates communicated per round is $\modelSize$ bits and the resulting latency is $\beta\modelSize$, where $1/\beta$ is available transmission bandwidth. Observe that the relative values of the fixed latency  $\alpha_{|\activeClients|}$ and the message transmission time $\beta\modelSize$ depend on the model size, the optimization algorithm as well as the communication infrastructure underlying the federated learning framework. In \Cref{tab:alg_cost}, we provide a few examples of the computation and communication costs in some recent federated optimization algorithms.

\begin{table}[!ht]
    \centering\small
    \begin{tabular}{c | c | c | c } \toprule
        \textbf{Algorithms} & \textbf{Message size} & \textbf{Comm. latency} & \textbf{Grad. evaluations} \\ \midrule
        \fedavg \citep{mcmahan17fedavg} & 2$\modelSize$ & $\alpha_{|\activeClients|}$ & $\localStep b$ \\
        \fedadam/\fedyogi \citep{reddi2021adaptive} & 2$\modelSize$ & $\alpha_{|\activeClients|}$ & $\localStep b$ \\
        \fedpa \citep{alshedivat2020federated} & 2$\modelSize$ & $\alpha_{|\activeClients|}$ & $\localStep b$ \\
        \scaffold \citep{karimireddy2019scaffold} & 4$\modelSize$ & $\alpha_{|\activeClients|}$ & $\localStep b$ \\
        \mime/\mimelite \citep{karimireddy2020mime} & 4$\modelSize$/3$\modelSize$ & $\alpha_{|\activeClients|}$ & $\localStep b + |\data_i|$ \\\bottomrule
    \end{tabular}
    \caption{Examples of communication and computation costs in some recent federated optimization algorithms. All metrics are evaluated for one round (\ie one global iteration). In the table, $\alpha$ is a function describing how the communication latency scales with the number of active clients. The total training time for one round can be formulated as a linear combination of the above three metrics and the weight of each metric varies in different systems. Most previous works use communication rounds as the measure of communication cost. This makes sense when the communication latency dominates the other two terms and the scaling function $\alpha_{|\activeClients|}$ is a constant.}
    \label{tab:alg_cost}
\end{table}

The computation and communication formulation $T_{\text{comp}}+T_{\text{comm}}$ of wall-clock time does not consider the work performed by the central server for global model update. Unlike the edge device participants, the central server is generally not resource constrained in the cross-device setting, and the computation for updating the global model is not significant for generalized \fedavg algorithms. Next we will propose a system model that enables using measurements of simulation time of federated optimization algorithms to achieve ballpark estimations of real world wall-clock time.

\subsection{Basic Model to Estimate Round Time of Cross-Device Training}\label{sec:basicmodel}

\paragraph{Motivation} We propose a basic system model to estimate the communication efficiency of deploying a federated algorithm to real-world systems. We hope this basic model can serve the following purposes. First, it represents a basic start for systems research to bridge FL simulation and deployment, which is challenging and lacks attention in academia so far. Second,  this model can provide a rough estimation to compare the performance of two federated learning algorithms primarily in execution time when deployed in a cross-device federated learning system. Third, this basic model can inspire federated optimization algorithms and system co-design, where federated optimization can be designed and adjusted considering various system constraints for potential practical deployment, and federated systems design can be influenced when considering support for various types of federated optimization algorithms. The model discussed here is simplified, we did not consider stragglers and many other real-world system constraints mentioned in \cref{sec:system_communication} and \cref{sec:system_computation}. We notice a few recent papers also discussed system modeling in simulation for federated learning \citep{yang2019characterizing,lai2021fedscale}.

\paragraph{Basic execution model}
Our basic model will estimate the execution time per round when deploying an optimization algorithm in cross-device federated learning system as follows,

\begin{align} \label{eq:system}
\begin{split}
T_\mathrm{round} (\algsymbol) = T_\mathrm{comm}(\algsymbol) + T_\mathrm{comp}(\algsymbol), \quad &  T_\mathrm{comm} (\algsymbol) = \frac{S_\mathrm{down}(\algsymbol)}{B_\mathrm{down}}+\frac{S_\mathrm{up}(\algsymbol)}{B_\mathrm{up}}, \\
T_\mathrm{comp} (\algsymbol) = \max_{j \in \data_\mathrm{round}} T^j_\mathrm{client} + T_\mathrm{server}(\algsymbol), \quad & T^j_\mathrm{client} (\algsymbol) = R_\mathrm{comp} T^j_\mathrm{sim}(\algsymbol) + C_\mathrm{comp} ,
\end{split}
\end{align}
where client download size $S_\mathrm{down}(\algsymbol)$, upload size $S_\mathrm{up}(\algsymbol)$, server computation time $T_\mathrm{server}(\algsymbol)$ and client computation time $T^j_\mathrm{sim}(\algsymbol)$ depend on model and algorithm $\algsymbol$.  Simulation time $T_\mathrm{server}(\algsymbol)$ and $T^j_\mathrm{sim}(\algsymbol)$ can be estimated from FL simulation in the data center \footnote{We assume the simulation environment is a data center similar to actual server in a cross-device FL system, and ignore the constant for server system setup time. The server simulation time is usually small for the variant of \fedavg algorithms where the aggregation is (weighted) summation of client updates and server computation is one step of optimizer update based on the aggregated update.}; download bandwidth $B_\mathrm{down}$, upload bandwidth $B_\mathrm{up}$, on-device to data center computational ratio $R_\mathrm{comp}$, and system constant $C_\mathrm{comp}$ are parameters of this system model.

We estimate parameters ($B_\mathrm{down}$, $B_\mathrm{up}$, $R_\mathrm{comp}$, $C_\mathrm{comp}$) of this model based on a real world cross-device federated learning system  \citep{bonawitz19sysml}, which supports Google products like Gboard \citep{hard18gboard}. We pick several production tasks in  \citep{bonawitz19sysml}, and collect the average metrics for anonymous users participating in these tasks in a limited time window. These tasks involve various models like RNN language models and MLP encoders. For download bandwidth $B_\mathrm{down}$ and upload bandwidth $B_\mathrm{up}$, we estimate the parameters based on the slowest device when 90\% of the devices can complete the round and report back to the server for each task, and take the average over different tasks. For computational ratio $R_\mathrm{comp}$ and system constant $C_\mathrm{comp}$, we first estimate simulation time for each task we picked by executing the same model and algorithm on a desktop CPU. Then a linear model in \eqref{eq:system} is fitted based on anonymous aggregated data of on-device training time. Finally, we average the coefficients of the linear model over the tasks picked for $R_\mathrm{comp}$ and $C_\mathrm{comp}$. 

The estimated parameters of the basic model are 
\begin{equation} \label{eq:basic_model_para}
B_\mathrm{down} \sim \text{0.75 MB/secs}, \, B_\mathrm{up} \sim \text{0.25 MB/secs}, \, R_\mathrm{comp} \sim \text{7}, \, \text{ and } C_\mathrm{comp}  \sim \text{10 secs}.
\end{equation}
The numbers in equation \cref{eq:basic_model_para} are only intended to provide some insights of real cross-device federated learning systems, and the scales of these parameters can be more important than the absolute values. 

\paragraph{Usage}
When there is no clear conclusion from applying analytical analysis mentioned in \cref{sec:evaluate_communication} for simulation experiments, the basic model and estimated parameters from the cross-device FL system \citep{bonawitz19sysml} can be used to compare the speed of federated optimization algorithms for on-device training. In addition, the system model can be used to identify parameter settings (e.g. $B_\mathrm{down}$, $R_\mathrm{comp}$) where the preferred candidate optimization algorithm may change. We only offer a rough estimation of parameters in the basic model, and these parameters can have large variance due to system configuration like population size and geographical distribution of devices. For example, if an algorithm can tolerate a higher dropout rate (less clients need to finish training and report back to the server), then the upload and download bandwidth parameter (estimated from the real system) in the basic model can be improved and communication cost can be reduced. An optimization algorithm can be claimed to be better than another one in scenarios defined by a certain range of the basic model parameters. 

This basic model has limitations. For example, this model is not designed to account for the effect of stragglers, or behavior of various clients. However the model is sufficient to help discuss the settings when a communication efficient algorithm saves more time than a computation efficient algorithm. This basic model is provided here to foster the discussion of more sophisticated models and system algorithm co-design for federated optimization in the future. A simple extension for future work could use a distribution instead of scalar to model the parameters for download bandwidth, upload bandwidth, computational ratio, and system constant.

\begin{figure}[ht]
\centering
\begin{subfigure}
    \centering
    \includegraphics[width=0.45\linewidth]{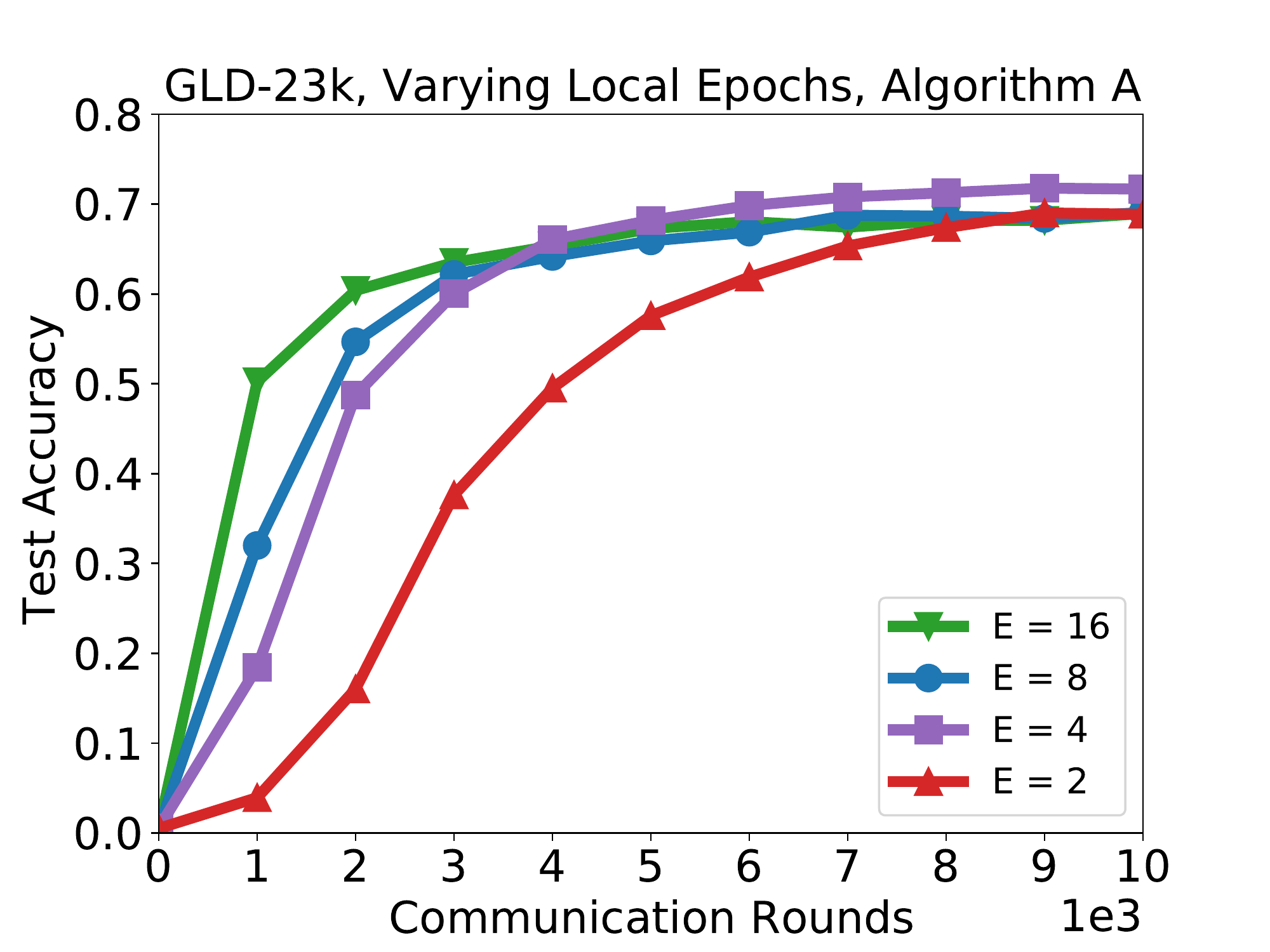}
\end{subfigure}
\begin{subfigure}
    \centering
    \includegraphics[width=0.45\linewidth]{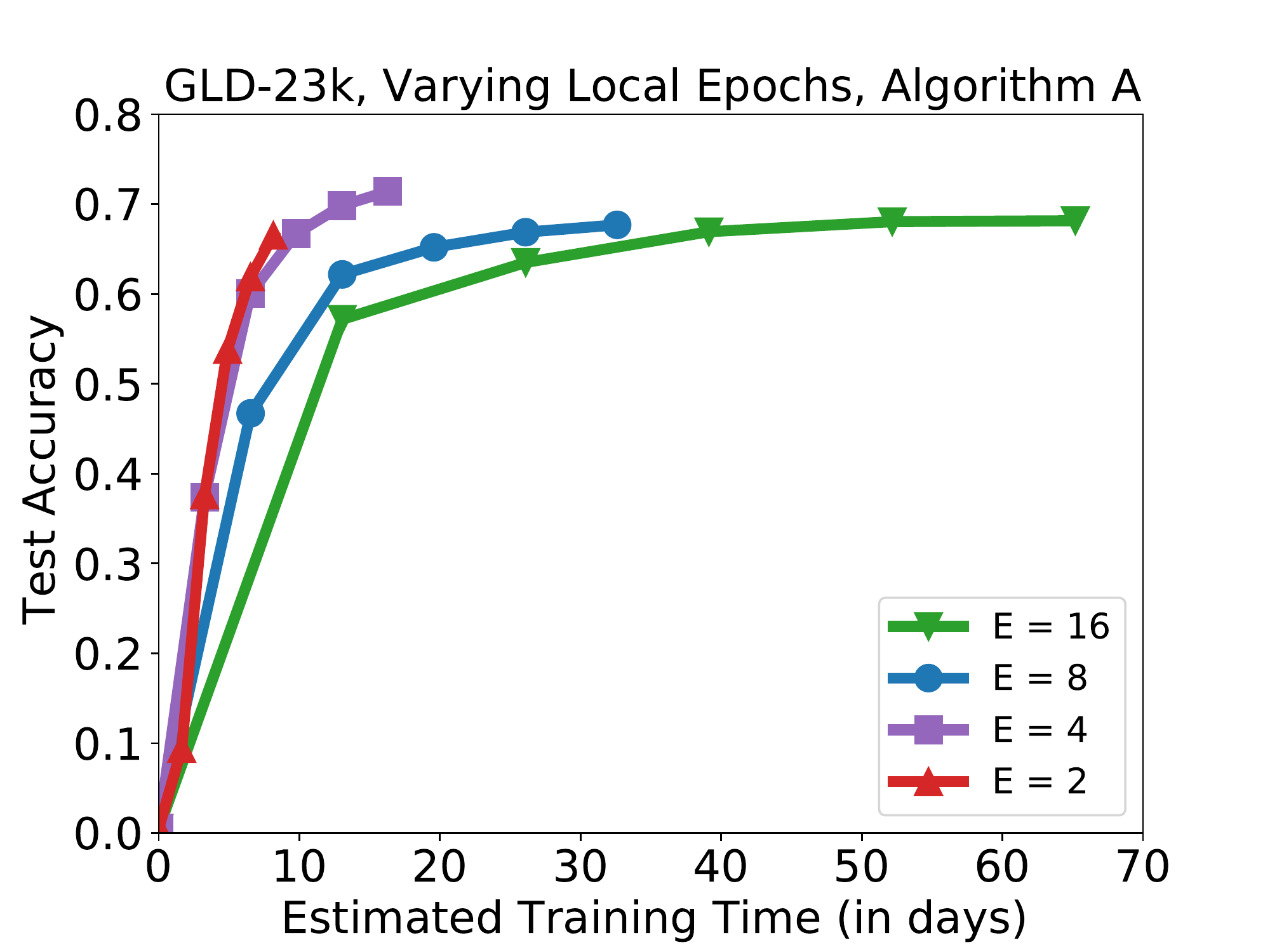}
\end{subfigure}
\caption{Test accuracy on GLD-23k for Algorithm A, for varying numbers of local epochs $E$ per round. We plot the test accuracy versus the number of communication rounds (left) and an estimate of the execution model round completion time $T_\mathrm{round} (\algsymbol)$ (right). We use learning rates $\eta = 0.1$ and $\eta_s = 1.0$.}
\label{fig:gld23k_system_model_alg_a}
\end{figure}

\paragraph{Example}
In \cref{fig:gld23k_system_model_alg_a} we plot the test accuracy of a model on the GLD-23k dataset (see \Cref{sec:evaluation} for the simulation setup), as a function of the number of communication rounds, and the estimated round completion time $T_\mathrm{round}(\algsymbol)$. We use Algorithm A\footnote{See the beginning of \cref{sec:evaluation} for a discussion on why we obfuscate the algorithm.}, and vary the number of local epochs of training $E$ each client performs per round. Based on simulations conducted on a single CPU (in order to better emulate the lightweight capabilities of cross-device clients), we estimate that the model used for GLD-23k requires roughly $0.127$ seconds to process a single data example. This allows us to estimate $T^j_\mathrm{sim}(\algsymbol)$ in this case as simply $0.127$ multiplied by the number of examples processed by a client. Using the estimated constants $B_\mathrm{down}, B_\mathrm{up}, R_\mathrm{comp}$, and $C_\mathrm{comp}$ above, we can then estimate $T_\mathrm{round}(\algsymbol)$ using \eqref{eq:system}.
While large local epoch $E$ settings converge faster if we look at the curves for the first 2000 communication rounds, the extra computation at the clients consumes extra training time such that the entire training time slows down drastically. 
It is worth mentioning that while more rounds can be accomplished under given wall-clock time constraint when $E$ is small, this can have potential negative consequences when combining federated optimization with differential privacy techniques. 
More results on Algorithm C and the Stack Overflow experiment can be found in \cref{appendix:additional_results}.

\subsection{Research Frameworks for Simulating Distributed Systems}

 In the last few years a proliferation of frameworks have been developed to enable federated optimization researchers to better explore and study algorithms, both from academia (\texttt{LEAF} \citep{caldas2018leaf}, \texttt{FedML} \citep{he2020fedml}, \texttt{Flower} \citep{beutel2020flower}) and from industry (\texttt{PySyft} \citep{ryffel2018generic}, \texttt{TensorFlow Federated} (TFF) \citep{TFF2019}, \texttt{FATE} \citep{yang2019federated}, \texttt{Clara} \citep{ClaraTraining}, \texttt{PaddleFL} \citep{ma2019paddlepaddle}, \texttt{Open FL} \citep{OpenFLFramework}, \texttt{FedJAX} \citep{fedjax2020github}). In this section we highlight some desirable aspects from such frameworks.

Traditionally many frameworks largely focused on enabling empirical studies of algorithms in a data center for experimental simulation. The community can benefit from frameworks that also facilitate simulating and analyzing the systems aspects such as elapsed training time, communication costs, and computation costs under different architectures.

A framework for designing federated algorithms that can be analyzed and deployed on real world systems might include features such as:

\begin{itemize}
    \item \emph{Platform agnostic}: Federated scenarios generally are characterized by heterogeneous compute nodes. The ability to write once and run everywhere is important to reach the participants of the algorithm.
    \item \emph{Expressive communication primitives}: A flexible programming interface that enables expression of diverse network topologies, variable information exchange among workers/clients, and various training procedures.
    \item \emph{Scalability}: Both cross-silo settings and cross-device settings have their own specific scaling needs. For example, cross-silo simulations may require scaling to a small number of data-rich clients, while cross-device simulations may require scaling to a large number of data- and compute-limited clients. A framework that can scale in multiple dimensions adds additional flexibility.
    \item \emph{Extensibility}: new algorithms are developed that can require new capabilities. Keeping pace with the new frontiers of research prevents a framework from becoming obsolete.
\end{itemize}

For more detailed challenges and practice in these topics, we refer to the guidance provided by \citep[Section 7]{kairouz2019advances}.

\subsection{Real-world Deployment Suggestions}

A central challenge in deploying practical federated learning systems is the need for modeling without access to raw centralized data. In a typical centralized training workflow, a modeling engineer is able to quickly test their hypotheses by inspecting data samples, \eg identifying mis-classified examples. This can lead to improved models and can also surface issues with the model or data pipeline. In federated settings, raw data is located on client devices and direct data access for modeling or debugging may be impossible.

\paragraph{De-risk training in simulation with proxy data} When available, proxy data (real or synthetic data that can serve as an approximate substitute for federated data) is useful for testing federated algorithms in simulation. Ideally, a practical federated learning system is built so that taking a working simulation and deploying it to real-world clients is as seamless as possible, to minimize bugs introduced in the process.

To ensure that observations made in simulated federated learning transfer well to production settings, simulated client data must have high fidelity. The most important factor for federated simulations is the clustering of data into pseudo-clients. Production federated data are highly heterogeneous, so insights from federated simulations on homogeneous (IID) client data are not guaranteed to generalize. The distribution of the number of training examples per client should also be matched in simulations, as they affect the optimal configuration of client and server optimizer parameters. Length distributions for sequence-based features should also be matched, as data augmentation techniques may need to be re-tuned. As an example, optimizations made for language modeling on long sentences in simulations are not guaranteed to generalize well to on-device settings with shorter sentence lengths. Typically, improvements to simulation and production federated training are iterative. Insights into production failures lead to simulation improvements, which can be used to optimize better models that eventually generalize better on-device.

\paragraph{Consider alternative supervision instead of labels in on-device training}
One of the primary challenges to real-world federated learning is the lack of supervised labels. Due to the no-peek nature of on-device data, manual labeling of individual training examples by raters is impossible. Given this constraint, three common approaches to federated training attempt to leverage the abundance of unlabeled data: unsupervised learning, semi-supervised learning, and supervised learning with contextual on-device signals. Several early applications of federated learning in production environments relied on unsupervised training (for example, the RNN language model in \citep{hard18gboard}). Techniques such as distillation or noisy student training can be used to build a student model with federated learning that outperforms an original teacher model \citep{hard20keyword}. And rater-assigned labels that are used for traditional server-based training can be replaced in certain circumstances with contextual signals. For example, accepted suggestions or manual corrections by users can be interpreted as supervision signals.

\paragraph{Use federated analytics to characterize client data and model behavior} Federated analytics \citep{federatedanalytics} can be used to get insights from federated data in a privacy-preserving way. These insights can improve the modeling workflow in federated settings. For example, federated analytics can be used to measure the fraction of users who type a particular token for a language modeling task for mobile keyboard applications, which can be useful for deciding whether the token should be part of the model's input vocabulary. Federated aggregation can also be applied on the model's outputs to better understand model behavior (a simple extension of federated aggregation of metrics like loss and accuracy).

\paragraph{Use generative models to characterize client data} \citet{augenstein2019generative} propose using differentially private federated generative models to generate realistic samples of client data. This can in principle be useful for debugging and generating and testing model hypotheses.

\section{Federated Optimization Theory}\label{sec:theory}

In this section, we will briefly discuss possible theoretical analysis for the convergence of current federated learning algorithms. 
In particular, we first introduce some theoretical tools for the convergence analysis in \Cref{sec:fedavg-proof} by presenting a simple proof for the vanilla \fedavg.
Based on the simple analysis, we discuss the effects of local updates and data heterogeneity and compare it with well-known baselines in optimization theory. 
Later in \Cref{sec:theory_advances}, we review more recent literature to demonstrate how to relax the assumptions made for simplicity and improve over the basic results in \Cref{sec:fedavg-proof}.

\subsection{Basic Convergence Analysis for Federated Optimization Algorithms}
\label{sec:fedavg-proof}

We present a simple proof for the vanilla \fedavg \citep{mcmahan17fedavg}, also known as Local SGD or parallel SGD for homogeneous data in the literature, to showcase the structure, tools, and tricks used for getting theoretical results in federated optimization.
We begin by considering a simple setting of the generalized \fedavg framework (\cref{algo:generalized_fedavg}) and later on give pointers to literature for how things change in different settings. 
The techniques and insights provided in this subsection can be found in much previous literature, including \cite{stich2018local,wang2018cooperative,yu2019parallel,karimireddy2019scaffold,bayoumi2020tighter,woodworth2020localSGD,woodworth2020minibatch}.

\subsubsection{Assumptions and Preliminaries}
Formally, we make the following assumptions for the analysis. Specifically, assumptions (i-iv) are about the algorithmic choices in \fedavg; assumption (v) and (vi) are about the properties of local objectives; assumption (vii) is about the data heterogeneity/similarity across clients.
\begin{enumerate}[label=(\roman*)]
    \item At any round $t$, each client takes a total of $\localStep$ local SGD steps with constant learning rate $\eta$, namely $\vx_i^{(t,k+1)} = \vx_i^{(t,k)} - \eta g_i(\vx_i^{(t,k)})$, where $k \in [0, \localStep)$ and $g_i$ is the stochastic gradient at client $i$.
    \item The \serveropt takes a unit descent step with server learning rate $1.0$, namely $\vx^{(t+1)} = \vx^{(t)} + \Delta^{(t)}$.
    \item There are finite ($M$) clients and each client contributes a uniform share of the global objective. 
    We label the clients as $\{1, 2, \ldots, M\}$.  Formally, we have $F(\vx) = \frac{1}{M} \sum_{i=1}^M F_i(\vx)$.
    \item Each client participates \emph{every round}, namely $\mathcal{S}^{(t)} \equiv \{1, 2, \dots, M\}$. 
    \item Each local objective $F_i(\vx)$ is \emph{convex} and $L$-\emph{smooth}. 
    \item Each client can query an \emph{unbiased} stochastic gradient with $\sigma^2$-\emph{uniformly bounded variance} in $\ell_2$ norm, namely 
    \begin{equation}
        \Exs [g_i(\vx_i^{(t,k)}) | \vx_i^{(t,k)}] = \nabla F_i(\vx_i^{(t,k)})
        ,
        \quad
        \Exs [\|g_i(\vx_i^{(t,k)}) - \nabla F_i(\vx_i^{(t,k)}) \|^2 | \vx_i^{(t,k)}] \leq \sigma^2.
    \end{equation}
    \item The difference of local gradient $\nabla F_i(\vx)$ and the global gradient $\nabla F(\vx)$ is $\zeta$-uniformly bounded in $\ell_2$ norm, namely
    \begin{equation}
        \max_{i} \sup_{\vx} \left\| \nabla F_i(\vx_i^{(t,k)}) - \nabla F(\vx_i^{(t,k)}) \right\| \leq \zeta.
    \end{equation} 
\end{enumerate}

In a typical convergence proof we want to show that our iterates $\vx^{(t, k)}$ is taking the objective function value $F(\vx^{(t,k)})$ closer to the optimal value $F(\vx^{\star})$.
In the federated learning setup, we have multiple local iterates on clients and we want to ensure all of them are approaching the minimizer.
To handle iterates from multiple clients, a concept of shadow sequence is introduced (commonly used in and originating from decentralized optimization literature~\citep{lian17decentralized,yuan2016convergence}), which is defined as $\overline{\vx}^{(t,k)} := \frac{1}{M} \sum_{i=1}^M \vx_i^{(t,k)}$. Given this notation, we have
\begin{align}\label{eqn:avg_iter}
    \overline{\vx}^{(t,k+1)} = \overline{\vx}^{(t,k)} - \frac{\lr}{\numClients}\sum_{i=1}^\numClients \sgrad_i(\vx_i^{(t,k)}).
\end{align}
From \Cref{eqn:avg_iter}, observe that the averaged iterates actually perform a perturbed stochastic gradient descent. Its gradients are evaluated at client models $\vx_i^{(t,k)}$ instead of $\overline{\vx}^{(t,k)}$. We want to show that:
\begin{equation}\label{eqn:defn_rate}
     \Exs \left[  \frac{1}{\tau T} \sum_{t=0}^{T-1} \sum_{k=1}^{\tau} F( \overline{\vx}^{(t,k)} ) - F(\vx^{\star}) \right]
        \leq \text{an upper bound decreasing with } T.
\end{equation}
How fast the above quantity decreases with $T$ is called the \emph{rate of convergence}. It is also worth noting that at the end of each round, we have $\overline{\vx}^{(t,\localStep)} = \overline{\vx}^{(t+1,0)}=\vx^{(t+1)}$. So the above quantity~\Cref{eqn:defn_rate} also quantifies the convergence of the global model.

\subsubsection{Main Results}  We introduce two lemmas to facilitate the proof.
\begin{enumerate}
    \item Show that we are making progress in each round, i.e. $\Exs \left[  \frac{1}{\tau} \sum_{k=1}^{\tau} F( \overline{\vx}^{(t,k)} ) - F(\vx^{\star}) \right]$ is bounded by the difference of a potential function evaluated at $t$ and $t+1$, measuring the progress of optimization, plus an additional small error term. 
    \item Show that all client iterates remain close to the global average/shadow sequence, i.e. in expectation  $ \|\vx_i^{(t,k)} -  \overline{\vx}^{(t,k)} \|^2 $ remains bounded.
\end{enumerate}
The first lemma is similar to analysis strategy for centralized optimization algorithms, but now we have added responsibility to ensure no client is deviating away.
Once we have these two, by simple telescoping over $t$ we can show that over $T$ rounds we have made significant progress towards the optimal value. We formally state the two lemmas below, and leave elaborate proof of the lemmas to \cref{appendix:proof}.
\begin{lemma}[Per Round Progress]
    \label{lem:1}
    Assuming the client learning rate satisfies $\eta \leq \frac{1}{4L}$, then
    \begin{align*}
    \Exs \left[  \frac{1}{\tau} \sum_{k=1}^{\tau} F( \overline{\vx}^{(t,k)} ) - F(\vx^{\star}) \middle| \mathcal{F}^{(t,0)} \right]
    \leq  & \underbrace{\frac{1}{2\eta \tau} \left( \left\|\overline{\vx}^{(t,0)} - \vx^{\star} \right\|^2 - \Exs\left[ \left\|\overline{\vx}^{(t,\tau)} - \vx^{\star} \right\|^2 \middle| \mathcal{F}^{(t,0)} \right]  \right)}_{\mathrm{Progress}} \\
    & + \underbrace{\frac{\eta \sigma^2}{M} + \frac{L}{M\tau} \sum_{i=1}^M \sum_{k=0}^{\tau-1} \Exs \left[ \left\| \vx_i^{(t,k)} -  \overline{\vx}^{(t,k)} \right\|^2 \middle | \mathcal{F}^{(t,0)} \right]}_{\mathrm{Deviation}},
    \end{align*}
    where $\mathcal{F}^{(t,0)}$ is the $\sigma$-field representing all the historical information up to the start of the $t$-th round.
\end{lemma}
\begin{lemma}[Bounded Client Drift]
    Assuming the client learning rate satisfies $\eta \leq \frac{1}{4L}$,
    \label{lem:2}
    \begin{equation*}
        \Exs \left[ \left\| \vx_i^{(t,k)} -  \overline{\vx}^{(t,k)} \right\|^2 \middle | \mathcal{F}^{(t,0)} \right] 
        \leq 
        18 \tau^2 \eta^2 \zeta^2 + 4 \tau \eta^2 \sigma^2,
    \end{equation*}
    where $\mathcal{F}^{(t,0)}$ is the $\sigma$-field representing all the historical information up to the start of the $t$-th round.
\end{lemma}
%
%
%
%
\noindent Combine \cref{lem:1,lem:2} and telescope $t$ from $0$ to $T-1$ to obtain the main theorem as follows.
\begin{theorem}[Convergence Rate for Convex Local Functions]
    \label{thm:1}
    Under the aforementioned assumptions (a)-(g), if the client learing rate satisfies $\lr \leq \frac{1}{4L}$, then
    one has
    \begin{equation}
        \Exs \left[  \frac{1}{\tau T} \sum_{t=0}^{T-1} \sum_{k=1}^{\tau} F( \overline{\vx}^{(t,k)} ) - F(\vx^{\star})  \right]
        \leq \frac{D^2}{2\eta \tau T} 
        + \frac{\eta \sigma^2}{M} + 4 \tau \eta^2 L \sigma^2   +  18 \tau^2 \eta^2 L \zeta^2,
    \label{eq:thm:1}
    \end{equation}
    where $D := \|\vx^{(0,0)} - \vx^{\star}\|$. Furthermore, when the client learning rate is chosen as
    \begin{equation}
        \eta = \min \left\{ \frac{1}{4L}, \frac{M^{\frac{1}{2}} D}{\tau^{\frac{1}{2}} T^{\frac{1}{2}} \sigma}, \frac{D^{\frac{2}{3}}}{\tau^{\frac{2}{3}} T^{\frac{1}{3}} L^{\frac{1}{3}} \sigma^{\frac{2}{3}}}, \frac{D^{\frac{2}{3}}}{\tau T^{\frac{1}{3}} L^{\frac{1}{3}} \zeta^{\frac{2}{3}}} \right\},
        \label{eq:thm:2}
    \end{equation}
    we have
    \begin{equation}
        \Exs \left[  \frac{1}{\tau T} \sum_{t=0}^{T-1} \sum_{k=1}^{\tau} F( \overline{\vx}^{(t,k)} ) - F(\vx^{\star}) \right]
        \leq
        \underbrace{\frac{2LD^2}{\tau T} 
        + \frac{2\sigma D}{\sqrt{M \tau T}}}_{\mathrm{Synchronous \ SGD}} 
        + \underbrace{\frac{5 L^{\frac{1}{3}} \sigma^{\frac{2}{3}} D^{\frac{4}{3}} }{\tau^{\frac{1}{3}} T^{\frac{2}{3}} } 
        + \frac{19 L^{\frac{1}{3}} \zeta^{\frac{2}{3}} D^{\frac{4}{3}} }{T^{\frac{2}{3}} }}_{\mathrm{Add'l \ errors \ from \ local \ updates \ \& \ non-IID \ data}}.
        \label{eq:thm:3}
    \end{equation}
\end{theorem}

\subsubsection{Discussion}
\Cref{thm:1} can be used to obtain insights for \fedavg or local-update algorithms. Below, we discuss the implications of \Cref{thm:1} in detail. Similar conclusions can be found in previous literature \cite{stich2018local,wang2018cooperative,yu2019parallel,karimireddy2019scaffold,bayoumi2020tighter,woodworth2020localSGD,woodworth2020minibatch}.

\paragraph{Effects of local steps}
An important benchmark to compare with is \emph{synchronous SGD} that synchronizes local models at each local iteration (that is, setting the number of local steps $\localStep$ to be~$1$ in \fedavg). In synchronous SGD, the last two terms in \Cref{eq:thm:1,eq:thm:3} will not appear at all and the rate of convergence is $\mathcal{O}(1/\localStep T + \sigma/\sqrt{\numClients \localStep T})$. However, when clients perform multiple local steps, there will be an additional error in the upper bound (\ie last two terms in \Cref{eq:thm:1,eq:thm:3}). Fortunately, observe that the additional error terms is in proportion to $\lr^2$ and can decay in a rate of $\mathcal{O}(1/T^{\frac{2}{3}})$ (when the learning rate is small enough according to \Cref{eq:thm:2}). When the total communication rounds $T$ is sufficiently large and the stochastic noise is not zero ($\sigma^2 \neq 0$), performing local steps will not degrade the rate of convergence.

\paragraph{Savings in communication rounds}
Taking local steps can save total communication rounds compared to synchronous SGD. To be more specific, when the total number of gradient evaluations/computations across all clients ($K = \numClients \localStep T$) is fixed and the local steps $\localStep$ satisfies
\begin{align}\label{eqn:upb_tau}
    \localStep \leq \min\left\{ \frac{\sigma}{D L}\frac{K^{\frac{1}{2}}}{M^2}, \frac{\sigma}{\zeta}\sqrt{\frac{\sigma}{D L}\frac{K^{\frac{1}{2}}}{\numClients^2}} \right\},
\end{align}
the error upper bound \Cref{eq:thm:3} will be dominated by the second term $\mathcal{O}(1/\sqrt{K})$, which is the same as synchronous SGD. On the other hand, the local-update algorithm only takes $T$ communication rounds within $K$ parallel SGD iterations, while synchronous SGD costs $\localStep T$ communication rounds. It is obvious that when the upper bound of local steps \Cref{eqn:upb_tau} becomes larger, there will be more communication savings. Therefore, the quantity \Cref{eqn:upb_tau} represents the largest savings in communication rounds and is reported by many previous works.

\paragraph{Effects of data heterogeneity}
\Cref{thm:1} also depicts that the data heterogeneity exacerbates the side effects of local updates on the convergence. Note that the additional error terms in \Cref{eq:thm:1} is in an order of $\mathcal{O}(\localStep^2)$. However, if all local objectives are identical so that $\zeta=0$, then the additional error terms will be just linear to the number of local steps. Besides, from \Cref{eqn:upb_tau} we know that, in the presence of high data heterogeneity ($\zeta \gtrsim \sigma$), the largest number of local steps is $\tau = \mathcal{O}(K^{\frac{1}{4}}M^{-1})$, which is much smaller than the IID data distribution case ($\tau = \mathcal{O}(K^{\frac{1}{2}}M^{-2})$).

\begin{remark}
While the above discussions around \Cref{thm:1} are mainly under the stochastic setting ($\sigma \neq 0$), the conclusions may change in the deterministic setting (\ie clients use full-batch gradients to perform local updates and $\sigma = 0$). In particular, the rate of convergence under heterogeneous data setting will be substantially slowed down to $\mathcal{O}(1/T^{\frac{2}{3}})$ from $\mathcal{O}(1/\localStep T)$.
\end{remark}


\paragraph{Comparison with large-batch synchronous SGD} So far, all the above discussions are based on the comparison with synchronous SGD. It is worth noting that there is another natural baseline to compare with. In particular, if we force all clients to use $\localStep$-fold larger mini-batch size in synchronous SGD algorithm, then this algorithm can also save $\localStep$ times communication, which is the same as local-update algorithms. We refer to this algorithm as \emph{large-batch synhcronous SGD}\footnote{Both large-batch and regular synchronous SGD are equivalent to mini-batch SGD. In order to distinguish the different batch sizes ($\numClients \localStep$ versus $\numClients$) and avoid confusions, we refer to them with different names.}. It has been well understood (see \cite{dekel2012optimal}) that the worst-case error of large-batch synchronous SGD is:
\begin{align}\label{eqn:error_sgd}
    \Theta\left(\frac{LD^2}{T} + \frac{\sigma D}{\sqrt{M \tau T}}\right).
\end{align}
Now we are going to compare \Cref{eqn:error_sgd} and \Cref{eq:thm:3} in the regime where the statistical term $\mathcal{O}(1/\sqrt{M \localStep T})$ does not dominate the rate\footnote{Otherwise, both algorithms have the same worst-case performance.}. The key insight is that \fedavg or local-update algorithms are certainly not universally better, but there do exist some regimes where they can improve over large-batch synchronous SGD. For example, when the data heterogeneity is very low such that $\zeta^2 \precsim 1/T$, the worst case error bound of \fedavg~\Cref{eq:thm:3} will be smaller than that of large-batch synchronous SGD~\Cref{eqn:error_sgd}. Similar conclusions hold for data homogeneous case (\ie when $\zeta=0$), in which \fedavg's guarantee \Cref{eq:thm:3} can be better than \Cref{eqn:error_sgd} when the number of local steps is sufficiently large $\localStep \gtrsim T$. Moreover, in the homogeneous setting, it has been shown in \cite{woodworth2020localSGD} that under possibly restricted conditions, such as Hessian smooth loss functions, \fedavg can be strictly better than large-batch synchronous SGD.

\paragraph{Lower bounds} 
Looking at lower bounds can enlighten us about what aspects can be improved and what cannot. For example, \citep{karimireddy2019scaffold} derived lower bounds showing that heterogeneity can be particularly problematic, showing degradation as heterogeneity increases. 
Putting this in light of above comparisons to large-batch synchronous SGD, the lower bounds of \citep{karimireddy2019scaffold} show that \Cref{eq:thm:3} cannot be substantially improved, and thus \fedavg can be strictly inferior to large-batch synchronous SGD when the level of data heterogeneity is large or the number of local steps is small (in the data homogeneous setting)~\citep{woodworth2020minibatch} .
%
%
%
Further insights into problems caused by heterogeneity can be obtained by looking at the communication lower bound of \cite{Arjevani2015complexity}, which matches the upper bound of accelerated gradient descent in the general heterogeneous setting.
So, in order to achieve any nontrivial improvements in communication complexity,
one has to introduce additional conditions on the
second order heterogeneity, as considered in the distributed computing 
literature (without device sampling) by
\citep{shamir2013dane,zhang2015disco,reddi2016aide} 
and in federated learning by \citep{karimireddy2019scaffold,karimireddy2020mime}.

\subsection{Advances in Convergence Guarantees}\label{sec:theory_advances}

The convergence property of \fedavg (or local-update algorithms) has been extensively investigated
in recent years. One can relax some assumptions in \Cref{sec:fedavg-proof} via more complex analysis. In this subsection, we briefly discuss some recent advances in convergence guarantees and how to possibly improve over the results in \Cref{sec:fedavg-proof}.

\subsubsection{Assumptions on data heterogeneity} \label{section:heterogeneity_assumptions}
The similarity, or dissimilarity, between the functions in~\eqref{eqn:global_obj} is an important quantity that determines the difficulty of the problem, both in practice and in theory. The challenge for theoretical analysis can be seen through a simple thought experiment. Consider a hypothetical situation where all clients own exactly the same set of training data. In this situation, from the perspective of any client, nothing could be gained from others' data, and hence there is no need for communication among the clients. On the other hand, in the hypothetical situation where all devices have IID samples from the same underlying distribution $\data$, there is incentive to learn jointly in a federated fashion as there is more statistical power in the union of these IID datasets, especially when the number of devices is large. However, this IID regime is still not particularly challenging as the high level of data homogeneity inherent in this regime can be exploited. The most challenging regime, and the one which often occurs in practice, is when the clients contain data samples from widely different distributions.

In order to quantify the dissimilarity between client data distributions, various measures have been proposed. Most common are quantities that are based on the gradients, in particular, the global bound on the variance of the gradients, taking the form
\begin{align} \label{eqn:gradient_similarity}
&\Exs_{i \sim \clientDist} \left\| \nabla \obj_i(\vx) - \nabla \obj(\vx) \right\|^2 \leq \zeta^2 + \beta^2 \|\nabla \obj (\vx) \|^2 \,, & &\forall \vx \in \R^d\,,
\end{align}
where $\zeta^2, \beta^2$ are parameters. When $\beta=0$, the above bound can recover assumption (vii) used in \Cref{sec:fedavg-proof}.  But it has been shown that setting $\beta>0$ can lead to refined rates~\citep{li2018federated,karimireddy2019scaffold,wang2020tackling,karimireddy2020mime,koloskova2020unified, LSGDunified2020}. When $\zeta=0, \beta\neq 0$, \Cref{eqn:gradient_similarity} recovers the assumption used in \cite{li2018federated}.

\paragraph{No data similarity} Given that the client's data are highly non-iid, it might happen that there is no finite $\beta,\zeta$ such that~\eqref{eqn:gradient_similarity} holds. Such an issue has been recently addressed by~\citet{bayoumi2020tighter} given that problem~\eqref{eqn:global_obj} has an unique minimizer $\vx^\star$ (as for instance for strongly convex functions, or non-convex functions under the Polyak-\L{}ojasiewicz condition). Specifically, it is enough to impose the condition~\eqref{eqn:gradient_similarity} only at a single point $\vx^\star$ instead of the full domain $\R^d$. Similar to the observation of \Cref{thm:1}, a price is paid for worse communication guarantees even with this weaker assumption~\citep{bayoumi2020tighter,karimireddy2020mime,koloskova2020unified}. 

\paragraph{Other heterogeneity assumptions} 
\citet{li2019convergence} quantify the data heterogeneity based on the quantity $\obj^\star - \E_{i\sim \clientDist}[\obj^\star_i]$, where $F^\star$ and $F_i^\star$ are the minimum values of $F$ and $F_i$ respectively. 
\citet{zhang2019overparametrization} assume---motivated by over-parametrized problems---that the intersection of the solution sets on the clients is nonempty.
Other works assume conditions on the Hessians~\citep{shamir2013dane,Arjevani2015complexity, karimireddy2020mime,karimireddy2019scaffold} or assume that the deviation between each client's and the global objective is smooth~\cite{cen2019dsvrg}. 
Given the strong lower bounds known for the performance of local update methods under the standard setting of \eqref{eqn:gradient_similarity}, exploring such alternate heterogeneity assumptions is an important direction to bridge the gap between the theoretical and practical performance of the methods.

\subsubsection{Extensions and Improvements} \label{subsec:theory_extension}

\paragraph{Analyses for various objective functions}
Instead of assuming the local objective function to be general convex, one can extend the results for strongly convex local objectives, see \cite{stich2018local,bayoumi2020tighter,karimireddy2019scaffold,woodworth2020minibatch,koloskova2020unified,LSGDunified2020}. In the strongly convex setting, the additional errors in \Cref{eq:thm:3} will decay faster along with the communication rounds $T$. We defer interested readers to Table 2 in \cite{woodworth2020minibatch} that summarizes the current state-of-the-art results. There is much literature analyzing  non-convex settings, the convergence rates (in terms of $\Exs[\|\nabla \obj(\vx^{(t)})]\|^2$) of which are similar to the general convex setting in \Cref{thm:1}, see \cite{wang2020tackling,karimireddy2019scaffold,koloskova2020unified}.

\paragraph{Analyses with client sampling}
Analyzing \fedavg with client sampling under certain constraints can generally be done by standard techniques. In particular, when using client sampling, the update rule of the virtual sequence \Cref{eqn:avg_iter} can be rewritten as follows
\begin{align}
    \overline{\vx}^{(t,k+1)} = \overline{\vx}^{(t,k)} - \lr\sum_{i\in \activeClients^{(t)}} w_i^{(t)} \sgrad_i(\vx_i^{(t,k)})
\end{align}
where $w_i^{(t)}$ can be chosen to ensure that $\Exs_{\activeClients}[\sum_{i\in \activeClients} w_i^{(t)} \sgrad_i(\vx_i^{(t,k)})] = \sum_{i=1}^\numClients \sgrad_i(\vx_i^{(t,k)})/\numClients$ so that the client sampling scheme just adds a zero-mean noise to the update rule \Cref{eqn:avg_iter} and the convergence of \fedavg can be naturally guaranteed. In the resulting error bound, the number of total clients $\numClients$ will be replaced by the number of active clients per round $| \activeClients^{(t)} |$. Interested readers can refer to \cite{li2019convergence,li2018federated,wang2020tackling,yang2021achieving,karimireddy2019scaffold,reddi2021adaptive,LSGDunified2020} for the technical details.

\paragraph{Analyses with advanced \serveropt and \clientopt}
While the basic analysis in \Cref{sec:fedavg-proof} assumes that the server takes a unit descent step to update the global model, one can  replace it with other \serveropt. It has been shown in \cite{karimireddy2019scaffold,reddi2021adaptive,karimireddy2020mime} that, with properly chosen server learning rate $\slr$ for $\vx^{(t+1)} = \vx^{(t)} + \slr \localChange^{(t)}$, the convergence rate can be improved. When the \serveropt is momentum SGD or adaptive optimization methods, \citet{reddi2021adaptive} proves that they preserve the same convergence rate as vanilla \fedavg but with significantly improved empirical performance. In another line of research, using momentum or accelerated SGD or adaptive optimizers on clients has been explored in \cite{yu2019linear,wang2020tackling,yuan2020federated,karimireddy2020mime,wang2021local}. In particular, \citet{yuan2020federated,karimireddy2020mime} proposed provably accelerated variants of \fedavg. 

\paragraph{Analyses with communication compression}
As discussed in \cref{sec:algo_comm_constraint}, combining \fedavg with compression can potentially boost the communication efficiency. The theoretical analysis without local steps was explored in \citep{alistarh2018sparse,stich2018sparse}, and with local steps in \cite{Qsparse-local-sgd19}. More recently there is much more literature studying communication compression, the update rule of which can be written as $\vx^{(t+1)} = \vx^{(t)} - \eta_t \sum_{i=1}^M p_i \cC_i^{(t)}(g_i(\vx^{(t)}))$ where $\cC_i^{(t)}$ denotes the suitably chosen compressor. In order to guarantee the convergence, certain constraints have to be enforced on the class of compressors. \citet{alistarh2017qsgd,Cnat,VR-DIANA} studied unbiased compressors that satisfy $\E[\cC(\vx)] = \vx$  and $\E[ \|\cC(\vx) -\vx \|^2 ] \leq \omega \|\vx\|^2$ for some $\omega\geq 0$ and all $\vx\in \R^d$. The theoretical analysis of unbiased compressor is by adding a zero-mean noise to the stochastic gradients. Moreover, biased compressors can be analyzed, for instance contractive compressors that satisfy $\E[ \|\cC(\vx) -\vx \|^2 ] \leq (1- 1/\delta) \|\vx\|^2$ for some $\delta\geq 1$ and all $\vx\in \R^d$. If applied naively, contractive compressors can diverge, even on simple low-dimensional piece-wise linear or quadratic problems \citep{karimireddy2019ef,biased2020}. A practical fix was suggested by \citet{seide2014compr} without theoretical guarantees, which is known under various names including {\em error compensation}, {\em error feedback} and {\em memory feedback}. Error compensation remained without any theoretical support until recently \citep{alistarh2018sparse,stich2018sparse,errorSGD,karimireddy2019ef}, with a steady stream of works achieving gradual progress in the theoretical understanding of this mechanism \citep{stich2019,Koloskova2019-DecentralizedEC-2019,DoubleSqueeze2019,biased2020,EC-LSVRG,EC-Katyusha,EC-SGD}.

\paragraph{Analyses with control variates}
As discussed in \Cref{sec:representative_techniques}, control variates methods have been
proposed to explicitly reduce the client shift problem
\cite{karimireddy2019scaffold,karimireddy2020mime}.
The basic idea is to debias the local model updates on each client using a global estimation over multiple
clients, similar to what was used
in communication efficient distributed computing \cite{shamir2013dane}.
Such a bias correction addresses the client shift problem, leading to better
convergence guarantees for local optimization methods. 

Specifically, \emph{without any assumptions on the data heterogeneity}, the \scaffold algorithm developed in \cite{karimireddy2019scaffold} converges at a rate of $O(\nicefrac{1}{M\localStep T})+O(e^{-T})$ for strongly convex functions and $O(\nicefrac{1}{\sqrt{M\localStep T}}))$ for non-convex functions. \scaffold, under second order regularity assumptions, shows strictly better rates than mini-batch SGD matching the lower-bound of \cite{Arjevani2015complexity}.  Similarly, when the \mime framework developed in \cite{karimireddy2020mime} is instantiated with momentum-based variance reduction, the resulting algorithm obtains the optimal convergence rate of $O((M\localStep T)^{-2/3})$ for non-convex functions assuming that the clients' gradients and Hessians satisfy certain closeness properties. This algorithm provably outperforms both large-batch synchronous SGD and \fedavg in the cross-device setting, matching the lower bound of~\cite{arjevani2019lower}.

\subsubsection{Gaps Between Theory and Practice}
In this subsection we discuss some of the pressing gaps between theoretical analyses of federated optimization algorithms, and practical algorithmic performance. We note that this is a critical topic, even in non-federated machine learning. There are a number of works dedicated to bridging the gap between the theory and practice of machine learning, especially deep learning, often focusing on topics such as generalization and the empirical success of optimization methods in non-convex settings (see ~\citep{jin2017escape, belkin2019reconciling, soltanolkotabi2018theoretical, chatterji2020does} for a few examples of such work, which we do not cover in this work).

We focus on gaps between theory and practice that are endemic to, or at least exacerbated by, federated learning. We stress that federated learning can be difficult to analyze theoretically due to its heterogeneous nature. Therefore, the topics discussed below are \emph{aspirational}, and should be viewed as significant open problems in the field. Moreover, this is not a complete list of open problems; some of the topics in \cref{sec:practical_algorithm_design} for practical consideration can become restrictions for theory. For example, algorithms that do not maintain persistent state and have low cost computation on clients are preferred for cross-device FL. For more details on such open problems in this vein, see \citep[Section 3]{kairouz2019advances}.

\paragraph{Asymptotic versus communication-limited performance}

One important caveat to many of the convergence rates discussed above is that they are \emph{asymptotic}, and may require the number of communication rounds $T$ to be sufficiently large in order to become valid. Such asymptotic results can belie the practical importance of an algorithm in settings where $T$ is small. This communication-limited nature was part of the motivation given by \citet{mcmahan17fedavg} for the development of \fedavg. As \citet{mcmahan17fedavg} show, \fedavg can lead to drastic improvements over algorithms such as \fedsgd (another name for large-batch synchronous SGD discussed in \Cref{sec:fedavg-proof}) in regimes where $T$ is limited, despite the fact that \fedavg and \fedsgd often have comparable asymptotic convergence rates~\citep{woodworth2020minibatch}. Empirical examples of how communication-limited regimes can alter the relative performance of federated algorithms were given in Section \ref{sec:evaluation}. Recent work suggests that empirically successful federated optimization algorithms such as \fedavg and \fedprox can quickly arrive in neighborhoods of critical points, but may not actually converge to such critical points~\citep{malinovskiy2020local, charles2021convergence}. It remains an open question how to theoretically quantify the communication-limited benefits of algorithms such as \fedavg, especially in non-convex settings.

\paragraph{Biased sampling of clients}

Another common facet of many theoretical analyses of federated learning is the assumption that we can sample clients in an unbiased or close to unbiased fashion. As discussed by \citet{bonawitz19sysml} and \citet{eichner2019semi}, practical cross-device federated learning usually must contend with non-uniform client availability. Whether clients are available to participate in a round depends on a number of factors (such as geographic location of the device, in the case of mobile phones). This presents significant challenges empirically and theoretically. Two outstanding challenges are understanding what types of biased sampling still result in good performance of existing algorithms \citep[e.g.][]{cho2020client,chen2020optimal}, and how to develop algorithms that can better contend with biased client sampling. While this is related to work on shuffling data in centralized learning (see ~\citep{safran2020good} for example) we stress that the degree of non-uniformity in the client sampling may be much greater in federated learning, especially if clients locally determine their availability for training~\citep{bonawitz19sysml}.

\paragraph{Client aggregation methods}

In addition, another area where theory and practice are sometimes not aligned is in the client aggregation methods used. Recall that in \cref{algo:generalized_fedavg}, the client updates are aggregated according to a weighted average, where client $i$ is weighted by $p_i/\sum_{i \in \activeClients^{(t)}} p_i$. The weights are \emph{time-varying} depending on $\activeClients^{(t)}$. Theoretical analyses \citep[e.g.][]{li2019convergence} often use $p_i = 1$ for all $i$ in order to reduce this aggregation to simple averaging across clients. However, as discussed by \citet{mcmahan17fedavg}, setting the weights to be in proportion to the number of examples in the local dataset of client $i$ may be more fruitful, as it allows one to recover a global loss function that is an average across all examples. The gap between this uniform aggregation and example-weighted time-varying aggregation can be especially important for extremely unbalanced datasets, where some clients may have only a few examples, while others may have thousands. Thus, deriving convergence theory that analyzes the performance of such aggregation methods in realistic settings could be a valuable area of exploration. 




\section{Privacy, Robustness, Fairness, and Personalization}\label{sec:connection_to_other_topics}

In this section, we discuss the relationship between federated optimization and other important considerations, such as privacy, robustness, fairness, and personalization. These problems are often posed as explicit or implicit constraints in the objective function used during optimization, as well as within the dynamics of federated optimization methods. We do not intend to give a comprehensive review of these topics. Our intention is to highlight how these added constraints impact federated optimization. For a more in-depth discussion of open problems, see sections 4-6 in \cite{kairouz2019advances}.

\subsection{Privacy} \label{sec:privacy}
While federated learning does not permit the server to access clients' data directly, federated optimization methods compute aggregate model updates using clients' data, and send these to the server. While these updates should not directly contain clients' data, it is inevitable that they contain downstream information about the clients' datasets. In fact, \citet{zhu2020deep} show that it is possible in some cases to reconstruct client data to some extent from model gradients.  Thus, it is critical that federated optimization methods are developed with privacy in mind. In order to aid the development of privacy-first methods, we describe two general techniques for privacy that are often used in federated optimization.

\subsubsection{Data Minimization} \label{sec:private_agg}



One class of technologies often used to strengthen privacy by reducing the surface of attack is generally referred to as data minimization. In federated learning, a default level of data minimization is achieved by ensuring that raw data is never shared with the server and that only focused, minimal updates that are intended for immediate aggregation are sent back to the server. 

In order to achieve greater privacy levels, we often use additional data minimization techniques. One such technique is secure computation. Secure computation provides cryptographic primitives that enable federated aggregation without revealing raw inputs to untrusted parties. \cite{bonawitz2017practical, bell2020secagg} propose secure aggregation (\secagg), which enables computing secure federated sums over client inputs. This permits the calculation of (weighted) averages of client updates without revealing individual updates to the server.

While \secagg enforces a greater level of privacy, it also imposes constraints and limitations that must be accounted for in the development of \secagg-compatible optimization methods. One notable constraint is that \secagg operates on the level of federated sums (\ie sums of client updates). Algorithms that require aggregations that cannot be performed through federated sums are therefore incompatible. While potentially restrictive, this constraint opens up the door to novel research on how to develop \secagg-compatible versions of federated optimization methods. For example, \citet{pillutla2019robust} shows that median-based aggregation can be well-approximated by a small number of federated sums (on the same set of clients),
and \citet{vogels2019powersgd, karimireddy2020learning} design new communication compression and robust aggregation protocols specifically compatible with \secagg. 
It remains an open question how to approximate other aggregation strategies (such as model distillation and fusion~\citep{lin2020ensemble}) using only federated sums, and how to implement practical secure computation method beyond summation for complicated aggregation methods.


\subsubsection{Data Anonymization} 

Another class of technologies used to strengthen privacy is data anonyimization. Such methods seek to introduce plausible deniability into the output of an algorithm. One such method is \emph{differential privacy}. Differential privacy is notable for its rigorous nature and well-tested formalization of the release of information derived from sensitive data. Informally, a differentially private mechanism is a randomized algorithm for a database query with a quantifiable guarantee that it will be difficult for an adversary to discern whether a particular item is present in the data, given the output of the query and arbitrary side information. The privacy guarantee is quantified by two values, $\epsilon$ and $\delta$, with smaller values implying increased privacy. In a machine learning context, the ``query'' is an algorithm (such as \sgd) that operates on training data and outputs a model. In centralized settings, many differentially private learning algorithms define the unit of privacy to be a single training example. However, in federated settings it is often preferable to guarantee privacy with respect to each users' entire data collection~\citep{mcmahan18learning}---an adversary should not be able to discern whether a user's dataset was used for training, or anything about it.

In the context of cross-silo FL, the unit of privacy can take on a different meaning. For example, it is possible to define a privacy unit as all the examples on a data silo if the participating institutions want to ensure that an adversary who has access to the model iterates or final model cannot determine whether or not a particular institution’s dataset was used in the training of that model. User-level DP can still be meaningful in cross-silo settings where each silo holds data for multiple users. However, enforcing user-level privacy may be challenging if multiple institutions have data from the same user. The conventional per data point privacy may be of interests in a lot of cross-silo settings.

In the context of cross-device FL, federated optimization provides a natural way to give this type of ``user-level'' differential privacy guarantee. A client's contribution to a training round is summarized by its update: the data that is sent from clients to server at each round. By applying privatizing transformations to these updates, we can derive user-level DP guarantees. For example, in the case of \fedavg, the client updates can be clipped (bounding their $\ell_2$ norm) and a calibrated amount of Gaussian noise can be added to the average, which is often sufficient to obscure the influence of any single client~\citep{mcmahan18learning}.



\paragraph{Central versus local privacy} Sharing the global model between clients or any external adversary could reveal sensitive information about the clients' data~\citep{chaudhuri2011differentially,abadi2016deep}. We consider two settings of privacy. The first is called the {\em central privacy} setting, where a trusted server can collect model updates from the clients, clip and aggregate the model updates, and then add noise to the aggregate before using it to update the server-side model and sharing the updated model with (possibly) untrusted parties (or clients in later federated rounds). Hence, this setting of differential privacy protects the privacy of the clients' data against external parties by relying on a trusted server. The second setting is called {\em local privacy}, where each client preserves the privacy of their own data against any (possibly) compromised entity including the central server, which could be untrusted. Here, the privacy guarantee is with respect to the server as well as other clients participating in the distributed optimization. This is achieved by clipping  and noising the model updates locally (on device) before sending the noised and clipped model updates back to the server. \\
 
 
The local privacy setting provides a stronger notion of privacy against untrusted servers; however, it suffers from poor learning performance in comparison with the central privacy setting because significantly more noise must be added to obscure individual updates~\citep{duchi2013local,kasiviswanathan2011can,kairouz2016discrete}. Fortunately, local privacy can be boosted when combined with the use of either a {\em secure shuffler} (that receives the private local updates of the participated clients and randomly permutes them prior to sending them to the untrusted server~\citep{erlingsson2020encode, ghazi2019scalable,balle2019privacy, girgis2020shuffled}) or a {\em secure aggregator} that sums the private updates of the participated clients prior to communicating the aggregate update to the untrusted server~\citep{kairouz2021distributed}. Observe that secure shuffling/aggregation plays the role of anonymization in which the server that observes the aggregate update cannot assign a certain update to a specific client, which can amplify the privacy of the local privacy model. Hence, the secure shuffler/aggregator can help us design private learning algorithms that have a reasonable learning performance without assuming the existence of a fully trusted server. These approaches are often considered distributed DP, and the mathematical privacy guarantee are expressed as central privacy bound. 

\paragraph{The impact of private training on optimization}

Many recent works have studied the impact of differentially private mechanisms on centralized optimization~\cite{bassily2014private,smith2017interaction,bassily2020stability,feldman2020private,bassily2019private,dagan2020interaction,girgis2020shuffled}. Such works are often concerned with developing better differential privacy mechanisms for model training, and for getting tight bounds on the $(\epsilon, \delta)$ privacy guarantees of such mechanisms. One particularly important method is DP-SGD~\citep{bassily2014private}, which applies \sgd, but adding properly calibrated noise to the gradients. Such work has since been extended to differentially private distributed optimization algorithms~\citep{kasiviswanathan2011can,erlingsson2020encode,girgis2020shuffled}, including differentially private federated learning~\citep{mcmahan18learning}. Much of the work in this area reveals fundamental \emph{privacy-utility trade-offs}. Generally speaking, to guarantee stronger privacy with smaller $(\epsilon, \delta)$, the utility (\eg accuracy in a recognition task) is often sacrificed to a certain level. While typically unavoidable (in formal, information-theoretic ways), one can often mitigate this trade-off and provide stronger privacy by introducing properly designed differential privacy and randomization mechanisms.

We note that developing and analyzing differentially private federated optimization methods is a relatively new area of study, and there are a number of open questions around how to combine the optimization techniques, such as those discussed in \cref{sec:representative_techniques}, with differentially private mechanisms. We list a few of the challenges below.

\paragraph{Unknown, unaddressable, and dynamic population} Providing formal $(\varepsilon, \delta)$ guarantees in the context of cross-device FL systems can be particularly challenging because the set of all eligible users is dynamic and not known in advance, and the participating users may drop out at any point in the protocol. There are few works discussed this issue: the recent work of \citet{balle2020privacy} shows that these challenges can be overcome using a novel random check-ins protocol (\ie clients randomly decide whether to and when to check in without any coordination); \citep{girgis_client-sampling_ISIT21} analyzes DP-SGD with alternate client self-sampling; the more recent work of \citep{kairouz2021practical} proposes a new online learning based DP algorithm, differentially private follow-the-regularized-leader
(DP-FTRL), that has privacy-accuracy trade-offs that are competitive with DP-SGD. DP-FTRL does not require amplification via sampling or a fixed, known population. Building an end-to-end protocol that works in production FL systems is still an important open problem.

\paragraph{Variance in differential privacy methods} Conventional differentially private optimization algorithms can increase the variance of the stochastic gradient method to the order of $d$,
where $d$ is the dimension of the model.
Since modern machine learning models may have millions or even billions of parameters, this scaling with $d$ can lead to significantly less accurate models. 
This issue can become a bottleneck for federated learning under privacy constraints, making it difficult to privately learn large network with reasonable accuracy.
Some possible mitigation strategies include augmenting the learning procedure with public data that does not have any privacy concerns~\citep{bassily2020learning} or developing new private optimization methods with dimension-independent variance under certain assumptions regarding the objective function and stochastic gradients~\citep{jain2014near,kairouz2020dimension}. In practical centralized learning setting, we can reduce the variance by increasing the number of samples at each round. In federated learning, this is analogous to increasing the number of clients participating in each round. While this may be possible in some setting (such as cross-device settings with high levels of client availability) it may not be possible in all settings (such as cross-silo settings with only a small number of clients). 

\paragraph{Communication and privacy} As discussed in \cref{sec:practical_algorithm_design}, communication-efficiency is typically an important requirement for federated optimization algorithms. While various communication-efficient optimization methods are discussed in \cref{sec:system_communication}, such methods may not preserve privacy. Furthermore, addressing privacy and communication-efficiency separately may lead to suboptimal methods. This has led to a budding area of communication-efficient private optimization algorithms. For example, \citet{agarwal2018cpsgd, kairouz2021distributed} propose a private optimization method where clients only needs to send $\mathcal{O}(d)$ bits to the server at each round. \citet{girgis2020shuffled, chen2020breaking} studied  communication-efficient and privacy-preserving mean estimation with each of them giving optimal methods for various $\ell_p$ geometries. In particular,~\citet{girgis2020shuffled} proposed and studied a variant of \sgd that requires $\mathcal{O}(\log(d))$ bits per client, showing that the same privacy-utility as full precision gradient exchange can be achieved. Hence, addressing the problems of communication and privacy together can lead to learning algorithms with a significant reduction on the communication cost per client.

\subsection{Robustness} \label{sec:robustness}
Machine learning can be vulnerable to failures -- from benign data issues such as distribution shift~\citep{lakshminarayanan2017simple} to adversarial examples~\citep{athalye2018obfuscated} and data poisoning~\citep{steinhardt2017certified}. Beyond these centralized robustness concerns, federated optimization, where hundreds of millions of client devices participate in the learning process with their local data, exposes the learning system to additional threats and novel versions of existing threats -- since, unlike centralized settings, the central server has limited control of the system devices and communication channels. Therefore, robustness is essential for the trustworthy, practical deployment of federated learning.

In this section, we describe some of the possible adversarial attacks in federated learning, and survey existing techniques for defending against these attacks. We also discuss other important aspects of robustness, including robustness due to non-malicious occurrences and its relation to privacy. While federated learning introduces new vulnerabilities to inference-time attacks (see \citep[Section 5.1.4]{kairouz2019advances} for a survey), we will focus on training-time attacks in our discussions below.


\subsubsection{Goals of an Adversary}

As discussed by \citep[Section 5.1.1]{kairouz2019advances}, there are a number of goals a training-time adversary may have. They may wish to reduce the global accuracy of a model (\emph{untargeted attack}), or they may wish to alter the model's behavior on a minority of examples (\emph{targeted attacks}). For example, an adversary could attempt to make a model misclassify all green cars as birds while maintaining its accuracy on all other samples~\citep{Xie2020DBA,Sun2019CanYR,bagdasaryan2020backdoor,wang2020attack}. Due to the average performance maintenance, the presence of a targeted attack can be difficult to detect. However, recent work has shown that untargeted attacks can also be introduced in ways that are difficult to detect from a statistical viewpoint~\citep{xie20empires, baruch2019little}.

Adversaries may not always actively alter the training procedure. Instead, they may function as eavesdroppers, and attempt to glean sensitive information about clients using information available at the server or by simply using the final model. This can include reconstructing client data from model updates~\citep{wei2020framework}, or performing label-inference attacks on aggregated model updates~\citep{wang2019eavesdrop}. While such instances are important failures of robustness, they may be better addressed via the privacy-preserving techniques discussed above. We will therefore focus on training-time attacks that attempt to alter the learned model in the following.



\subsubsection{Model Poisoning and Data Poisoning Attacks}\label{sec:adversarial_attacks}
Since federated learning is typically a collaborative and iterative process, and the final model is often deployed on edge devices for inference, there are a number of ways in which an adversary may attack a federated learning system. One particularly important distinction is whether an adversary is present during the training or during inference. In training-time attacks, the adversary corrupts a subset of clients and attacks the system either by corrupting the data at the compromised clients (\emph{data poisoning attacks}), or by having the corrupt clients send spurious updates to the server during the learning process (\emph{model poisoning attacks}). See \citep[Section 5.1]{kairouz2019advances} for more a more thorough taxonomy and discussion of such attacks.

Rather than focusing on specific attacks, we discuss how data and model poisoning attacks differ, especially in federated contexts. While both can significantly mar the capabilities of a learned model, data poisoning attacks are special cases of model poisoning attacks, as the clients which have poisoned data report corrupted updates to the server. Thus, data poisoning attacks may be strictly weaker; while a single malicious client can compromise a entire learning process via model poisoning~\citep{Krum_Byz17}, significantly degrading model performance with data poisoning may require poisoning many clients~\citep{bagdasaryan2020backdoor}. While data poisoning can be strengthened with collaborative attacking~\citep{sun2020data} and defense-aware poisoning~\citep{bhagoji2019analyzing, fang2020local}, a full understanding of the difference in capabilities between data poisoning and model poisoning remains unknown. This is particularly important in federated learning. Data poisoning attacks are generally easier for an adversary to engage in, requiring only influencing the data collection process within a client, not actually circumventing the core operation of the federated learning process.

Another key notion is that data poisoning attacks are not necessarily the result of an explicit adversary. Data collection processes, especially in settings with heterogeneous clients, are often noisy and can inadvertently lead to outlier data. If data poisoning occurs before training begins, it may lead to inaccurate models, even in systems where clients train honestly, according to the specified algorithm. In contrast, model poisoning attacks are often modelled by assuming corrupt clients can send arbitrary and adversarially chosen vectors to the server throughout training. The dynamic nature of such attacks makes them significantly more challenging to safeguard against. It may be difficult to even detect such attacks in a distributed system, as the downstream failure of a model may have any number of possible explanations.



\subsubsection{Defenses Against Training-Time Attacks}\label{sec:defense}
There are a number of proposed defenses against training time attacks, which can broadly be divided into two categories: coding-theoretic defenses and statistical defenses. Coding-theoretic solutions \citep{DataSoDi_CodingByz20,Draco_ByzGD18,DataDi_ByzCD19,Detox_ByzSGD19} introduce redundancy across compute nodes in a distributed system, and do not require statistical assumptions on the data. Direct application of these methods in federated learning could require introducing redundancy across clients' local data, and may violate data minimization principles. 

Statistical solutions typically require some kind of robust aggregation to mitigate the effect of corrupt updates. For example, one popular class of statistical defenses replaces the averaging step in \serveropt by a robust variant of the average such as the geometric median~\citep{ChenSX_Byz17,pillutla2019robust}, coordinate-wise median/trimmed-mean~\citep{Yin_ByzICML18,SLSGD19},  heuristics-based~\citep{Krum_Byz17,Bulyan_Byz18}, or other robust mean estimation techniques from statistics~\citep{SuX_Byz19,Yin_ByzICML19,DataDi_Byz-SGD_Heterogeneous20,DataDi_ByzFL20}. 
Under assumptions on the distribution of the client updates and the number of corrupt clients, this can guarantee that the aggregate update is close to the true average.

Heterogeneous clients present a challenge to robust federated learning since it is hard to distinguish a client with unique data from a clever attacker~\citep{he2020byzantine}. Solutions for this setting attempt to re-use existing robust algorithms by either clustering the clients into relatively homogeneous clusters~\citep{YinRobustFL19}, or by properly resampling clients during training~\citep{he2020byzantine}, and by using gradient dissimilarity properties~\citep{DataDi_Byz-SGD_Heterogeneous20,DataDi_ByzFL20}. Such work often requires new theoretical approaches in order to contend with heterogeneous data, such as analyses incorporating non-uniform sampling~\citep{he2020byzantine} and novel concentration inequalities~\citep{DataDi_Byz-SGD_Heterogeneous20,DataDi_ByzFL20}. A possible solution might be to use trusted data at the server to filter the client updates ~\citep{xie2019zeno,xie2020zeno}. However, trusted data is often not available at the server in federated learning and so such solutions are not feasible.
Instead, the client updates can be directly utilized to build up estimates of non-outlier model updates over time, and memory-based solutions for combating adversarial attacks have been employed by~\citep{Alistarh_Byz-SGD18,karimireddy2020learning,allenzhu2020byzantineresilient}. These defenses however are currently only designed for the cross-silo setting and extending them to the cross-device setting is an important open question.
Client heterogeneity also enables targeted attacks where an attacker may manipulate only a small subset of clients' data in order to introduce backdoors~\citep{bagdasaryan2020backdoor}. Detecting and defending against such targeted attacks is challenging and remains an open problem~\citep{Sun2019CanYR,wang2020attack}.

Another challenge in FL is that of communication efficiency, and only a few works have considered this with Byzantine adversaries~\citep{Ghosh_Byz_ISIT20,DataDi_ByzFL20,SLSGD19}. Among these, \citet{DataDi_ByzFL20} works with heterogeneous data and high-dimensional model learning, and incorporates local iterations in the analysis. However, the polynomial time method in \citep{DataDi_ByzFL20} for filtering out corrupt updates is not as efficient as other cited methods. An important and interesting direction is to devise a communication efficient method that works with the challenges posed by the setting of federated learning.

\subsubsection{Other Robustness Concerns}\label{sec:other_concerns}

\paragraph{Robustness and aggregation protocols}
An existing challenge is to ensure that robust optimization is compatible with privacy primitives in federated learning such as \secagg~\citep{bonawitz2017practical}. Many robust methods rely on non-linear operations such as medians, which are often not efficient when combined with cryptographic primitives. Some emerging solutions address this by replacing median operations with iterative Weizfeld's algorithm approximation~\citep{pillutla2019robust},
or to clip the client updates before aggregation~\citep{karimireddy2020learning}. 
\citet{he2020robustsecure} uses secure two-party computation with a non-colluding auxiliary server in order to perform non-linear robust aggregation. In addition to restricting the non-linearity of defenses, aggregation protocols such as \secagg also make training-time attacks more difficult to detect, as the server cannot trace suspected adversarial updates to specific clients~\citep{bagdasaryan2020backdoor}. Therefore, when designing a new defense for federated learning, we suggest considering its compatibility with secure aggregation protocols, and to be explicit about the type of system required for a method.

\paragraph{Robustness to distribution shift} 
In heterogeneous settings where the client population evolves over time, it may be critical to train a model that is robust to distribution shift in data. In fact, in federated settings where we train a global model over some training set of clients and test its performance on a held-out set of clients, we are effectively measuring performance under some amount of distributional shift, as the two populations may be very different from one another. To this end, \citet{lin2020ensemble} uses ensembling throughout their federated training to produce more distributionally robust models. Another approach to distributional robustness involves having clients learn both a model and a ``worst-case'' affine shift of their data distribution~\citep{reisizadeh2020robust}.

\paragraph{Rate of corruption}
Many of the aforementioned works on defenses against training-time attacks provide resilience against that a constant fraction of clients are malicious (say, one-tenth of the clients are malicious). However, the effective rate of corruption may be extremely small in cross-device settings, where there may be hundreds of millions of clients, only a tiny fraction of which are actually malicious. In cross-silo settings, where there may be only a small number of clients, the rate of corruption may be higher. Given the already onerous difficulties of developing federated optimization methods that contend with heterogeneity, communication constraints, privacy, and compatibility with \secagg, we emphasize that developing methods that can also tolerate a tiny fraction of malicious clients is still an important achievement, and may be a much more tractable problem.

\subsection{Fairness}
\label{sec:fairness}
 

Algorithmic fairness is an emerging research area under machine learning that attempts to understand and mitigate the unintended or undesirable effects the learned models may have on individuals or sensitive groups (races, genders, religions, etc.) \cite{barocas2017fairness, kearns2019ethical}. For example, if individuals with similar preferences and characteristics receive substantially different outcomes, then we say that the model violates individual fairness \cite{dwork2012fairness}. If certain sensitive groups receive different patterns of outcomes, this can violate certain criteria of group fairness (e.g. demographic parity fairness or equality of opportunity fairness \cite{hardt2016equality}).

Fairness concerns are critical, and can be exacerbated, in federated learning, due to systems and data heterogeneity. While models trained via federated learning may be accurate on average, these models can sometimes perform unfairly or even catastrophically on subsets of clients~\citep{li2019fair}. For instance, the unfairness/bias could come from periodic changes of data patterns, over-represented data due to the large size of samples on some specific devices, or under-represented users who do not own devices~\citep{kairouz2019advances}.
 It is thus important to produce models that go beyond average accuracy to also ensure fair performance, e.g., maintaining a minimum quality of service for all devices.

This notion of fairness (i.e., uniformity of model performance distribution), also known as representation disparity~\citep{hashimoto2018fairness}, poses additional challenges for federated optimization. Some works propose min-max optimization~\citep{mohri2019agnostic,deng2020distributionally} to optimize the model performance under the worst case data distributions. As solving a min-max problem in federated settings can be especially challenging due to the scale of the network, other works propose alternative objectives to reweight samples less aggressively, allowing for a more flexible tradeoff between fairness and accuracy~\citep{li2019fair, li2020tilted}. Fairness can also be enforced by optimizing the dual formulation of min-max robust optimization, e.g., via superquantile methods or conditional Value-at-Risk which only minimizes the losses larger than a given threshold~\citep{li2020tilted, laguel2020device}.
 Different from the motivation described above, there are also other works considering varying notions of fairness (e.g., proportional fairness~\citep{zhang2020hierarchically}) for federated learning. We note that it is important for any fair federated learning algorithms to be resilient to heterogeneous hardware by allowing for low device participation and local updating, which are basic desirable properties for federated optimization methods, especially for cross-device settings where fairness concerns are prominent.
 
While not originally motivated by the problem of fairness, another line of work that can mitigate unfairness/sampling bias is alternate client selection strategies (as opposed to random selection)~\citep{cho2020client,chen2020optimal,k-fed}. Intuitively, selecting representative devices could produce more informative model updates, thus helping speed up convergence and encouraging a more uniform performance distribution across all devices. However, discovering the underlying structure/similarities of the federated networks may be prohibitively expensive. Therefore, we suggest that any client selection methods need to account for and be compatible with practical systems and privacy constraints of FL~\citep{bonawitz19sysml}.

There are still many open problems for fairness in FL. For instance, it would be interesting to explore the connections between the existing FL fairness notions and the broader fairness literature. It is also worth investigating the intersection between fairness and other constraints like robustness and privacy, as discussed in the next section.

 
\subsection{Tensions Between Privacy, Robustness, and Fairness}
\label{sec:tensions}
It is also critical to consider the interplay and tension between fairness, robustness, and privacy. Fairness constraints attempt to ensure that the learned model works well for clients with data distributions that are not particularly well aligned with the average population distribution. Robustness constraints attempt to ensure that the output of a learning algorithm is not affected by outlier data. Privacy constraints attempt to ensure that the learned model and intermediate model updates do not overfit to or retain information about outlier client training examples. Intuitively, these constraints can potentially lead to conflicting objectives.

For example, training with differential privacy requires bounding the clients' model updates and adding noise to the model in each round, therefore preventing the model from memorizing unique training examples. Though this often helps with robustness \cite{Sun2019CanYR}, it leads to disparate impact on accuracy and therefore unfairness \cite{bagdasaryan19disparate}. The use of secure aggregation to hide client updates strengthens privacy but often makes it impossible for the server to use robust optimization algorithms (e.g. computing a high dimensional median or a trimmed mean which require access to the individual model updates). Worse still, fairness and robustness are difficult to maintain while simultaneously ensuring privacy, as user privacy constraints limit the ability of a central provider to examine, test, and validate training data. On the other hand, robustness to distributional shift in user data distributions (discussed in \cref{sec:robustness}) can be thought of as a generalized notion of fairness. 

It is an interesting area of future research to optimize for fairness in federated learning while taking into account robustness and privacy constraints. As we will see in the next subsection, personalization might offer a good solution to break the tension between these constraints~\citep{ditto}. 

\subsection{Personalization}\label{sec:personalization}


As discussed in Section~\ref{sec:intro}, a defining characteristic of federated learning is that client data is likely to be heterogeneous. One natural solution is to provide models that are \textit{personalized} to each client. To enable personalization, a simple approach is to incorporate client-specific features. These features may be naturally occurring in the data, or may take the form of some auxiliary meta data. However, in lieu of (or in addition to) including such expressive features, we can use federated optimization methods to learn personalized models for clients. For example,  \citet{li2019fedmd} propose an algorithm that allows clients to have different model architectures via transfer learning, provided the server has access to a representative public dataset. However, for brevity, in this section we focus on the case when the personalized models have the same architecture.

The idea that every client gets a good model for its own data not only improves the overall statistical performance, but also potentially improve fairness or robustness of private algorithms~\citep{ditto, yu2020salvaging}. As alluded to in Section~\ref{sec:tensions}, using personalization to mitigate tension between privacy, robustness, and fairness is an active research area in federated learning. Before discussing the specific personalization algorithms (e.g., multi-task learning, clustering, fine-tuning, and meta-learning), we give two general categories of personalization algorithms:
\begin{itemize}
    \item \textbf{Algorithms that require client-side state or identifier:} Here ``client-side state'' means that every client needs to maintain some state variables locally during training, and is responsible for carrying these variables from the previous round and updating them in the next round. As pointed out in Section~\ref{section:application_patterns}, the assumption of stateful clients is more appropriate in the cross-silo setting, where the number of total clients is relatively small, and most (if not all) clients can participate in every round of training. Algorithms in this category sometimes also assume that every client has an identifier known by the server, and that the server can store updates from individual clients. This assumption can incur extra privacy concerns because it may be difficult to obtain strong privacy guarantees under this setting (see Section~\ref{sec:privacy} for more discussions on privacy considerations in federated learning).
    \item \textbf{Algorithms that do not require client-side state or identifier:} In contrast to the first category, algorithms in this category do not require the server to know the client's identifier; the clients also do not need to carry a state from the previous round to the next round. This makes these algorithms more attractive in the cross-device setting, where the population size is huge (e.g., millions of devices), only a small number of clients (e.g., a few hundreds) participate in each round, and a device usually participates only once during the entire training process. Furthermore, it is often straightforward to combine these algorithms with the privacy techniques discussed in Section~\ref{sec:privacy}.
\end{itemize}

\subsubsection{Algorithms that Require Client-side State or Identifier}\label{sec:stateful_p13n_algo}

A popular technique used in this category is \textbf{multi-task learning}\footnote{Note that not all algorithms developed under the multi-task learning framework fall into this category, as shown in Section~\ref{sec:stateless_p13n_algo}.}. To model the (possibly) varying data distributions on each client, it is natural to consider learning a separate model for each client's local dataset. If we view learning from the local data on each client (or possibly a group of clients) as a separate ``task'', we can naturally cast such a problem as an instance of multi-task learning. In multi-task learning, the goal is to learn models for multiple related tasks simultaneously. The ERM objective function usually takes the following form:
\begin{equation}
    \min_{\vx_1, \vx_2, \hdots, \vx_{\numClients} \in R^{\modelSize}} \sum_{i=1}^\numClients  f_i(\vx_i) + \phi(\vx_1, \vx_2, \hdots, \vx_{\numClients}), \label{eqn:multitask_obj_erm}
\end{equation}
where $\vx_i$ and $f_i(\cdot)$ denote the local model and loss function for the $i$-th client, and $\phi(\cdot)$ is a regularization term capturing the relations between the clients. 

\citet{smith2017federated} first proposed a primal-dual multi-task learning method for jointly learning the optimal local models and the relationships among these models in the federated setting. A similar formulation is considered in~\citep{FL-personal-mixture2020, personalized-optimal-2020, huang2021personalized, ditto}, where the key differences are their regularization terms. The regularization term in~\citep{FL-personal-mixture2020, personalized-optimal-2020, ditto} is the distance between the local models and the global model, while \citet{huang2021personalized} uses the sum of pairwise functions to capture the relations among the clients. The algorithms proposed by~\citep{smith2017federated, FL-personal-mixture2020, ditto, personalized-optimal-2020} require the clients to maintain certain local states and update them in every round, such as the local dual variables~\citep{smith2017federated} or the local models~\citep{FL-personal-mixture2020, personalized-optimal-2020, ditto}. The algorithms developed in~\citep{smith2017federated, huang2021personalized} also require the server to store individual updates from the clients. \citet{sattler2020clustered} also considered a multi-task learning framework, and proposed to use {clustering} in the federated setting. The algorithm developed in \citep{sattler2020clustered} requires the server to know the clients' identifiers so that it can compute the pairwise similarities between the clients and then partition them into different clusters.

Another common approach, which can be viewed as a special case of the multi-task learning framework, is to split the entire model into two parts: a shared part that is jointly learned by all clients, and a local part that is personalized to each client. Examples include~\citep{arivazhagan2019federated, liang2020think,deng2020adaptive, PFL_universal2021}, where their algorithms require that every client stores the local part of the model as the local state, which is updated every round. \citet{agarwal2020federated} considers an online-learning scenario and assumes that the client can observe new samples in every round. However, the algorithm proposed by~\citet{agarwal2020federated} requires the server to store the local predictions from individual clients, which might violate the data privacy constraint required by many federated learning applications.

\subsubsection{Algorithms that Do Not Require Client-side State or Identifier}\label{sec:stateless_p13n_algo}

A popular technique used in this category is \textbf{meta-learning}. The goal of meta-learning is to learn a learning algorithm from a variety of tasks, so that this learning algorithm can solve new tasks using a small number of training samples. In the federated setting, every client can be treated as a different task, and the goal is to meta-learn a learning algorithm that can generalize to unseen clients. A typical objective function is as follows:
\begin{equation}
    \min_{\vx} \sum_{i=1}^{\numClients} f_i(g_i(\vx)),
\end{equation}
where $g_i(\cdot)$ represents a learning algorithm that produces the local model on the $i$-th client (i.e., the local personalization algorithm), and $\vx$ denotes the parameters that configure the learning algorithm. Note that meta-learning and \textbf{multi-task learning} are closely related. In fact, meta-learning is a popular approach for developing algorithms for solving multi-task learning problems.


\citet{fallah2020personalized} proposed an algorithm under the MAML (model-agnostic meta-learning) framework~\citep{finn2017model}, where $g_i(\vx) = \vx - \alpha \nabla f_i(\vx)$ denotes a learning algorithm that performs one step of gradient descent starting from $\vx$. Exactly computing the gradient of the MAML objective is usually computationally inefficient, so first-order approximations such as Reptile~\citep{nichol2018first} might be preferred. On the other hand, MAML may have lower sample complexity than Reptile and as a result works much better when the number of examples per task is low~\citep{alshedivat2021deml}.
\citet{Qupel_personalized20} proposed QuPeL, a \textit{quantized} and \textit{personalized} FL algorithm, that facilitates collective training with heterogeneous clients while respecting resource diversity. For personalization, they allow clients to learn \textit{compressed personalized models} with {\em different} quantization parameters depending on their resources. 
The authors develop an alternating proximal gradient update for solving their quantized personalization problem and analyzed its convergence properties. QuPeL also outperforms the other competing methods in personalized learning.
\citet{khodak2019adaptive} showed that the \fedavg algorithm~\citep{mcmahan17fedavg}, followed by fine-tuning the global model on new client's local data during the inference time, is closely related to the Reptile algorithm. Specifically, \citet{khodak2019adaptive} proposed an online meta-learning algorithm, which was later generalized by \citet{li2019differentially} with privacy guarantees. Similar connections were later pointed out by~\citet{jiang2019improving}.
Despite being the simplest possible approach to personalization, local fine-tuning is shown to work well in large-scale datasets~\citep{yu2020salvaging} as well as real-world on-device applications~\citep{wang2019federated}. Besides local fine-tuning, another popular local adaptation algorithm is to interpolating between local and global models~\citep{mansour2020three,dinh2020personalized,shen2020federated}. Splitting the entire model into a shared part and a local part is another natural approach to personalization. Unlike the stateful algorithms mentioned in Section~\ref{sec:stateful_p13n_algo}, \citet{singhal2021federated} proposed an algorithm where the local part is reconstructed locally each time a client participates in a round, and showed that the proposed algorithm also fits in the meta-learning framework.

Besides meta-learning,  \textbf{clustering} is also a popular approach to personalization. The idea is to cluster the clients and learn a single model for each cluster. Unlike the clustering algorithm~\citep{sattler2020clustered} mentioned in Section~\ref{sec:stateful_p13n_algo}, the algorithms developed by~\citep{mansour2020three, ghosh2020efficient} do not require client-side state or identifier, and they work as follows: in each round, the server sends all models (one for each cluster) to a set of clients; each client, after receiving the models, first identifies which cluster it belongs to and then computes updates for that specific model. Besides iterative methods,  \citet{k-fed} propose a one-shot federated clustering technique which leverages data heterogeneity to weaken the separation assumptions for clustering. It can be used as a light-weight pre-processing step with applications to personalization and client selection.



\section{Concluding Remarks} \label{sec:conclusion}

Federated learning is an active and interdisciplinary research area, in which many challenging problems lie at the intersections of machine learning, optimization, privacy, security, cryptography, information theory, distributed systems and many other areas.  Research on federated optimization centers around formulating and solving the optimization problems arising in federated learning settings. This manuscript is intended to provide concrete recommendations and guidance for practitioners and researchers who are new to the field on how to design and evaluate federated optimization algorithms (see \cref{sec:intro,sec:formulation,sec:practical_algorithm_design,sec:evaluation}). Moreover, this manuscript extensively discussed the practical considerations as constraints for designing federated optimization algorithms, and briefly touched system deployment in \cref{sec:system}. A concise yet self-contained discussion of the basic theory analysis is provided in \cref{sec:theory} to help readers get familiar with the theoretical concepts and gain insights. Some elementary introductions to the broader scope of FL research are also given to help researchers understand how federated optimization intersects with other important considerations in FL and why certain recommendations are made (see \cref{sec:connection_to_other_topics}). 

This manuscript places a greater emphasis on guidelines for the practical implementation of federated optimization methods than most of previous works that share a similar spirit. Federated optimization algorithms are often complicated because of the natural partitioning of server/client computation and the various constraints imposed by system design and privacy considerations. The evaluation of FL methods may require extra caution regarding train/validation/test set splits, metrics and hyperparameters. We extensively use generalized FedAvg-like algorithms to showcase the suggested evaluation principles, rather than directly comparing the performance of different algorithms to recommend a specific set of algorithms. These evaluation principles are not intended to be a comprehensive checklist for readers to strictly follow, but are used to inspire researchers with ideas for how to perform their own evaluations. Among the suggestions, we want to emphasize the importance of specifying the application settings and comparison under the same set of constraints. Understanding the scenarios where an algorithm can be applied and has advantages can be particularly interesting for practitioners.  However, as deploying algorithms in a real-world FL system can be a privilege not accessible to many FL researchers, the experimental evaluations mostly depend on carefully designed simulations. This suggestion can be universal for the general machine learning research, but particularly important for federated learning where the simulation experiments and system deployment environments can be very different. 

FL applications are often categorized as either cross-device or cross-silo settings. Much discussion in this manuscript is biased towards cross-device settings for several reasons: cross-device settings usually impose more strict constraints (like local computation, communication efficiency, client availability etc), which makes the practical consideration different from conventional distributed optimization;  and, many of the authors are more familiar with the cross-device settings in practice. Particularly, the real-world system deployment discussed in \cref{sec:system} is primarily based on the Google FL system \citep{bonawitz19sysml}. We hope that most of the suggestions can also be useful in cross-silo settings. However, cross-silo FL may impose a different set of system and privacy constraints that empower researchers to design sophisticated algorithms beyond the scope of this manuscript. 

Previous federated optimization research extensively discussed communication constraints, data heterogeneity, and to some extent computational constraints. Though many open problems and challenges still exist in these areas, as discussed in \cref{sec:federated_optimization}, we believe privacy, security and system issues should be considered equally important factors. Privacy and system consideration often introduce constraints that are challenging for designing optimization algorithms, and a co-design of optimization and these factors can be important. 

Optimization remains one of the key and active research areas in federated learning. Though this manuscript is probably too long to be considered concise, we hope the extended discussions provide clear pointers and won’t interfere with the main messages. We sincerely hope that the guidelines discussed can be useful for both researchers and practitioners to design and apply their federated learning algorithms.

\section*{Acknowledgements and Notes}
\label{sec:ack_notes}
The authors thank the early feedback and review by Sean Augenstein, Kallista Bonawitz, Corinna Cortes, and Keith Rush.
This work is supported by GCP credits provided by Google Cloud. The simulation experiments are implemented with the TensorFlow Federated package. This paper originated at the discussion session moderated by Zheng Xu and Gauri Joshi at the Workshop on Federated Learning and Analytics, virtually held June 29–30th, 2020. During the discussion, a general consensus about the need for a guide about federated optimization is reached.

\paragraph{Editors} We thank the extra work of section editors for helping incorporating contributions, organizing contents and addressing concerns. This is not a complete list of contributors or section authors. The editors\footnote{Zheng Xu  \{\url{xuzheng@google.com}\} for \cref{sec:intro,sec:conclusion}, Gauri Joshi \{\url{gaurij@andrew.cmu.edu}\} for \cref{sec:formulation}, Jianyu Wang \{\url{jianyuw1@andrew.cmu.edu}\} for \cref{sec:practical_algorithm_design}; Zachary Charles \{\url{zachcharles@google.com}\} for \cref{sec:evaluation}; Zachary Garrett \{\url{zachgarrett@google.com}\} for \cref{sec:system}, Manzil Zaheer  \{\url{manzilzaheer@google.com}\}  for \cref{sec:theory}, and Peter Kairouz \{\url{kairouz@google.com}\} for \cref{sec:connection_to_other_topics}.} would be happy to field questions from interested readers, or connect readers with the authors most familiar with a particular section.
Zheng Xu edited \cref{sec:intro,sec:conclusion} with the help of Jakub Kone\v{c}n\'{y} and Brendan McMahan; Gauri Joshi edited \cref{sec:formulation} with the help of Brendan McMahan, Jianyu Wang, and Zheng Xu; Jianyu Wang edited \cref{sec:practical_algorithm_design} with the help of Gauri Joshi, Zachary Charles, and Satyen Kale; Zachary Charles edited \cref{sec:evaluation} with the help of Brendan McMahan and Sebastian Stich; Zachary Garrett edited \cref{sec:system} with the help of Zachary Charles and Zheng Xu; Manzil Zaheer edited \cref{sec:theory} with the help of Jianyu Wang and Honglin Yuan; Peter Kairouz edited \cref{sec:connection_to_other_topics} with the help of Galen Andrew, Tian Li, Zachary Charles, Shanshan Wu, and Deepesh Data. 

\paragraph{Experiments}
Zachary Charles, Jianyu Wang, and Zheng Xu designed and coordinated the simulation experiments in \cref{sec:evaluation}, and provided necessary infrastructure. Zachary Charles, Weikang Song and Shanshan Wu worked on the GLD-* MobileNetV2 experiments; Zhouyuan Huo, Zheng Xu and Chen Zhu worked on the StackOverflow transformer experiments; Advait Gadhikar, Luyang Liu and Jianyu Wang worked on the CIFAR-10 ResNet cross-silo  experiments.  Hubert Eichner, Katharine Daly, Brendan McMahan, Zachary Garrett,  Weikang Song, and Zheng Xu designed and collected data for the basic model in \cref{sec:basicmodel}.

\bibliographystyle{plainnat}
\bibliography{refer}

\appendix

\section{Datasets}\label{appendix:datasets}

\subsection{GLD-23k and GLD-160k Datasets}

\begin{figure}[ht]
    \centering
    \includegraphics[width=\textwidth]{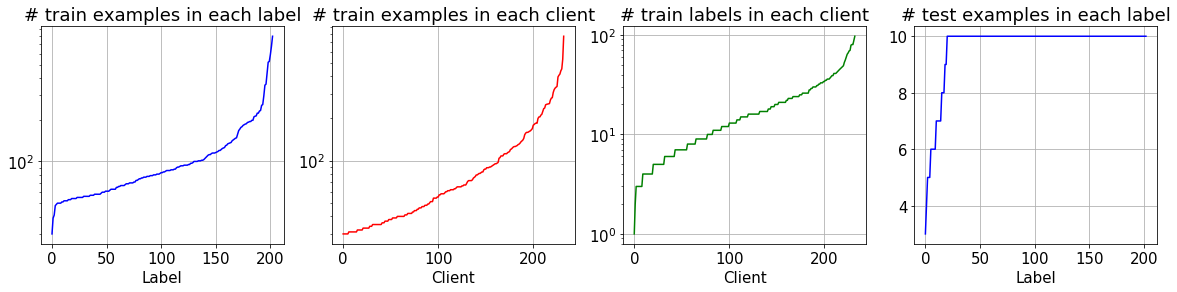}
    \caption{Statistics of the GLD-23k dataset.}
    \label{fig:gld23k}
\end{figure}

The GLD-160k dataset is proposed by~\citet{hsu2020fedvision} (under the name Landmarks-User-160k). It is a federated version of the centralized Google Landmark Dataset~\citep{noh2017large}. Each client represents an author, and the images are partitioned into clients based on their original authorship. GLD-160k contains 164,172 training images, 2028 landmark labels, and 1,262 clients. It has a single test set (i.e., the test dataset is not federated) containing 19,526 images. The GLD-23k dataset is a subset of the GLD-160k dataset. GLD-23k contains 23,080 training images, 203 landmark labels, and 233 clients. Same as GLD-160k, GLD-23k has a single test set containing 1,959 test examples, which covers all the labels in the federated training set (see the rightmost plot of Figure~\ref{fig:gld23k}). Though smaller, the GLD-23k dataset maintains the imbalanced training characteristics of the GLD-160k dataset~\citep{hsu2020fedvision}. As shown in the left three plots of Figure~\ref{fig:gld23k}, the number of training examples per label and the number of training examples per client both can vary for an order of magnitude, and the number of training labels in each client can vary for up to two orders of magnitude. These characteristics make the GLD-23k dataset representative of some issues endemic to real-world federated learning problems, and the smaller size makes it easier to work with in simulations.

\subsection{Stack Overflow Dataset}
Stack Overflow is a language dataset consisting of question and answers from the Stack Overflow online forum. The questions and answers have associated metadata, including tags (\eg ``javascript''), the creation time of the post, the title of the associated question, the score of the question, and the type of post (\ie whether it is a question or an answer). Each client corresponds to a user, whose examples are all of their posts. We use the version from~\citep{stackoverflow}, which partitions the dataset among training, validation, and test clients. There are 342,477 train clients, 38,758 validation clients, and 204,088 test clients. Notably, the train clients only have examples from before 2018-01-01 UTC, while the test clients only have examples from after 2018-01-01 UTC. The validation clients have examples with no date restrictions, and all validation examples are held-out from both the test and train sets. The dataset is relatively unbalanced, with some users having very few posts, and other having many more (one client has over 80,000 examples).

\subsection{CIFAR-10 Dataset}
The CIFAR-10 dataset~\citep{krizhevsky2009learning} is a computer vision dataset consisting of $32 \times 32 \times 3$ images with 10 possible labels. There are 50,000 training examples and 10,000 test examples, for a total of 6000 images per label. While this dataset does not have a natural partition among clients, \citet{hsu2019measuring} propose a technique for creating non-IID partitions of the dataset across clients. Each client draws a multinomial distribution over the 10 labels from an underlying symmetric Dirichlet distribution with parameter $\alpha$. To assign the client examples, we draw labels from this multinomial distribution, and sample (without replacement) one of the images in CIFAR-10 with the corresponding label. We proceed until the entire dataset has been exhausted. As discussed by \citet{hsu2019measuring}, this recovers a purely IID split when $\alpha \to \infty$, while for $\alpha = 0$ each client draws examples with a single label. We apply this technique with a Dirichlet parameter of $\alpha = 1$ to partition the training set of CIFAR-10 across 10 clients, each with 5000 images.

\section{Empirical Evaluation - Details}\label{appendix:empirical_details}

\subsection{Algorithms}
In \cref{sec:evaluation}, we refer to three algorithms, Algorithm A, B, and C. The purpose of this name obfuscation was to avoid making comparisons between algorithmic efficacy, as the purpose of that section was to provide suggestions for \emph{how} to evaluate algorithms. For the interested reader, these algorithms are the following:
\begin{itemize}
    \item Algorithm A: \fedavg~\citep{mcmahan17fedavg}
    \item Algorithm B: \fedavgm~\citep{hsu2019measuring} 
    \item Algorithm C: \fedadam~\citep{reddi2021adaptive}
\end{itemize}

All three algorithms are special cases of the generalized \fedopt framework proposed by \citet{reddi2021adaptive} and presented in \cref{algo:generalized_fedavg}. All three algorithms use \sgd as the client optimizer (\clientopt in \cref{algo:generalized_fedavg}), but use different server optimizers (\serveropt in \cref{algo:generalized_fedavg}). Specifically, \fedavg, \fedavgm, and \fedadam use \sgd, \sgd with momentum, and \adam~\citep{kingma2014adam}. Note that all three methods therefore perform the same amount of client computation and communication at each round.

In our experimental implementations, we only vary the client learning rate $\eta$ and server learning rate $\eta_s$. For \fedavgm, we use a momentum parameter of $0.9$. For \fedadam, we use server \adam with parameters $\beta_1 = 0.9$, $\beta_2 = 0.99, \epsilon = 0.001$. These are used for first-moment momentum, second-moment momentum, and numerical stability, respectively (see \citep{kingma2014adam} for more details).

\subsection{Models}
In this section, we give a brief summary of the models used for each task in \cref{sec:evaluation}. We use one model per dataset.

\paragraph{CIFAR-10} For CIFAR-10 experiments, we train a modified ResNet-18 model, where the batch normalization layers are replaced by group normalization layers~\citep{wu2018group}. We use two groups in each group normalization layer. This replacement of batch normalization by group normalization was first proposed by \citet{hsieh2020non}, who show that it can lead to significant gains in heterogeneous federated settings.

\paragraph{GLD-23k and GLD-160k} For both GLD-23k and GLD-160k experiments, we use a modified version of the MobileNetV2 model~\citep{sandler2018mobilenetv2}. As with CIFAR-10, we replace all batch normalization layers with group normalization layers~\citep{wu2018group}. Each group normalization layer has two groups. We also do not use the dropout layers (setting their dropout probability to $0$).

\paragraph{Stack Overflow} For Stack Overflow experiments, we train a modified 3-layer Transformer model~\citep{vaswani2017transformer}, where the dimension of the token embeddings is 96, and the hidden dimension of the feed-forward network (FFN) block is 1536. We use 8 heads for the multi-head attention, where each head is based on 12-dimensional (query, key, value) vectors. We use ReLU activation and set dropout rate to 0.1.

\section{Additional Experimental Results} \label{appendix:additional_results}

\subsection{More results for \cref{subsec:evaluation_guide}}

\begin{figure}[ht]
\centering
\begin{subfigure}
    \centering
    \includegraphics[width=0.32\linewidth]{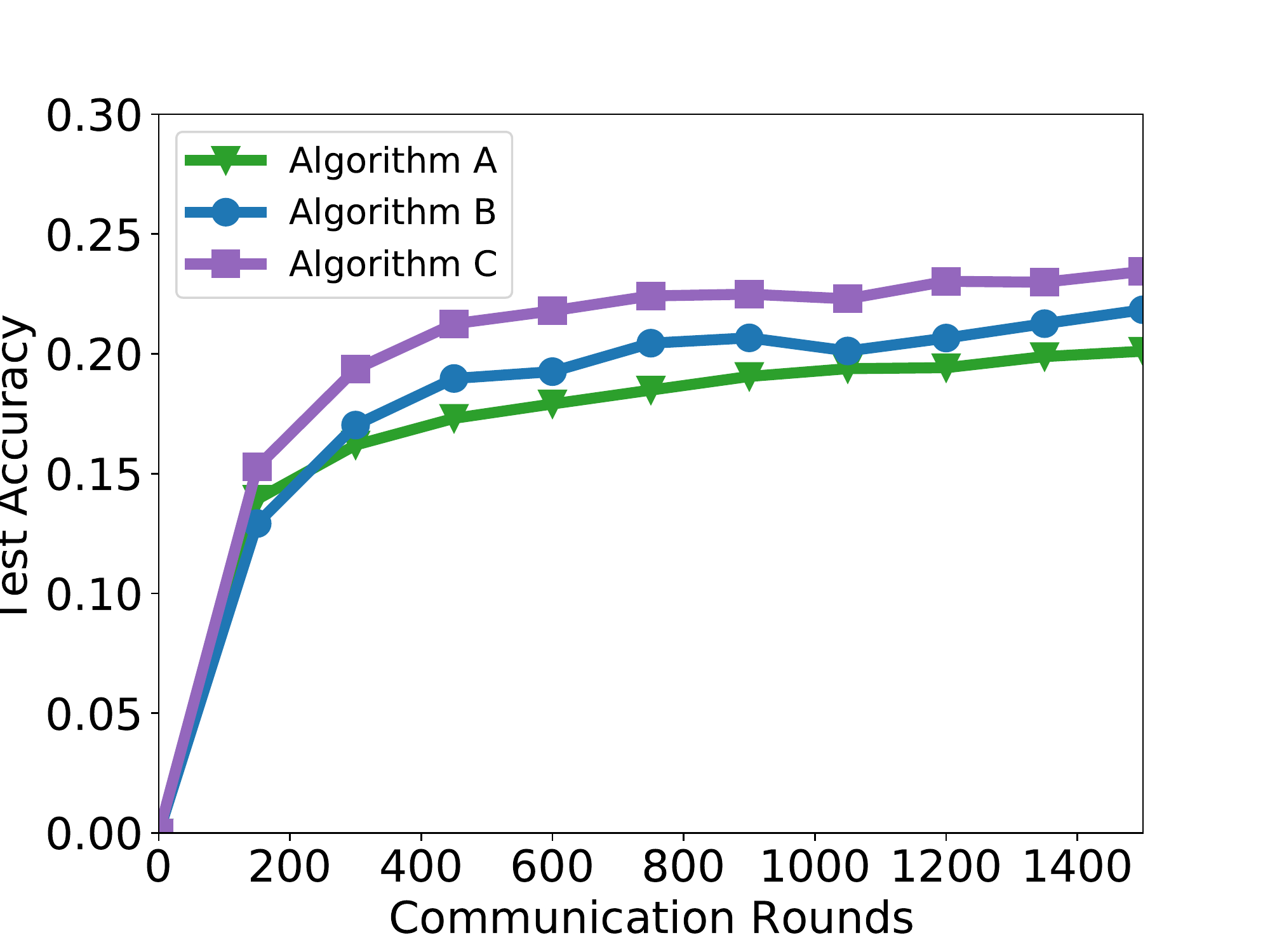}
\end{subfigure}
\begin{subfigure}
    \centering
    \includegraphics[width=0.32\linewidth]{images/stackoverflow/stackoverflow_compare_algorithms_num_rounds_validation_tuning_v1.pdf}
\end{subfigure}
\begin{subfigure}
    \centering
    \includegraphics[width=0.32\linewidth]{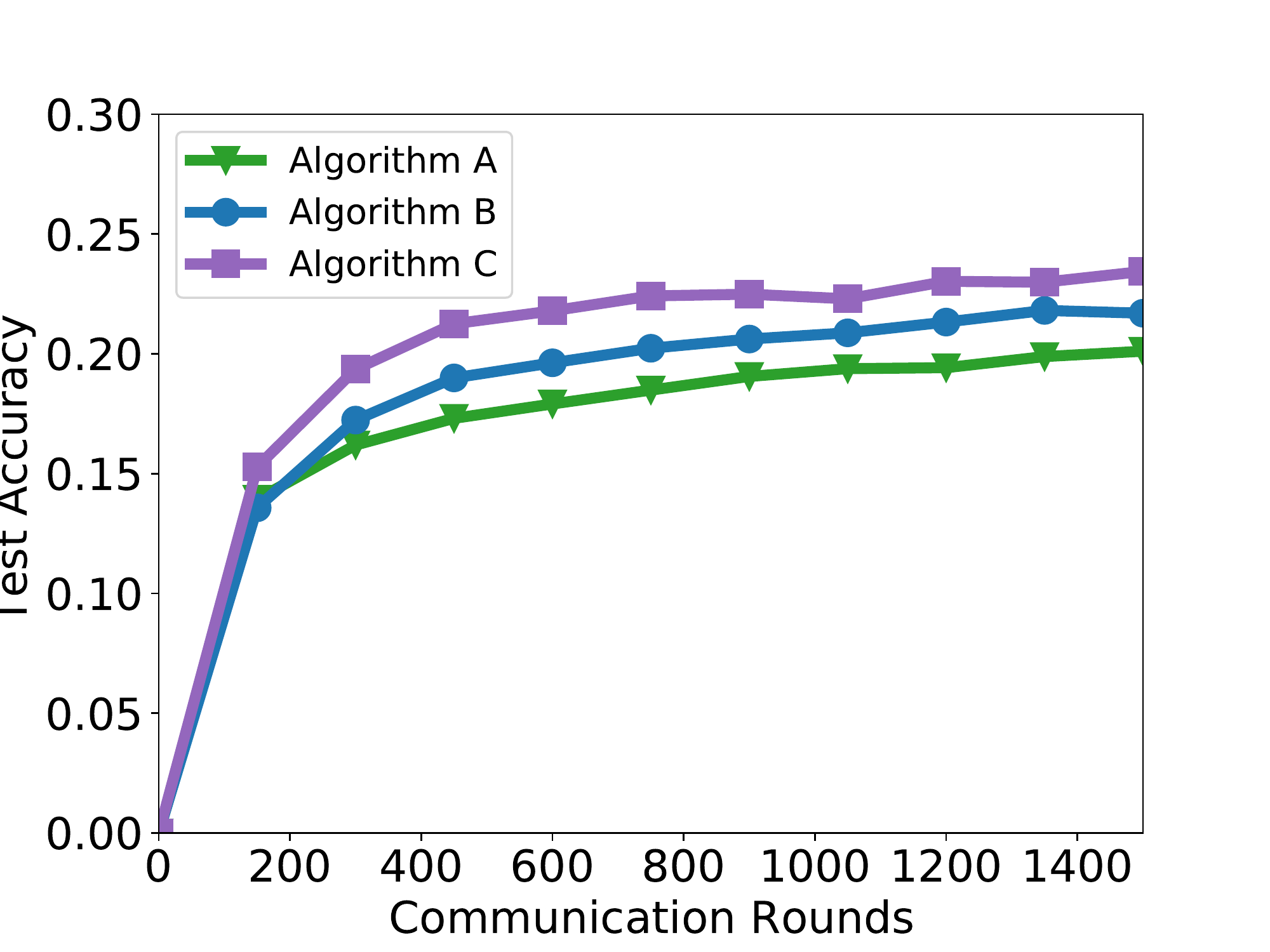}
\end{subfigure}
\caption{Test accuracy on Stack Overflow for Algorithms A, B, and C. The best performing client and server learning rate combinations are selected for each algorithm based on (a) training (b) validation (also see \cref{fig:stackoverflow_compare_tuning}) (c) testing accuracy. The trend of the curves are similar. For large-scale cross-device dataset, selecting by training and selecting by validation can be closely related because the probability of a client being sampled multiple times can be low. Selecting by testing can show the capacity (upper bound) of an algorithm, but is not a practical tuning strategy. More discussion in \cref{subsec:tuning strategy}. }
\end{figure}


\begin{figure}[ht]
\centering
\begin{subfigure}
    \centering
    \includegraphics[width=0.3\linewidth]{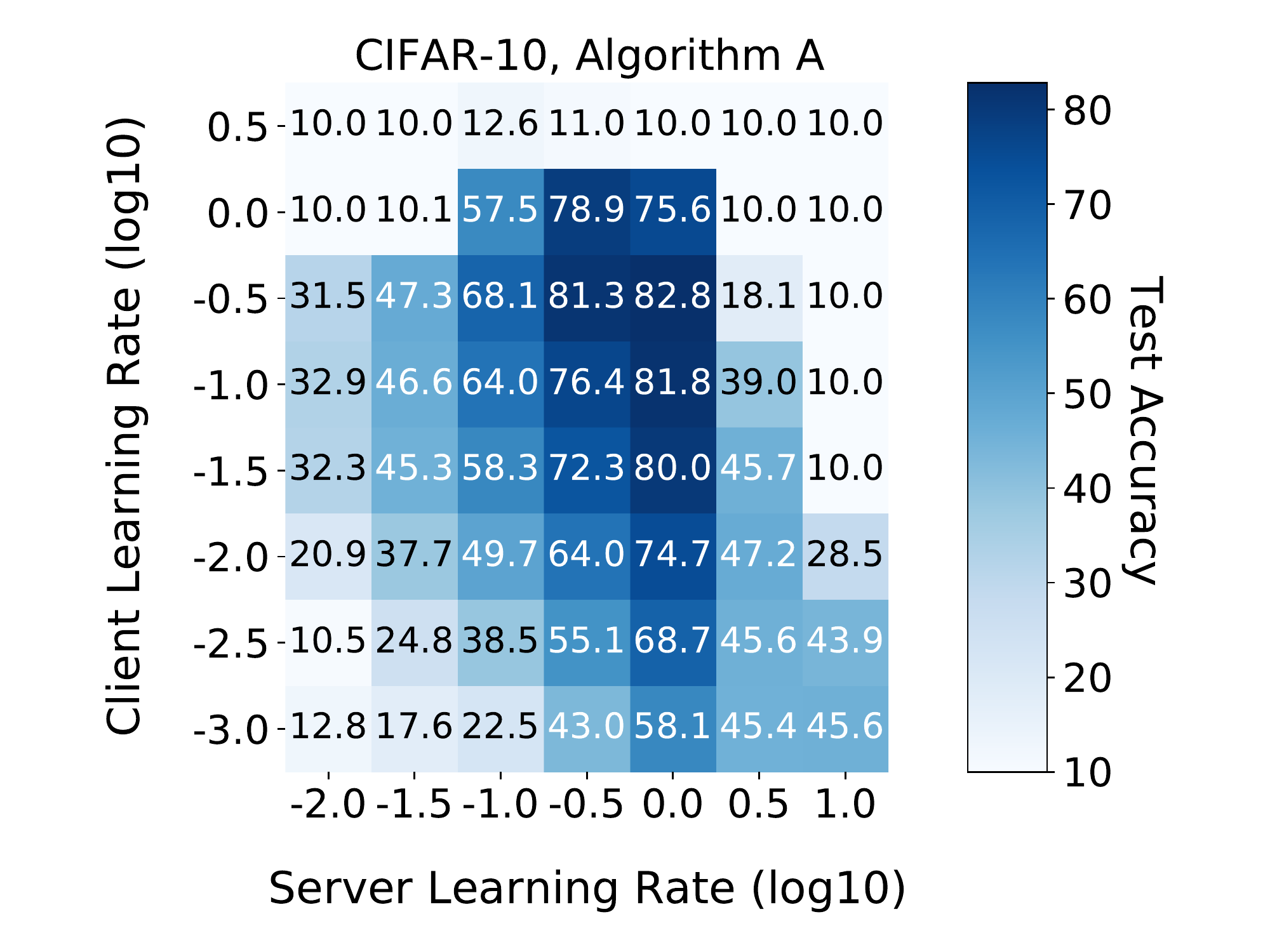}
\end{subfigure}
\begin{subfigure}
    \centering
    \includegraphics[width=0.3\linewidth]{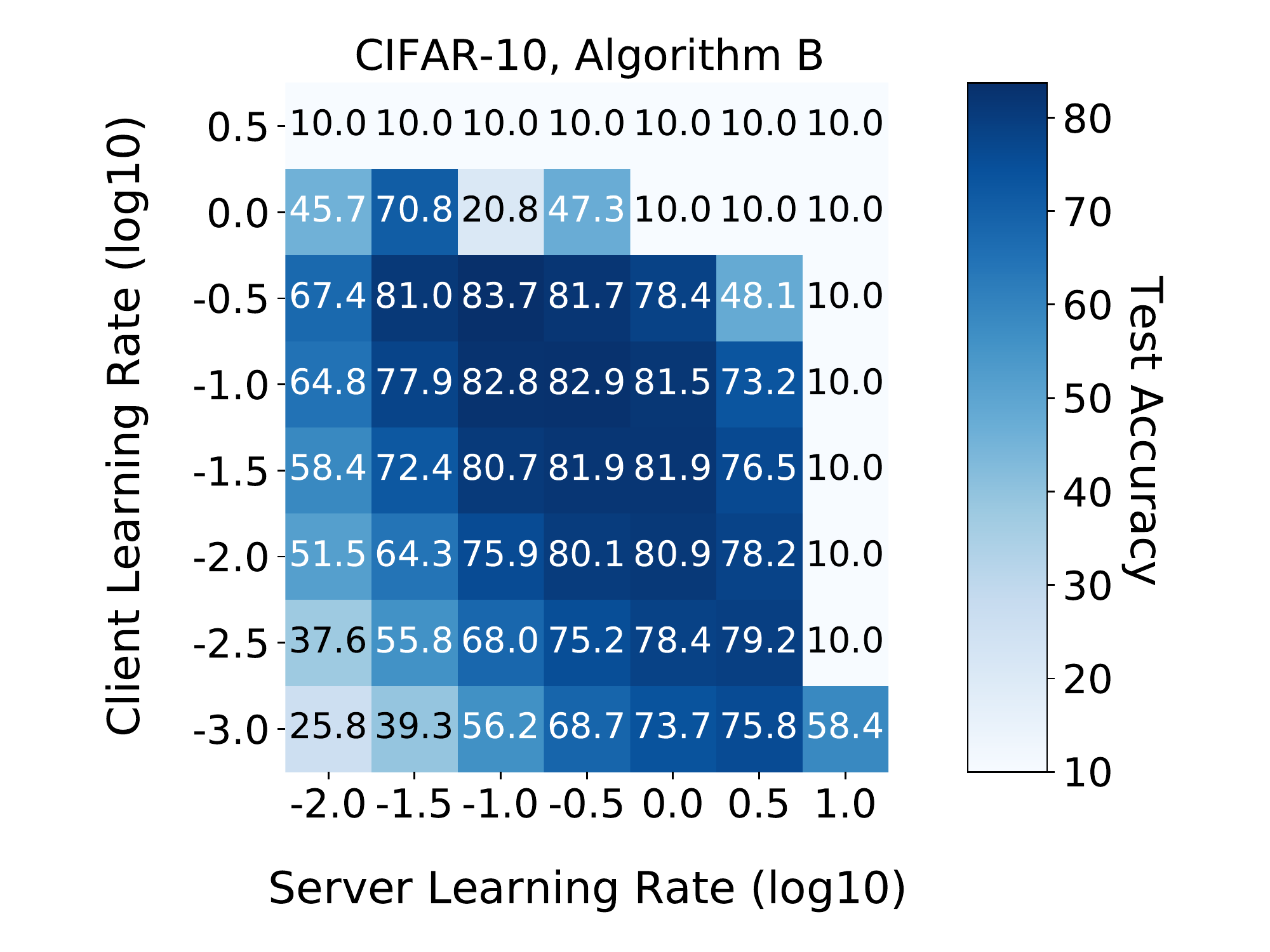}
\end{subfigure}
\begin{subfigure}
    \centering
    \includegraphics[width=0.3\linewidth]{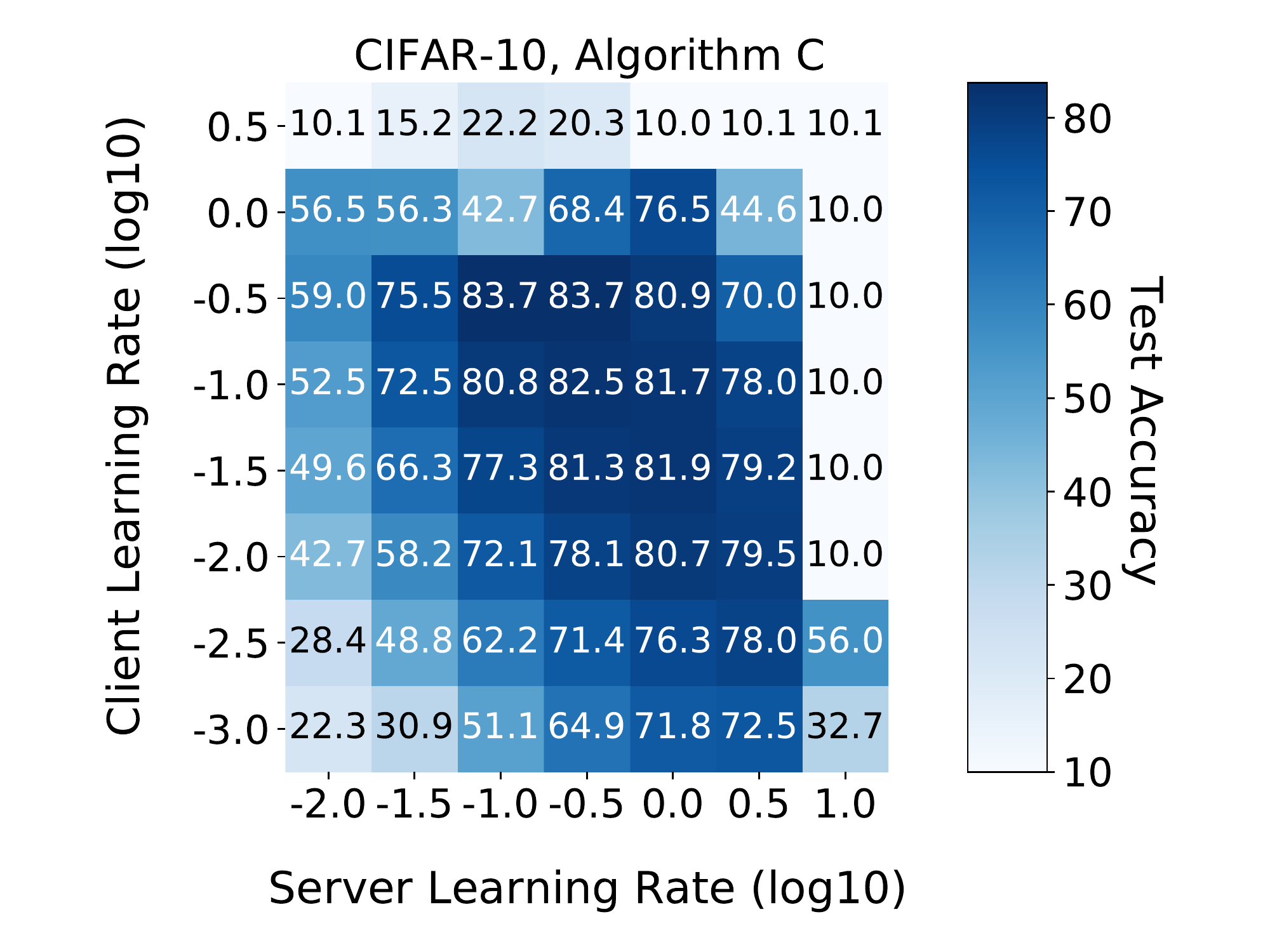}
\end{subfigure}
\caption{Test accuracy on CIFAR10 for various client and server learning rates. Results for Algorithms A, B, and C are given in the left, middle, and right plots, respectively. The test accuracy can show the hyperparameter sensitivity, while is unpractical for tuning.}
\end{figure}

\begin{figure}[ht]
\centering
\begin{subfigure}
    \centering
    \includegraphics[width=0.3\linewidth]{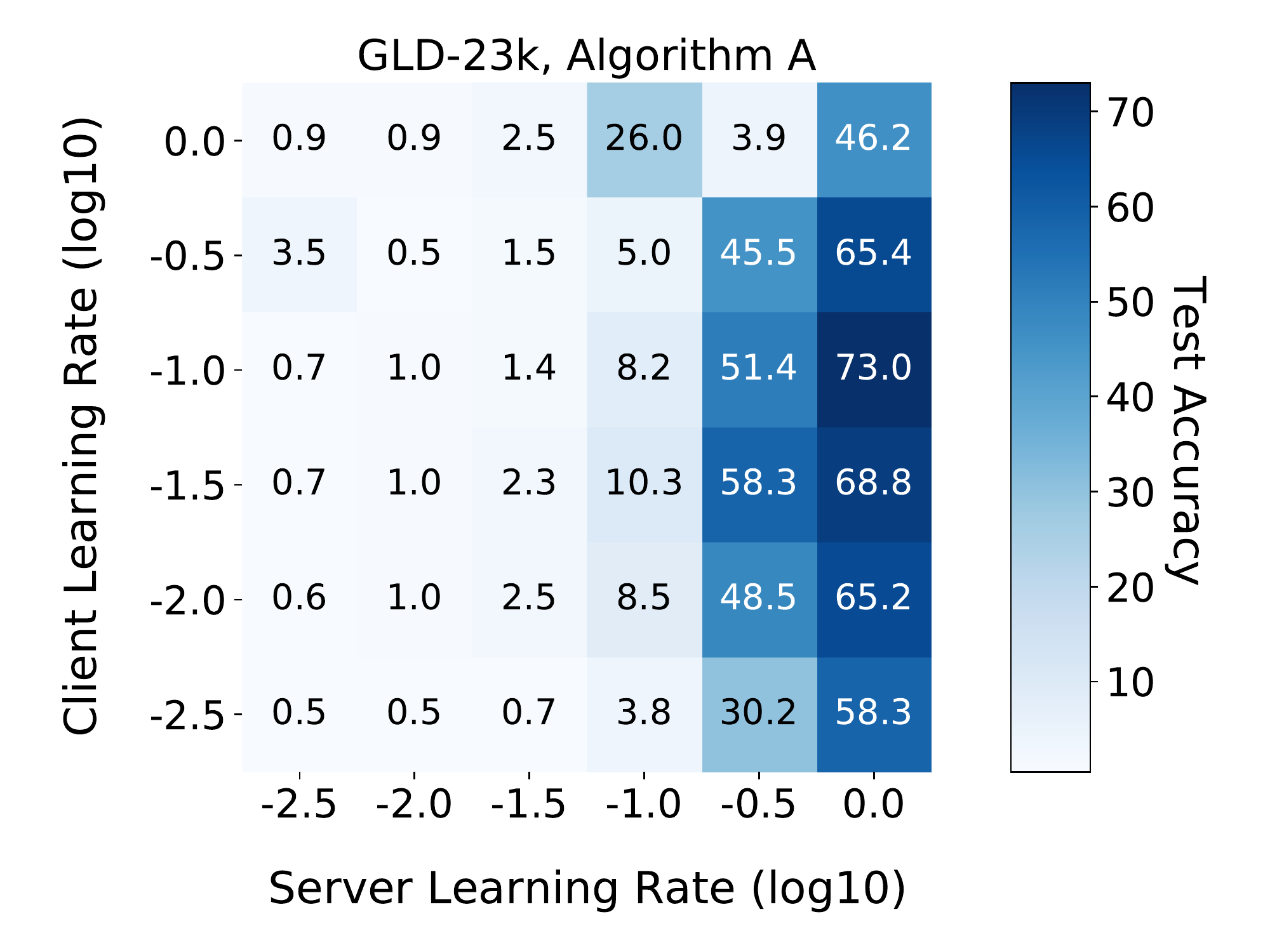}
\end{subfigure}
\begin{subfigure}
    \centering
    \includegraphics[width=0.3\linewidth]{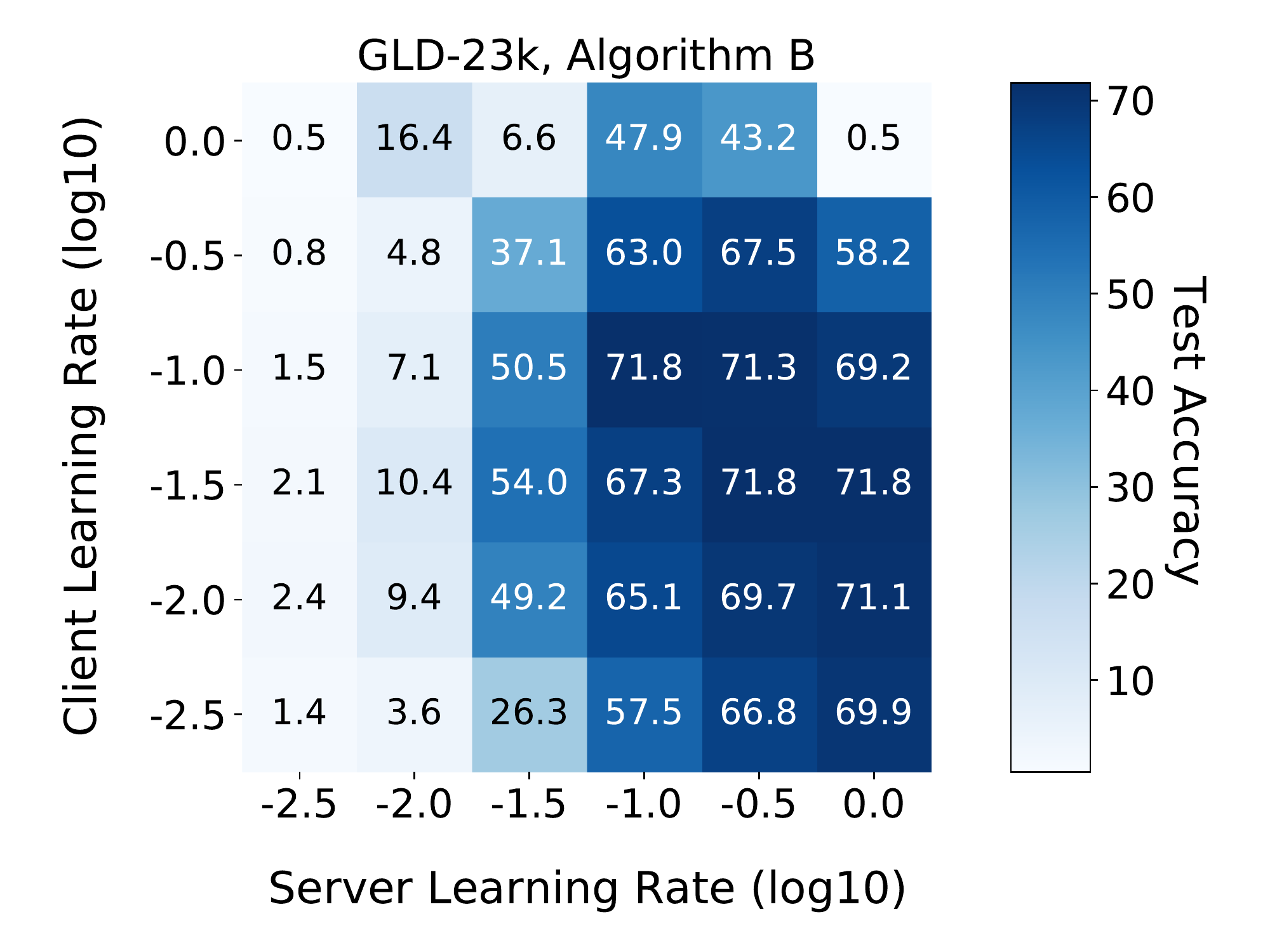}
\end{subfigure}
\begin{subfigure}
    \centering
    \includegraphics[width=0.3\linewidth]{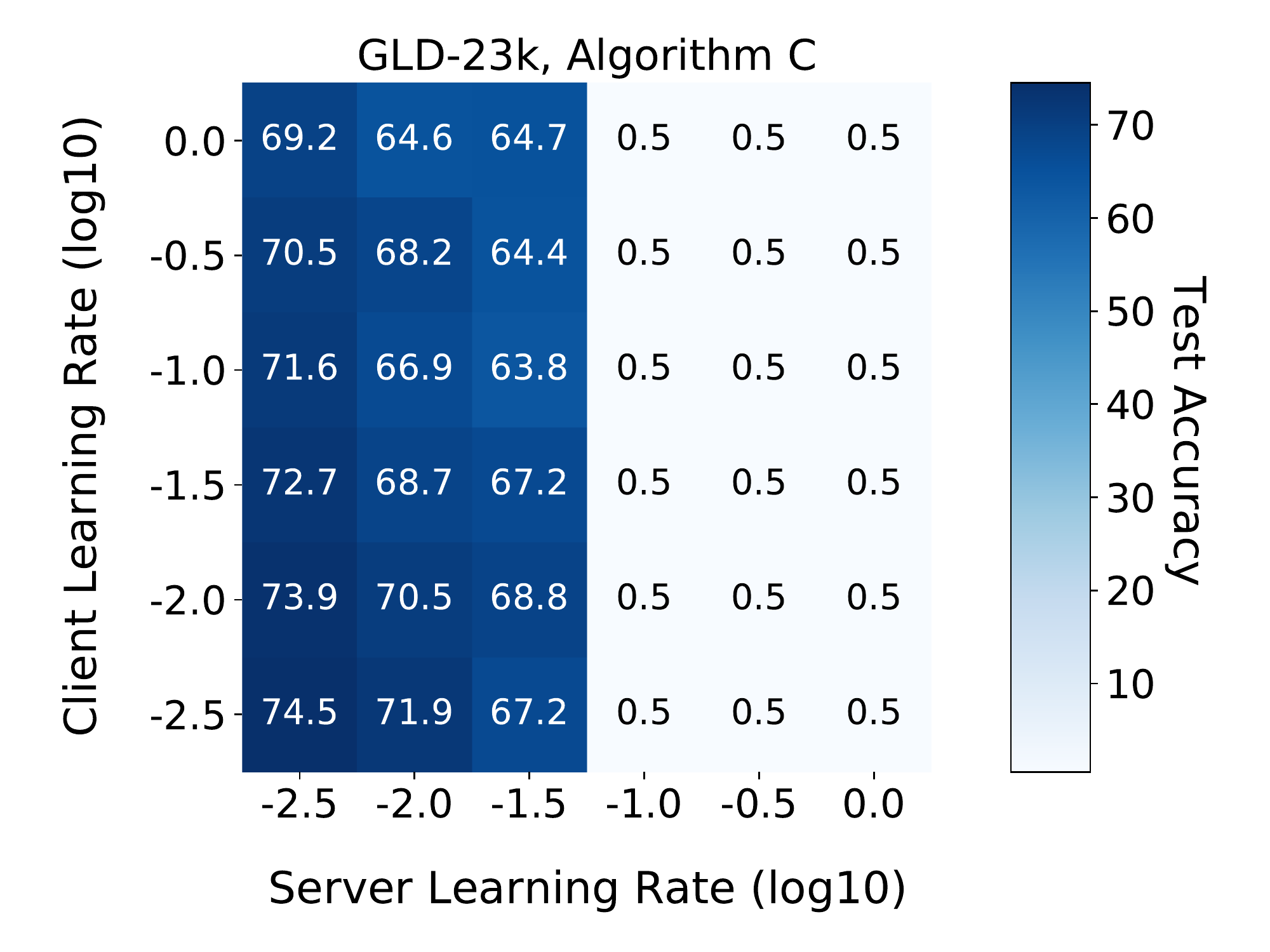}
\end{subfigure}
\caption{Test accuracy on GLD23k for various client and server learning rates. Results for Algorithms A, B, and C are given in the left, middle, and right plots, respectively.  The test accuracy can show the hyperparameter sensitivity, while is unpractical for tuning.}
\end{figure}

\begin{figure}[ht]
\centering
\begin{subfigure}
    \centering
    \includegraphics[width=0.3\linewidth]{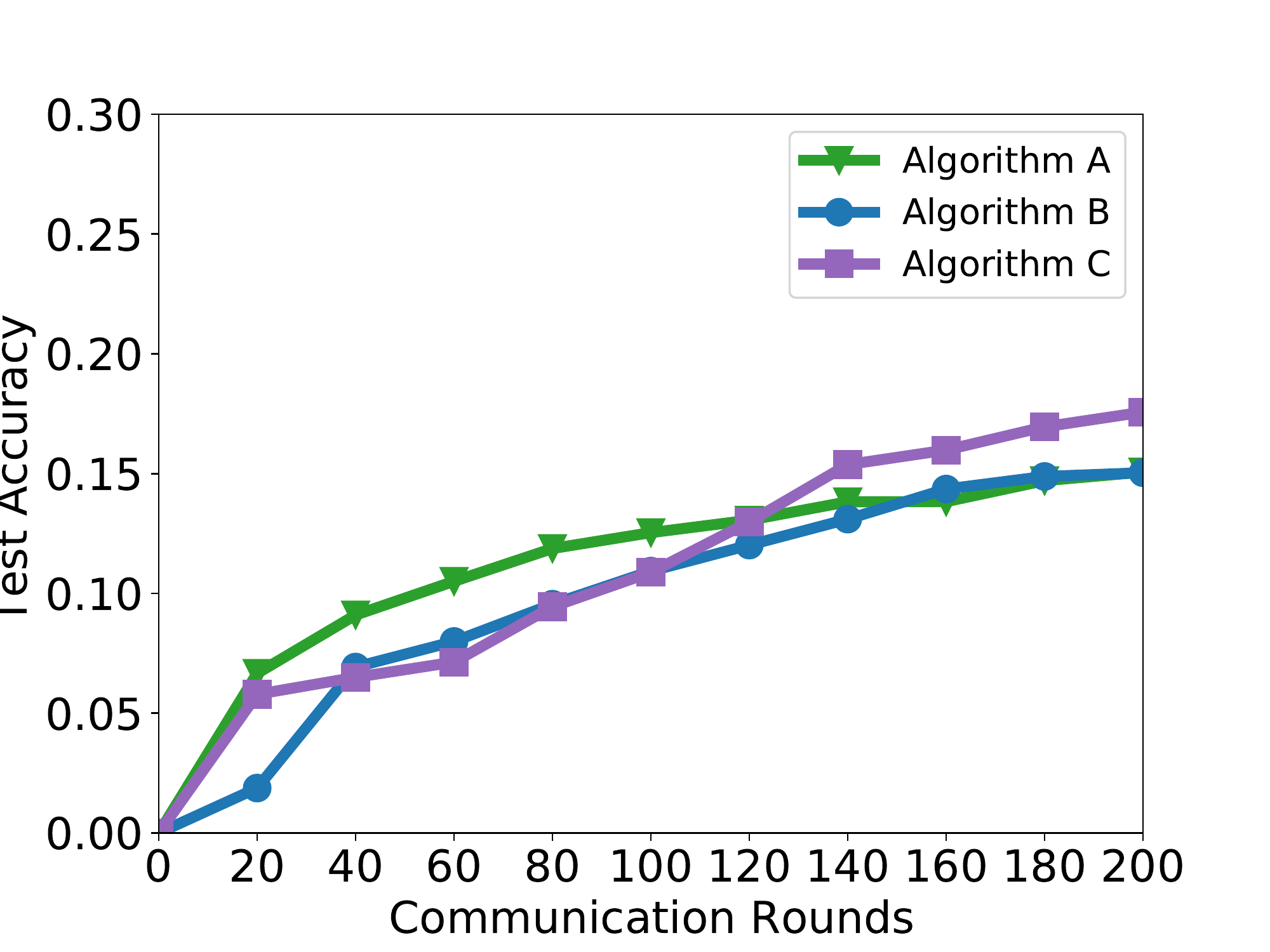}
\end{subfigure}
\begin{subfigure}
    \centering
    \includegraphics[width=0.3\linewidth]{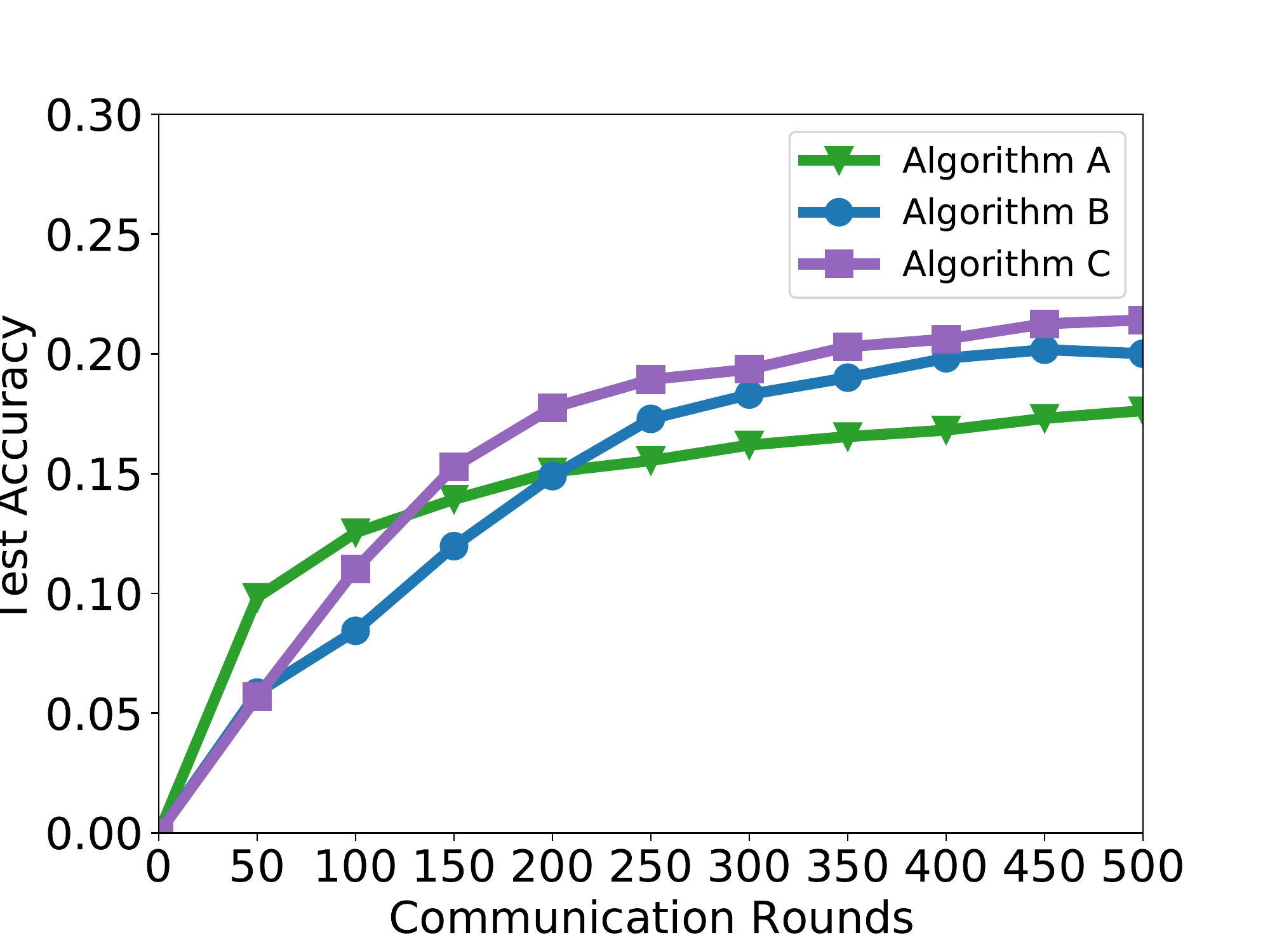}
\end{subfigure}
\begin{subfigure}
    \centering
    \includegraphics[width=0.3\linewidth]{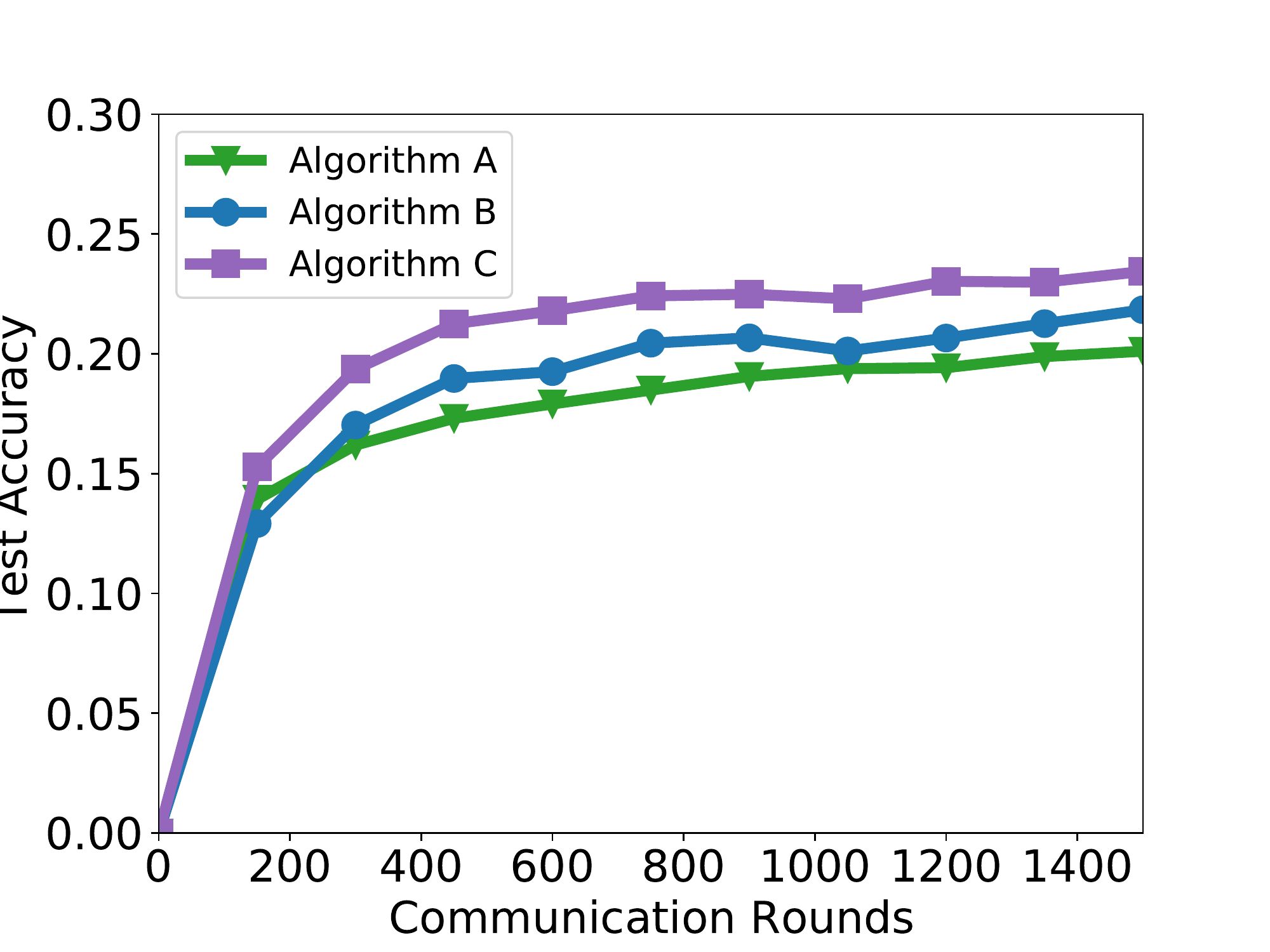}
\end{subfigure}
\caption{Validation accuracy on StackOverflow for a total of 200, 500, and 1000 communication rounds (left, middle, and right, respectively). }
\end{figure}

\begin{figure}[ht]
\centering
\begin{subfigure}
    \centering
    \includegraphics[width=0.45\linewidth]{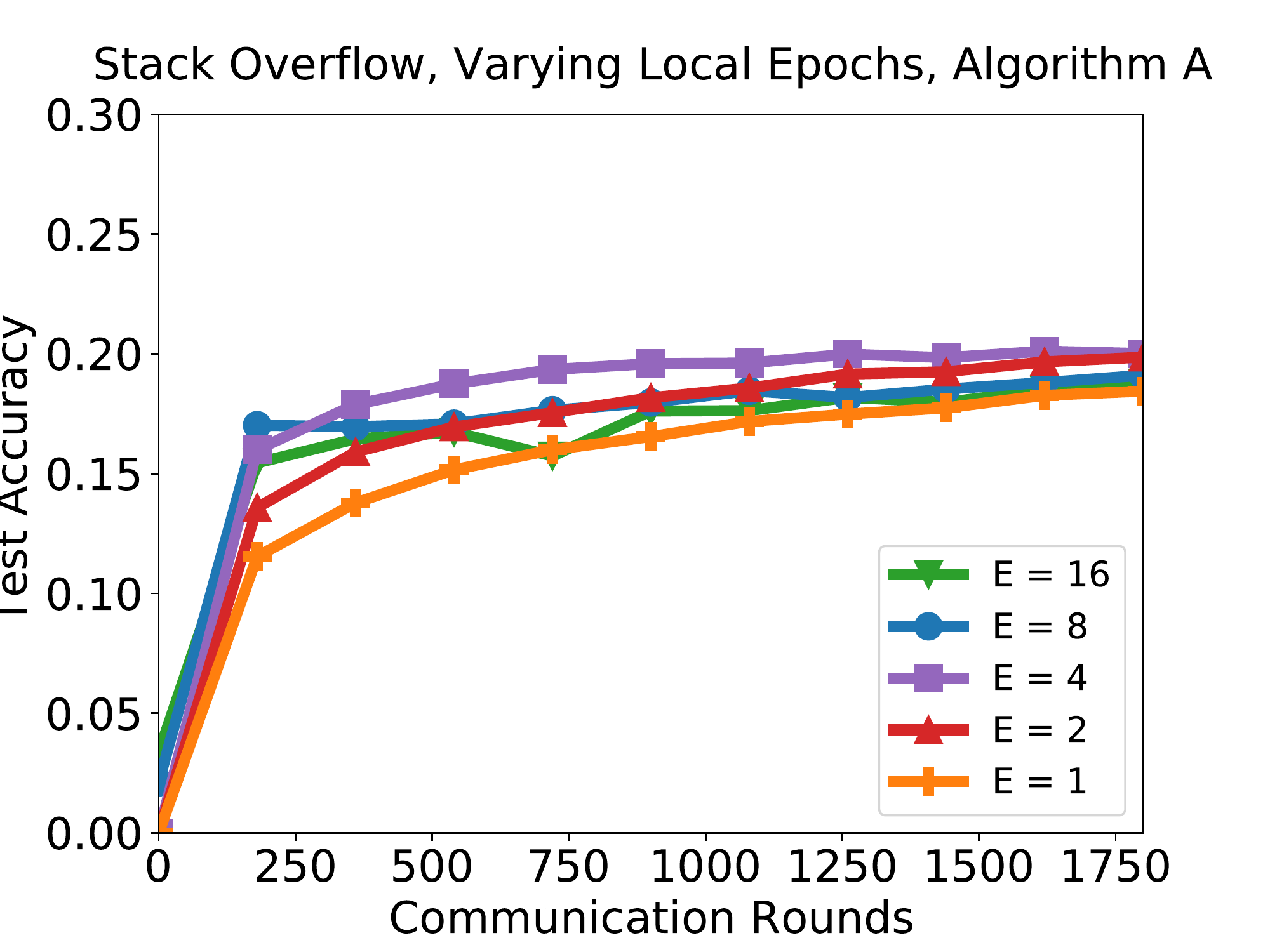}
\end{subfigure}
\begin{subfigure}
    \centering
    \includegraphics[width=0.45\linewidth]{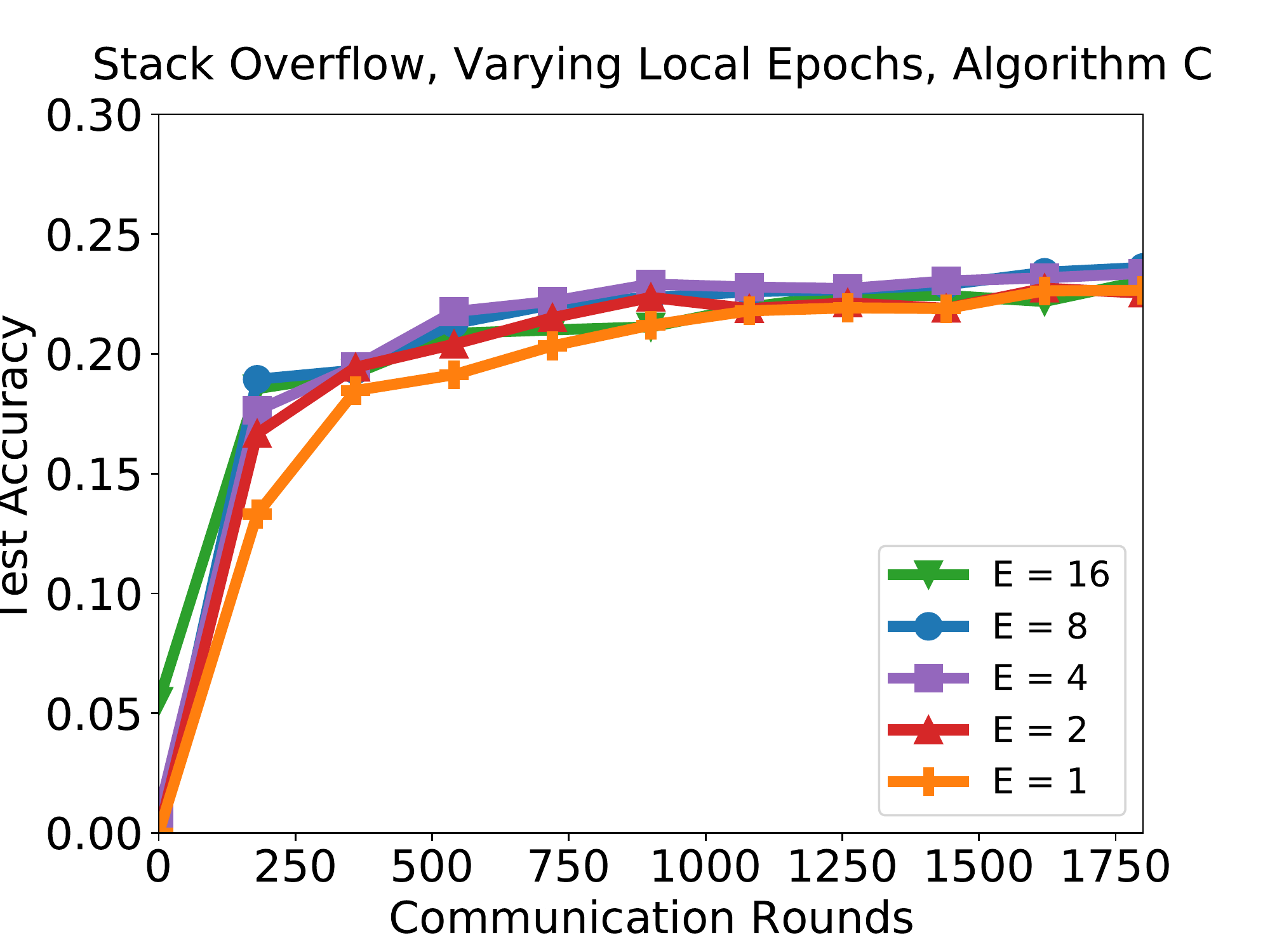}
\end{subfigure}
\caption{Validation accuracy on StackOverflow for Algorithms A (left) and C (right) for various numbers of local epochs per round $E$, versus the number of communication rounds.}
\end{figure}

\begin{figure}[ht]
\centering
\begin{subfigure}
    \centering
    \includegraphics[width=0.45\linewidth]{images/stackoverflow/stackoverflow_num_epochs_versus_num_rounds_algorithm_c.pdf}
\end{subfigure}
\begin{subfigure}
    \centering
    \includegraphics[width=0.45\linewidth]{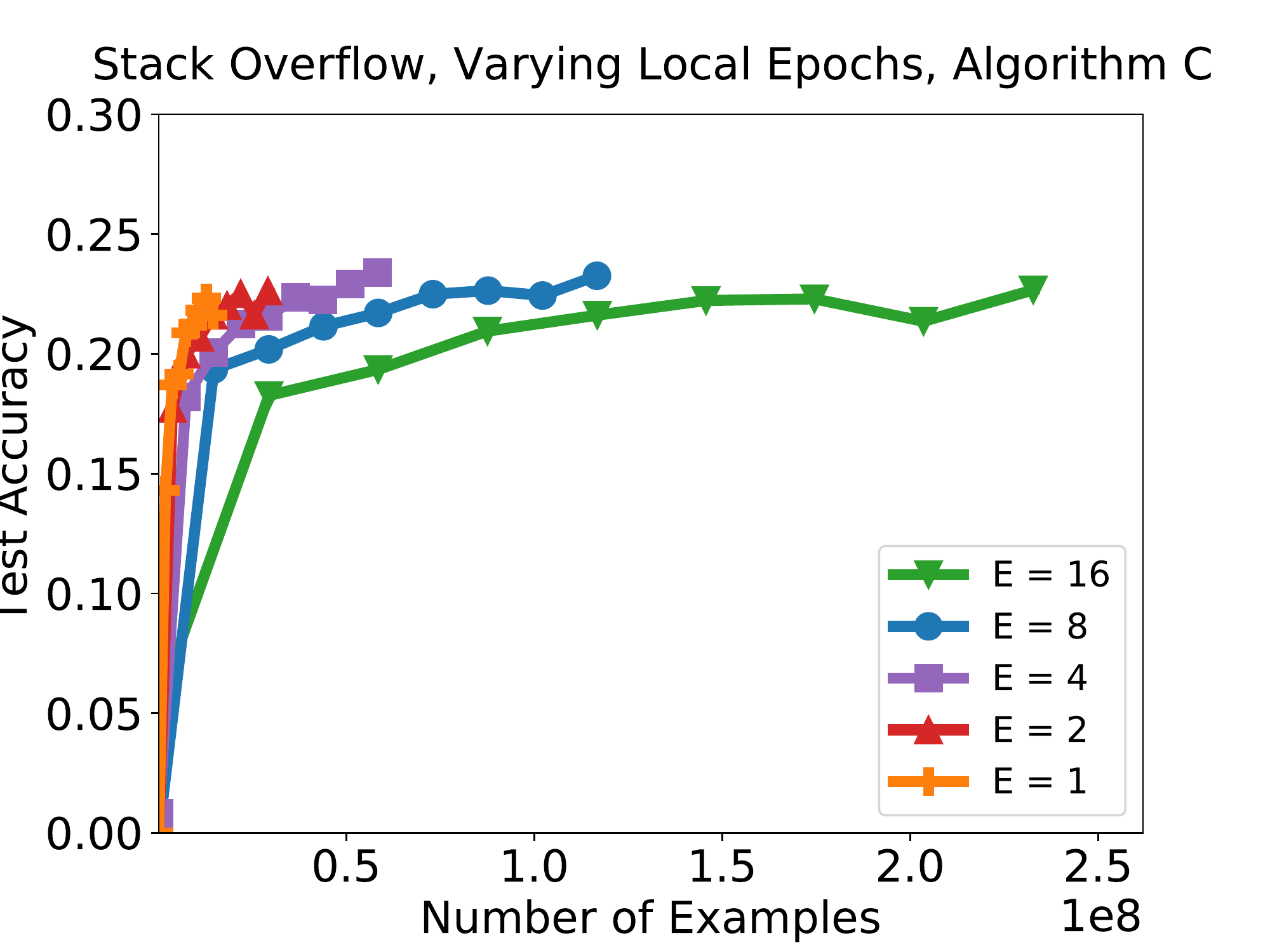}
\end{subfigure}
\caption{Validation accuracy on StackOverflow for Algorithms C. We plot validation accuracy versus the number of communication rounds (left) and versus the total number of examples processed by clients (right) for various numbers of local epochs per round $E$. We use learning rates $\eta = 0.01, \eta_s = 1.0$.}
\end{figure}

\begin{figure}[ht]
\centering
\begin{subfigure}
    \centering
    \includegraphics[width=0.45\linewidth]{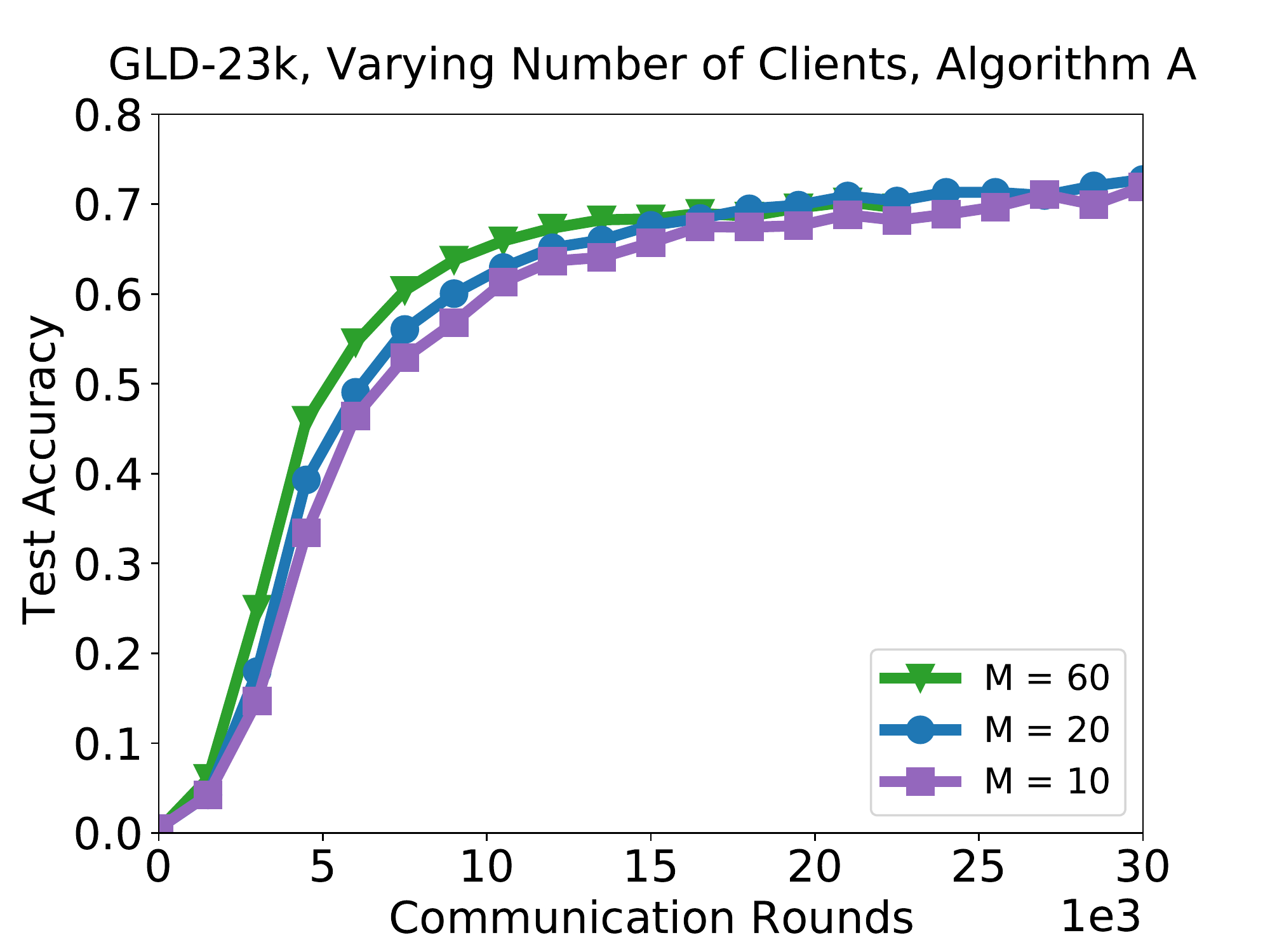}
\end{subfigure}
\begin{subfigure}
    \centering
    \includegraphics[width=0.45\linewidth]{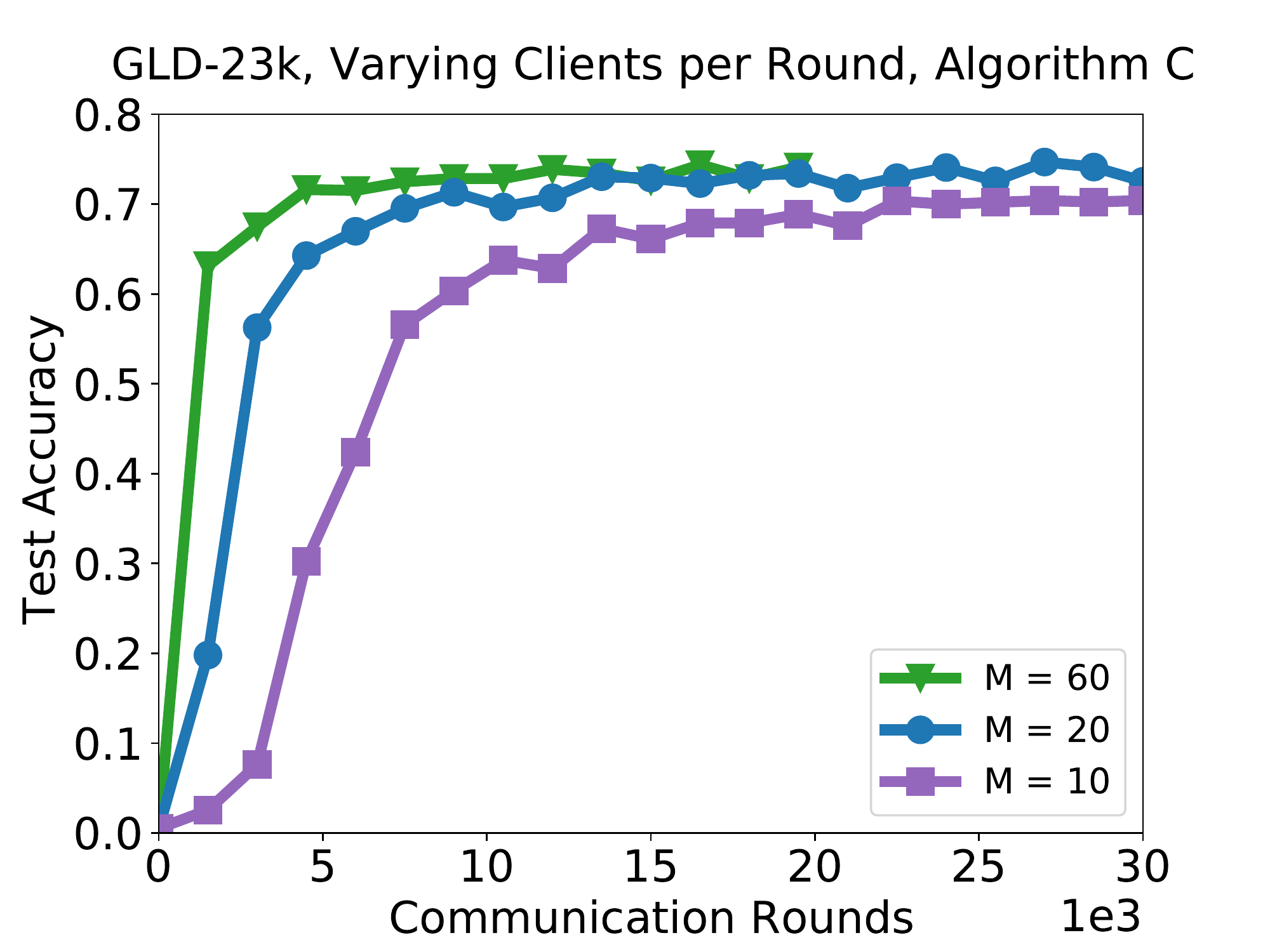}
\end{subfigure}
\caption{Test accuracy on GLD-23k for Algorithms A (left) and C (right) for various cohort sizes $M$, versus the number of communication rounds.}
\end{figure}

\begin{figure}[ht]
\centering
\begin{subfigure}
    \centering
    \includegraphics[width=0.45\linewidth]{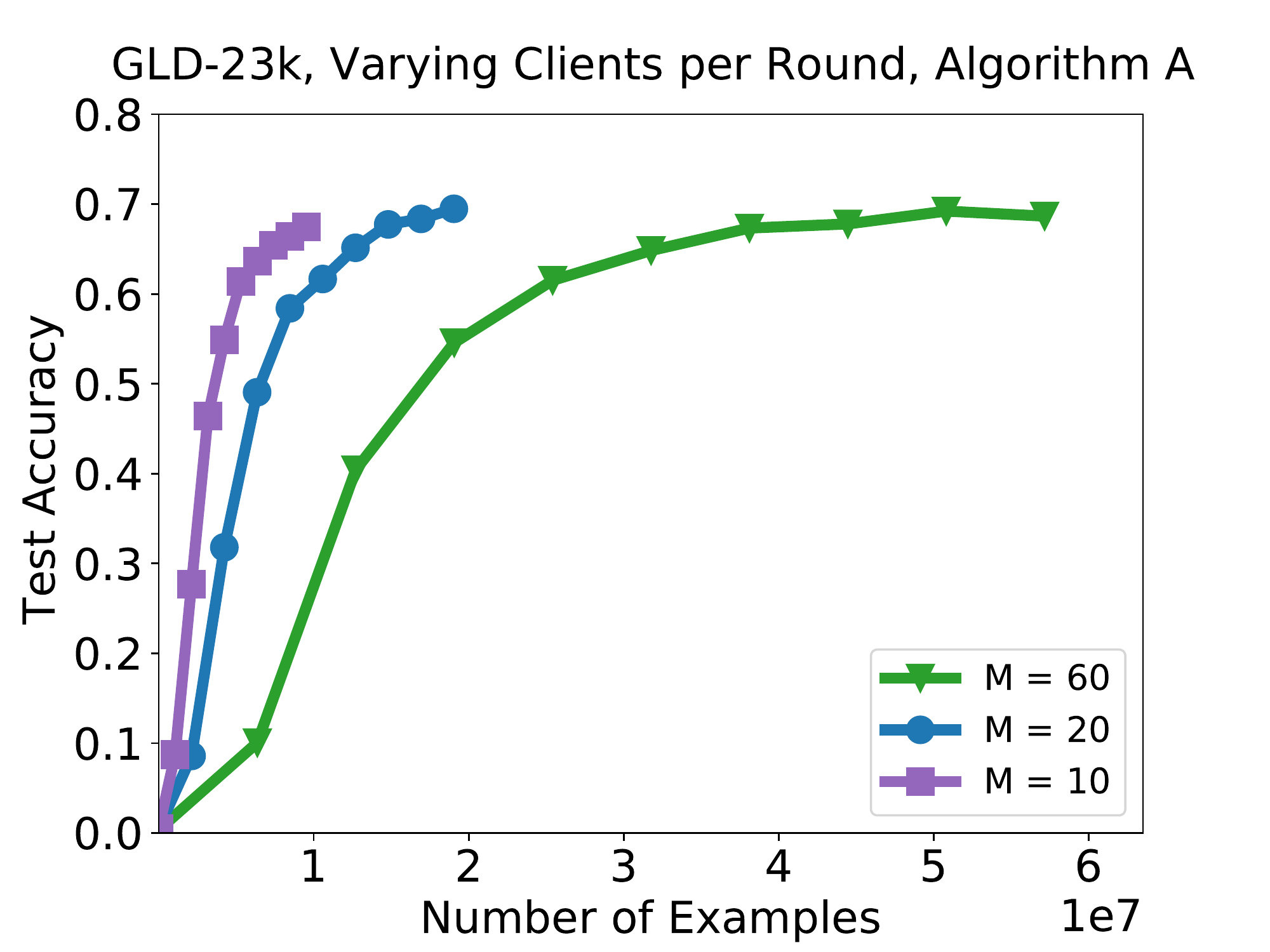}
\end{subfigure}
\begin{subfigure}
    \centering
    \includegraphics[width=0.45\linewidth]{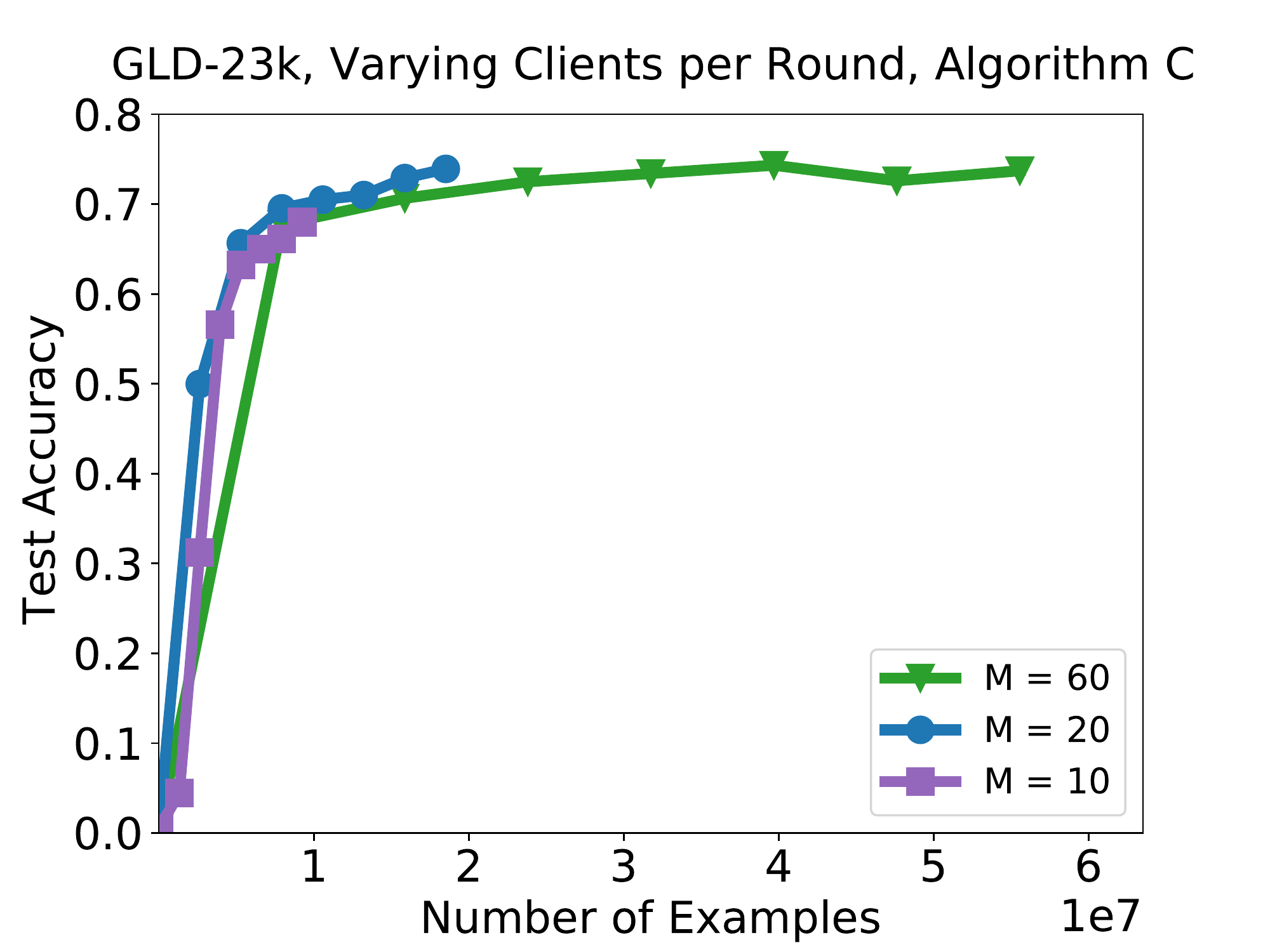}
\end{subfigure}
\caption{Test accuracy on GLD-23k for Algorithms A (left) and C (right) for various cohort size $M$, versus the number of examples processed by the clients.}
\end{figure}

\begin{figure}[ht]
\centering
\begin{subfigure}
    \centering
    \includegraphics[width=0.45\linewidth]{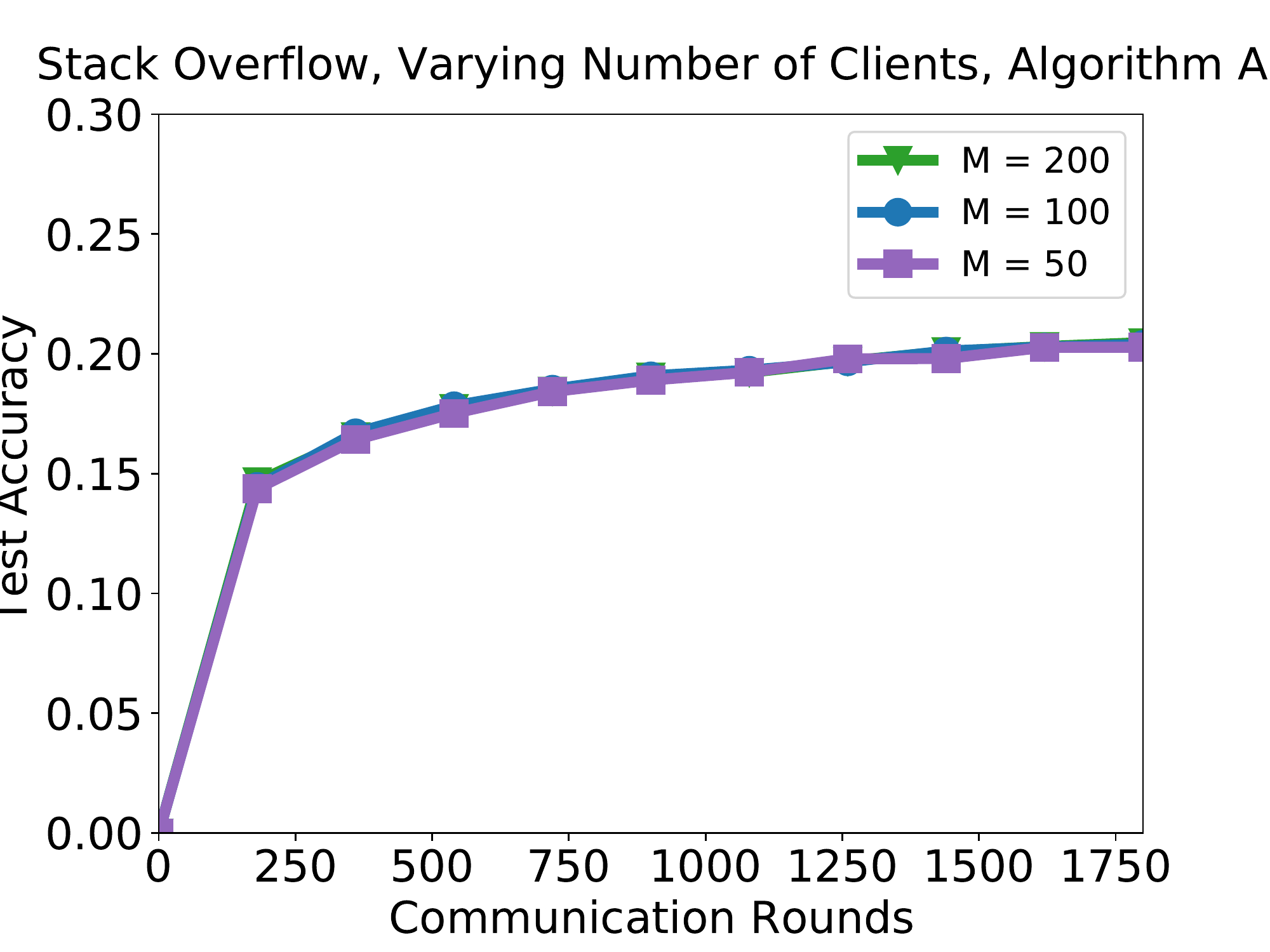}
\end{subfigure}
\begin{subfigure}
    \centering
    \includegraphics[width=0.45\linewidth]{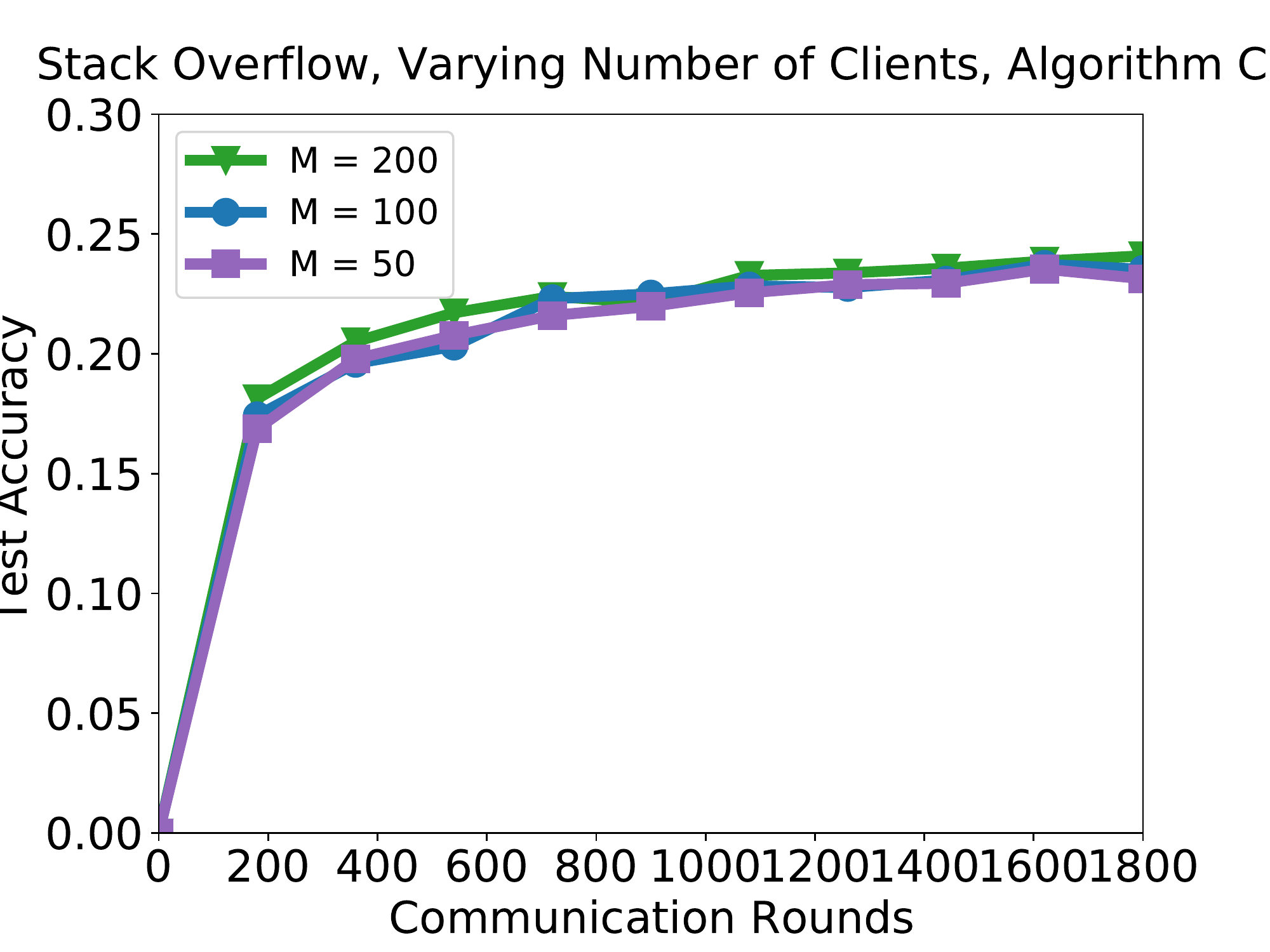}
\end{subfigure}
\caption{Validation accuracy on StackOverflow for Algorithms A (left) and C (right) for various cohort size $M$, versus the number of communication rounds.}
\end{figure}

\clearpage

\subsection{Basic Model to Estimate On-Device Training Times in \Cref{sec:basicmodel} }

\begin{figure}[ht]
\centering
\begin{subfigure}
    \centering
    \includegraphics[width=0.45\linewidth]{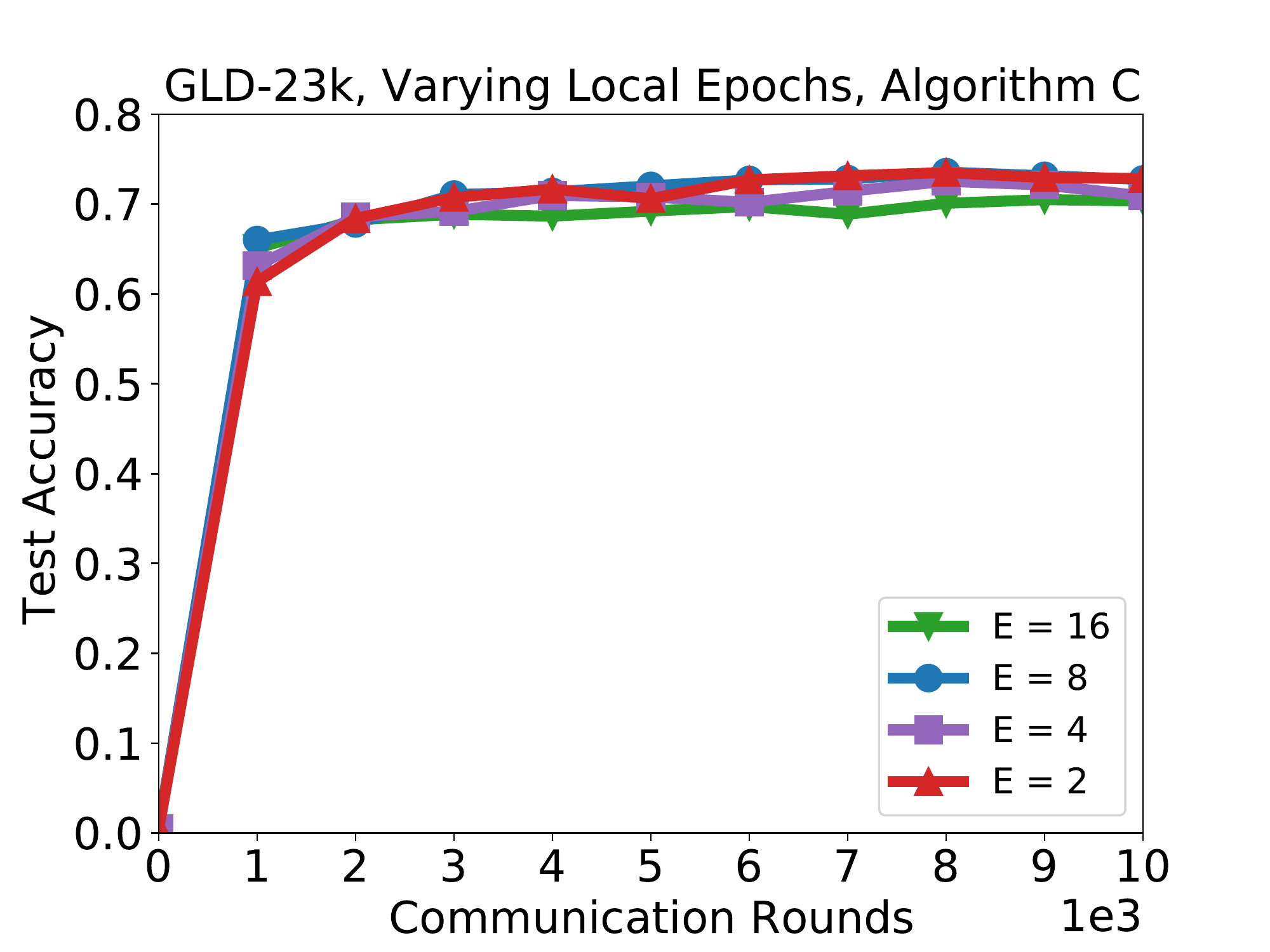}
\end{subfigure}
\begin{subfigure}
    \centering
    \includegraphics[width=0.45\linewidth]{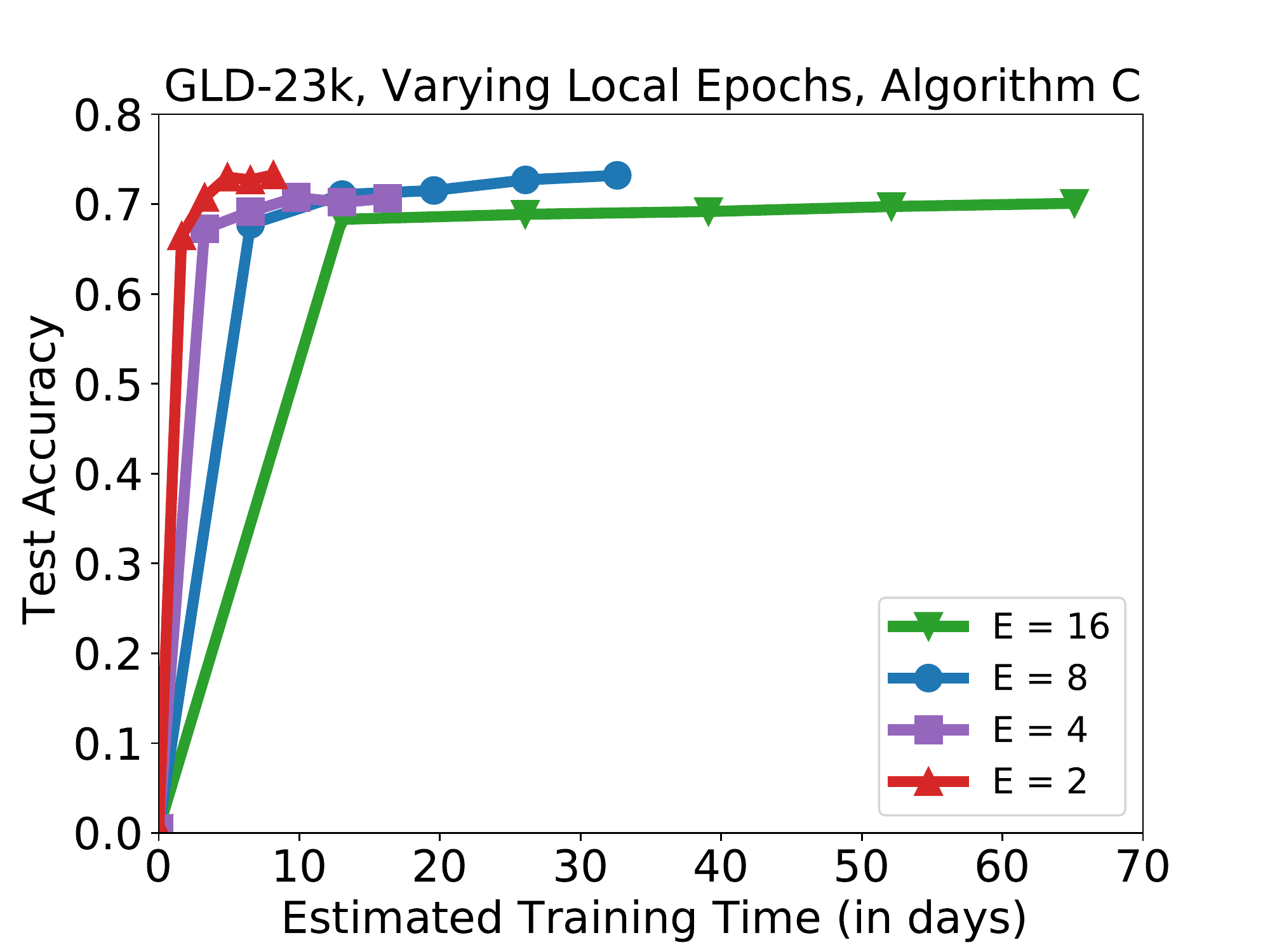}
\end{subfigure}
\caption{Test accuracy on GLD-23k for Algorithm C, for varying numbers of local epochs per round. We plot the test accuracy versus the number of communication rounds (left) and the estimate completion time using the model above (right).}
\label{fig:gld23k_system_model_alg_c}
\end{figure}

\begin{figure}[ht]
\centering
\begin{subfigure}
    \centering
    \includegraphics[width=0.45\linewidth]{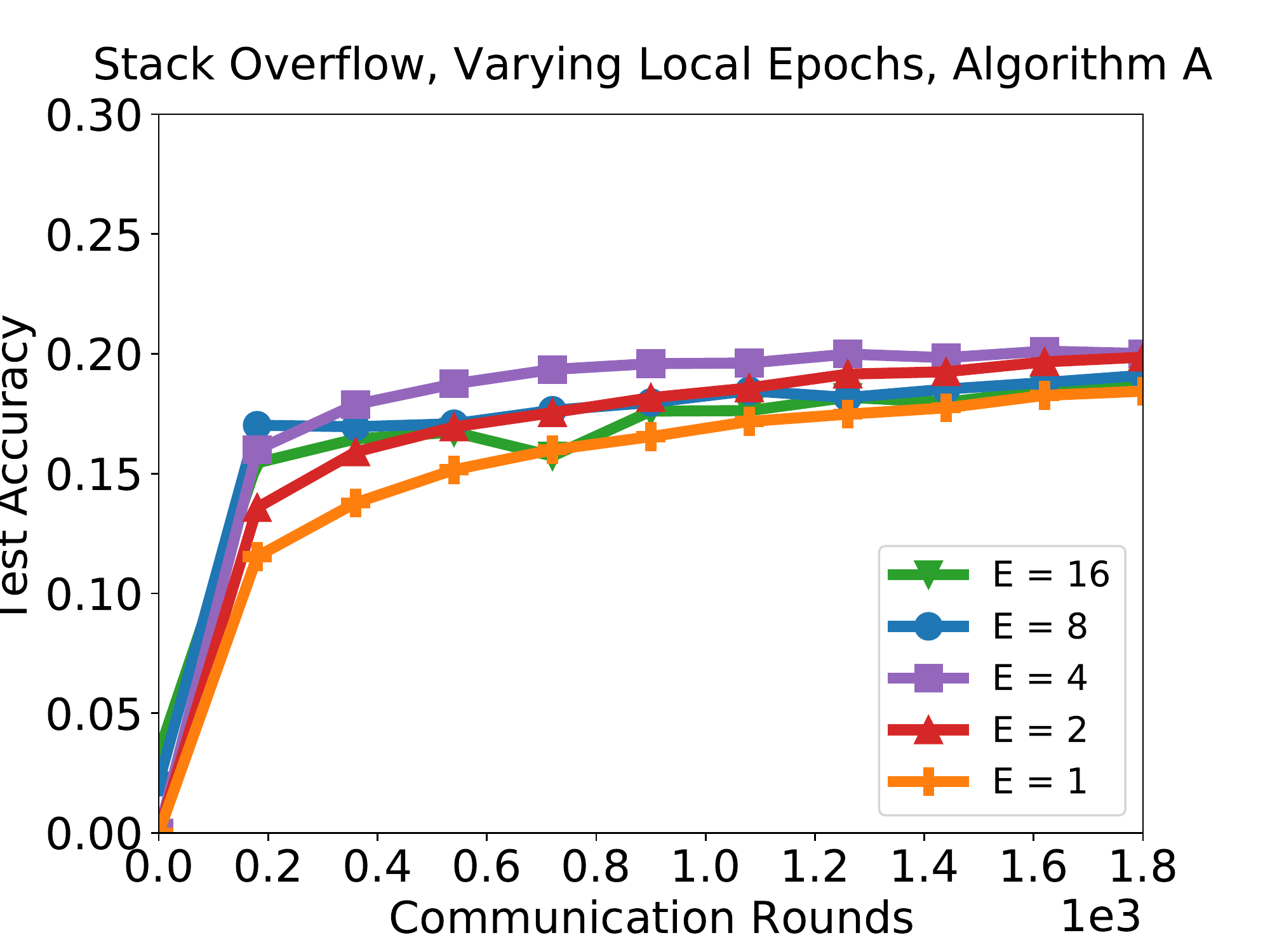}
\end{subfigure}
\begin{subfigure}
    \centering
    \includegraphics[width=0.45\linewidth]{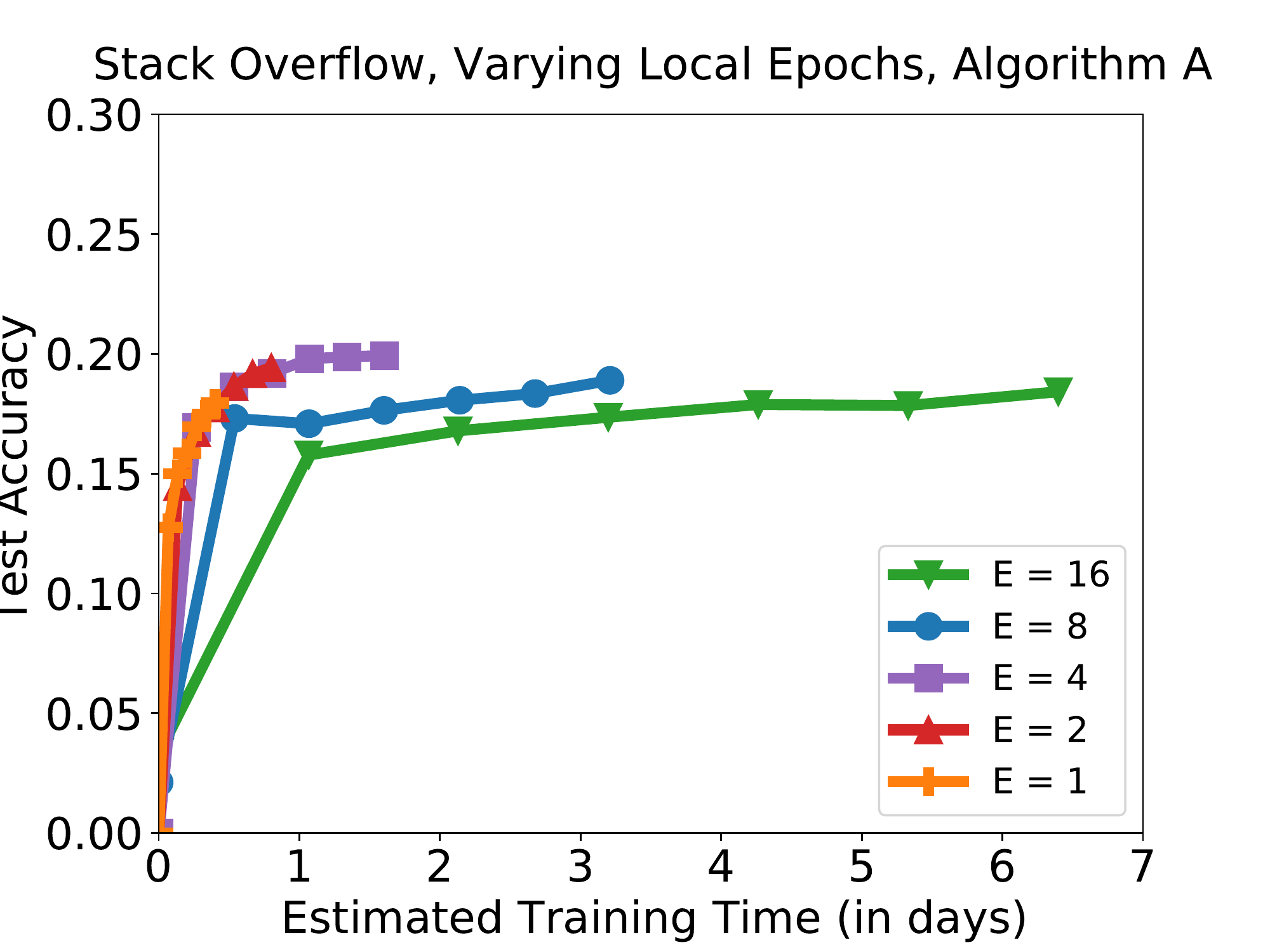}
\end{subfigure}
\caption{Accuracy of Algorithm A on a fixed set of 10,000 randomly selected test examples, for varying numbers of local epochs per round $E$. We plot the test accuracy versus the number of communication rounds (left) and the estimate completion time using the model above (right).}
\label{fig:stackoverflow_system_model_alg_a}
\end{figure}

\begin{figure}[ht]
\centering
\begin{subfigure}
    \centering
    \includegraphics[width=0.45\linewidth]{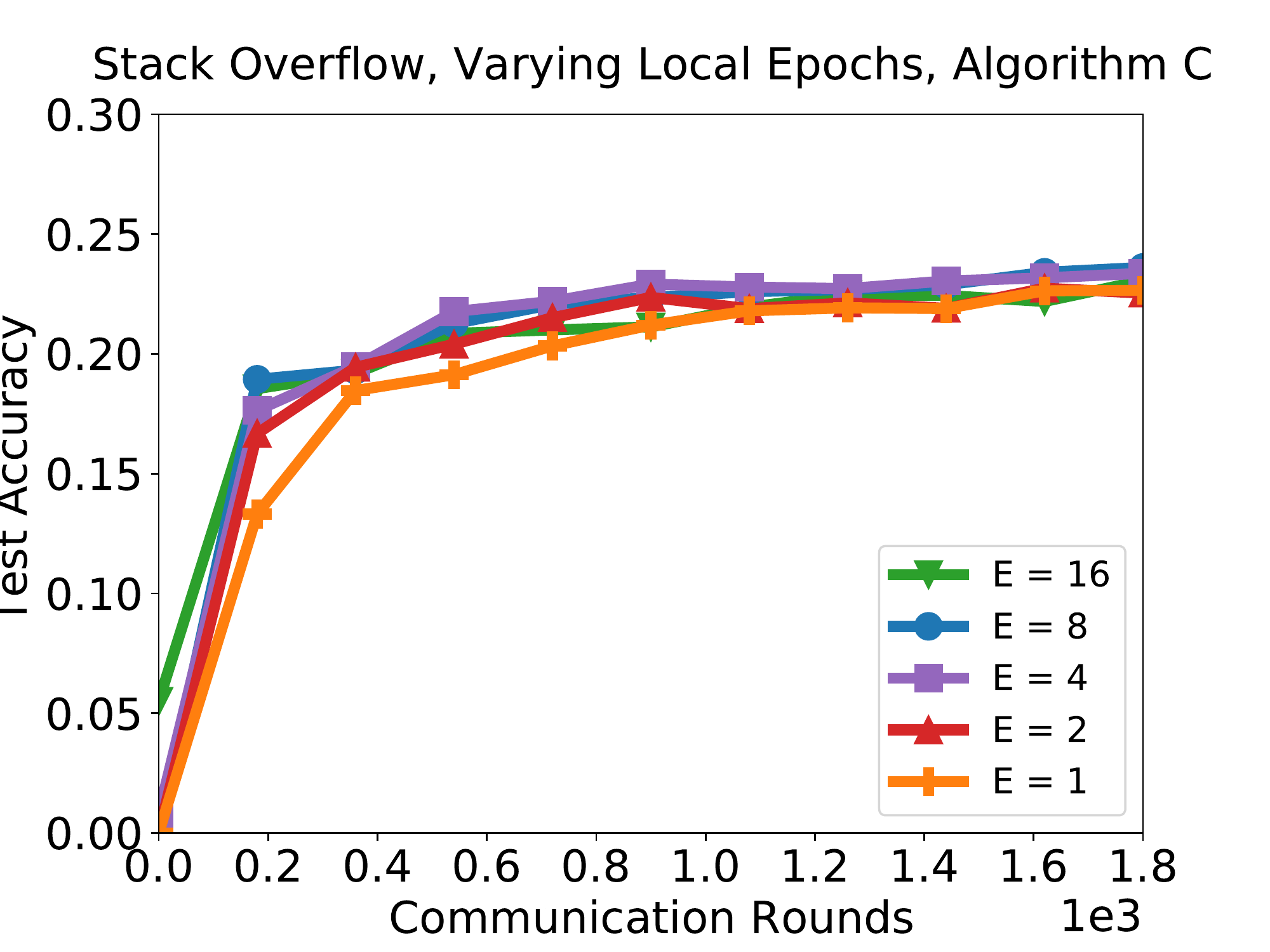}
\end{subfigure}
\begin{subfigure}
    \centering
    \includegraphics[width=0.45\linewidth]{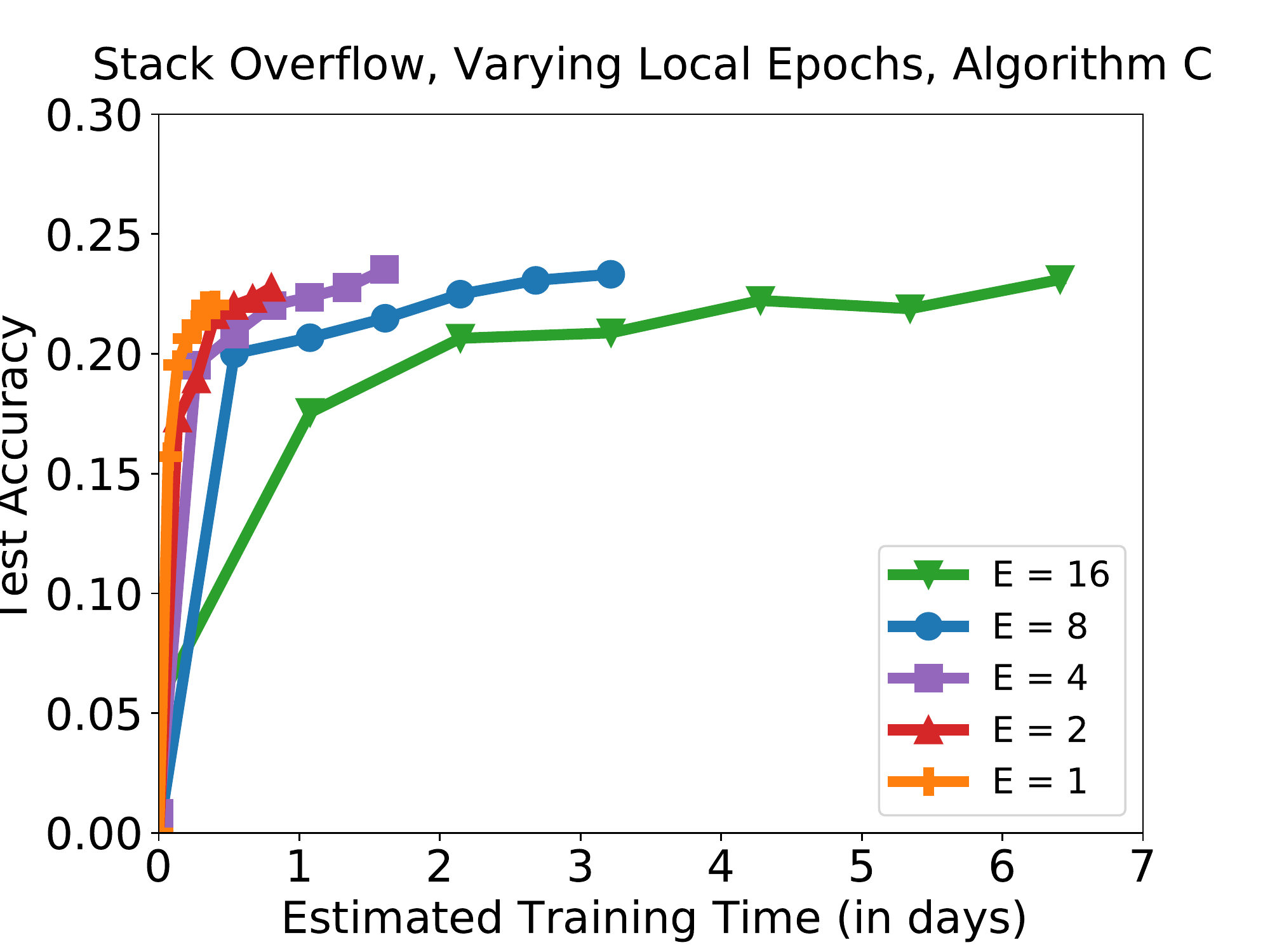}
\end{subfigure}
\caption{Accuracy of Algorithm C on a fixed set of 10,000 randomly selected test examples, for varying numbers of local epochs per round $E$. We plot the test accuracy versus the number of communication rounds (left) and the estimate completion time using the model above (right).}
\label{fig:stackoverflow_system_model_alg_c}
\end{figure}

\clearpage

\section{Proofs} \label{appendix:proof}
\subsection{Deferred Proof of \cref{lem:1}}
    \begin{proof}[Proof of \cref{lem:1}]
    Since $\overline{\vx}^{(t,k+1)} = \overline{\vx}^{(t,k)} - \eta \frac{1}{M} \sum_{i=1}^M g_i(\vx_i^{(t,k)})$, by parallelogram law
    \begin{equation}
        \frac{1}{M} \sum_{i=1}^M \left\langle g_i (\vx_i^{(t,k)}), \overline{\vx}^{(t,k+1)} - \vx^{\star}  \right\rangle 
        =
        \frac{1}{2\eta} \left( \left\|\overline{\vx}^{(t,k)} - \vx^{\star} \right\|^2 
        - \left\| \overline{\vx}^{(t,k+1)} -  \overline{\vx}^{(t,k)}  \right\|^2 
        - \left\|\overline{\vx}^{(t,k+1)} - \vx^{\star} \right\|^2  \right).
        \label{eq:lem:1:1}
    \end{equation}
    By convexity and $L$-smoothness of $F_i$, one has
    \begin{align}
        & F_i ( \overline{\vx}^{(t,k+1)} ) 
        \leq
        F_i ( \vx_i^{(t,k)})  + \left\langle \nabla F_i ( \vx_i^{(t,k)}), \overline{\vx}^{(t,k+1)} - \vx_i^{(t,k)} \right\rangle + \frac{L}{2} \left\|  \overline{\vx}^{(t,k+1)} -  \vx_i^{(t,k)}  \right\|^2
        \tag{$L$-smoothness}
        \\
        \leq &  F_i ( \vx^{\star})  + \left\langle \nabla F_i ( \vx_i^{(t,k)}), \overline{\vx}^{(t,k+1)} - \vx^{\star} \right\rangle
        + \frac{L}{2} \left\|  \overline{\vx}^{(t,k+1)} -  \vx_i^{(t,k)}  \right\|^2
        \tag{convexity}
        \\
        \leq &  F_i ( \vx^{\star})  + \left\langle \nabla F_i ( \vx_i^{(t,k)}), \overline{\vx}^{(t,k+1)} - \vx^{\star} \right\rangle
        + L \left\|  \overline{\vx}^{(t,k+1)} -  \overline{\vx}^{(t,k)} \right\|^2 
        + L \left\| \vx_i^{(t,k)} -  \overline{\vx}^{(t,k)} \right\|^2 
        \label{eq:lem:1:2}
    \end{align}
    Combining \cref{eq:lem:1:1,eq:lem:1:2} yields
    \begin{align}
        & F ( \overline{\vx}^{(t,k+1)} ) - F(\vx^{\star}) = \frac{1}{M} \sum_{i=1}^M \left(  F_i ( \overline{\vx}^{(t,k+1)} ) - F(\vx^{\star})  \right)
        \nonumber
        \\
    \leq  &
    \frac{1}{M} \sum_{i=1}^M \left\langle  \nabla F_i ( \vx_i^{(t,k)}) - g_i(\vx_i^{(t,k)}), \overline{\vx}^{(t,k+1)} - \vx^{\star} \right\rangle 
    + {L} \left\|  \overline{\vx}^{(t,k+1)} -  \overline{\vx}^{(t,k)} \right\|^2 
    + \frac{L}{M} \sum_{i=1}^M \left\| \vx_i^{(t,k)} -  \overline{\vx}^{(t,k)} \right\|^2 
    \nonumber
    \\
        & +   \frac{1}{2\eta} \left( \left\|\overline{\vx}^{(t,k)} - \vx^{\star} \right\|^2 
        - \left\|\overline{\vx}^{(t,k+1)} - \vx^{\star} \right\|^2 
        - \left\| \overline{\vx}^{(t,k+1)} -  \overline{\vx}^{(t,k)}  \right\|^2 
        \right).
    \label{eq:lem:1:3}
    \end{align}
    Since $\Exs \left[ \nabla F_i ( \vx_i^{(t,k)}) - g_i ( \vx_i^{(t,k)}) \middle| \mathcal{F}^{(t,k)} \right] = 0$ we have
    \begin{align}
        & \Exs \left[ \frac{1}{M} \sum_{i=1}^M \left\langle  \nabla F_i ( \vx_i^{(t,k)}) - g_i ( \vx_i^{(t,k)}), \overline{\vx}^{(t,k+1)} - \vx^{\star} \right\rangle  \middle| \mathcal{F}^{(t,k)} \right]
        \nonumber
        \\
    =   & \Exs \left[ \frac{1}{M} \sum_{i=1}^M \left\langle  \nabla F_i ( \vx_i^{(t,k)}) - g_i ( \vx_i^{(t,k)}), \overline{\vx}^{(t,k+1)} - \overline{\vx}^{(t,k)}\right\rangle  \middle| \mathcal{F}^{(t,k)} \right]
    \nonumber
        \\
    \leq & \eta \cdot \Exs \left[ \left\| \frac{1}{M} \sum_{i=1}^M (\nabla F_i ( \vx_i^{(t,k)}) - g_i ( \vx_i^{(t,k)})) \right\|^2 \middle| \mathcal{F}^{(t,k)} \right] 
    +  \frac{1}{4\eta} \cdot \Exs \left[ \left\| \overline{\vx}^{(t,k+1)} - \overline{\vx}^{(t,k)} \right\|^2 \middle| \mathcal{F}^{(t,k)} \right]
    \tag{Young's inequality}
        \\
    \leq & \frac{\eta \sigma^2 }{M} 
    +
    \frac{1}{4\eta} \cdot \Exs \left[ \left\| \overline{\vx}^{(t,k+1)} - \overline{\vx}^{(t,k)} \right\|^2 \middle| \mathcal{F}^{(t,k)} \right],
    \label{eq:lem:1:4}
    \end{align}
    where the last inequality is by bounded covariance assumptions and independence across clients.
    Plugging \cref{eq:lem:1:4} back to the conditional expectation of \cref{eq:lem:1:3} and noting that $\eta \leq \frac{1}{4L}$ yield
    \begin{align}
        & \Exs \left[ F( \overline{\vx}^{(t,k+1)} ) - F(\vx^{\star}) \middle| \mathcal{F}^{(t,k)} \right]
        +   \frac{1}{2\eta} \left( \Exs\left[ \left\|\overline{\vx}^{(t,k+1)} - \vx^{\star} \right\|^2 \middle| \mathcal{F}^{(t,k)} \right] 
         - \left\|\overline{\vx}^{(t,k)} - \vx^{\star} \right\|^2  \right)
         \nonumber
        \\
    \leq & \frac{\eta \sigma^2}{M} 
        - \left(\frac{1}{4\eta} - L \right) \Exs \left[ \left\| \overline{\vx}^{(t,k+1)} - \overline{\vx}^{(t,k)} \right\|^2 \middle| \mathcal{F}^{(t,k)} \right]
        +
        \frac{L}{M} \sum_{i=1}^M \left\| \vx_i^{(t,k)} -  \overline{\vx}^{(t,k)} \right\|^2 
         \nonumber
        \\
    \leq & \frac{\eta \sigma^2}{M} + \frac{L}{M} \sum_{i=1}^M \left\| \vx_i^{(t,k)} -  \overline{\vx}^{(t,k)} \right\|^2 
    \tag{since $\eta \leq \frac{1}{4L}$}.
    \end{align}
    Telescoping $k$ from $0$ to $\tau$ completes the proof of \cref{lem:1}.
\end{proof}

\subsection{Deferred Proof of \cref{lem:2}}
\begin{proof}[Proof of \cref{lem:2}]
    \begin{align*}
          & \Exs \left[ \left\| \vx_1^{(t,k+1)} - {\vx_2^{(t,k+1)}} \right\|^2 \middle| \mathcal{F}^{(t,k)} \right]
        = \Exs \left[ \left\| \vx_1^{(t,k)} - {\vx_2^{(t,k)}} - \eta \left( g_1(\vx_1^{(t,k)}) - g_2(\vx_2^{(t,k)})  \right) \right\|^2 \middle| \mathcal{F}^{(t,k)} \right]
        \\
        \leq & \left\| \vx_1^{(t,k)} - {\vx_2^{(t,k)}} \right\|^2 
        - 2 \eta  \left\langle  \nabla F_1(\vx_1^{(t,k)}) - \nabla F_2(\vx_2^{(t,k)})  , \vx_1^{(t,k)} - {\vx_2^{(t,k)}}   \right\rangle
        \\
        & 
        + \eta^2  \left\| \nabla F_1(\vx_1^{(t,k)}) - \nabla F_2(\vx_2^{(t,k)})  \right\|^2 + 2 \eta^2 \sigma^2
        \label{eq:lem:2:1}
    \end{align*}
    Since $ \max_i \sup_{\vx} \|  \nabla F_i(\vx) - \nabla F(\vx) \| \leq \zeta$, the second term is bounded as
    \begin{align*}
        & - \left\langle  \nabla F_1(\vx_1^{(t,k)}) - \nabla F_2(\vx_2^{(t,k)})  , \vx_1^{(t,k)} - {\vx_2^{(t,k)}}   \right\rangle
        \\
        \leq &   - \left\langle  \nabla F(\vx_1^{(t,k)}) - \nabla F(\vx_2^{(t,k)})  , \vx_1^{(t,k)} - {\vx_2^{(t,k)}}   \right\rangle + 2 \zeta \left\| \vx_1^{(t,k)} - {\vx_2^{(t,k)}} \right\| 
        \\
        \leq & - \frac{1}{L} \left\|  \nabla F(\vx_1^{(t,k)}) - \nabla F(\vx_2^{(t,k)}) \right\|^2  + 2 \zeta \left\| \vx_1^{(t,k)} - {\vx_2^{(t,k)}} \right\| 
        \tag{by smoothness and convexity}
        \\
        \leq & - \frac{1}{L} \left\|  \nabla F(\vx_1^{(t,k)}) - \nabla F(\vx_2^{(t,k)}) \right\|^2  + 
        \frac{1}{2 \eta \tau}\left\| \vx_1^{(t,k)} - {\vx_2^{(t,k)}} \right\|^2 + 2 \eta \tau \zeta^2 
        \tag{by AM-GM inequality}
    \end{align*}
    Similarly the third term is bounded as
    \begin{equation}
        \left\|  \nabla F_1(\vx_1^{(t,k)}) - \nabla F_2(\vx_2^{(t,k)})  \right\|^2 
        \leq
        3 \left\|  \nabla F(\vx_1^{(t,k)}) - \nabla F(\vx_2^{(t,k)})  \right\|^2 + 6 \zeta^2.
    \end{equation}
    Replacing the above two bounds back to \cref{eq:lem:2:1} gives (note that $\eta \leq \frac{1}{4L}$)
    \begin{align*}
        \Exs \left[ \left\| \vx_1^{(t,k+1)} - {\vx_2^{(t,k+1)}} \right\|^2 \middle| \mathcal{F}^{(t,k)} \right] 
        & \leq
        \left( 1 + \frac{1}{\tau} \right) \left\| \vx_1^{(t,k)} - {\vx_2^{(t,k)}} \right\| ^2 + 4 \tau \eta^2 \zeta^2 + 6 \eta^2 \zeta^2 + 2 \eta^2 \sigma^2
        \\
        & \leq  \left( 1 + \frac{1}{\tau} \right) \left\| \vx_1^{(t,k)} - {\vx_2^{(t,k)}} \right\| ^2 + 10 \tau \eta^2 \zeta^2 + 2 \eta^2 \sigma^2.
    \end{align*}
    Telescoping
    \begin{equation}
        \Exs \left[ \left\| \vx_1^{(t,k)} - {\vx_2^{(t,k)}} \right\|^2 \middle| \mathcal{F}^{(t,0)} \right] 
        \leq
        \frac{\left( 1 + \frac{1}{\tau} \right)^k - 1}{\frac{1}{\tau}} \cdot \left( 10 \tau \eta^2 \zeta^2 + 2 \eta^2 \sigma^2 \right)
        \leq
        18 \tau^2 \eta^2 \zeta^2 + 4 \tau \eta^2 \sigma^2.
    \end{equation}
    By convexity, for any $i$,
    \begin{equation}
        \Exs \left[ \left\| \vx_i^{(t,k)} - \overline{\vx}^{(t,k)} \right\|^2 \middle| \mathcal{F}^{(t,0)} \right] 
        \leq
        18 \tau^2 \eta^2 \zeta^2 + 4 \tau \eta^2 \sigma^2.
    \end{equation}
\end{proof}

\end{document}